\documentclass{article} %

\PassOptionsToPackage{dvipsnames,table}{xcolor}
\usepackage{xcolor}

\usepackage{iclr2026_conference,times}

\usepackage{graphicx}

\usepackage[utf8]{inputenc} %
\usepackage[T1]{fontenc}    %

\usepackage{float}
\usepackage{wrapfig}

\usepackage[colorlinks,linktoc=all]{hyperref}
\usepackage[all]{hypcap}
\hypersetup{citecolor=blue!50!green}
\hypersetup{linkcolor=Mahogany}
\hypersetup{urlcolor =Mahogany}

\usepackage{url}            %
\usepackage{booktabs}       %
\usepackage[detect-all]{siunitx}
\usepackage{amsfonts}       %
\usepackage{nicefrac}       %
\usepackage{microtype}      %
\usepackage[ruled,linesnumbered,noend]{algorithm2e}

\SetCommentSty{mycommfont}

\usepackage{amssymb,amstext,amsmath,amsthm,amsfonts,pifont,array}
\usepackage{mathtools}

\usepackage{nicematrix}

\usepackage[capitalize,nameinlink]{cleveref}
\crefname{section}{\S}{\S\S}
\Crefname{section}{\S}{\S\S}
\creflabelformat{equation}{#2\textup{#1}#3}

\definecolor{mylightgray}{gray}{0.95}
\definecolor{mylightgray2}{gray}{0.9}

\usepackage{pifont}
\newcommand{\cmark}{\textcolor{green!60!black}{\ding{51}}\xspace}
\newcommand{\xmark}{\textcolor{red!60!black}{\ding{55}}\xspace}

\usepackage{enumitem}

\usepackage[acronym,nowarn]{glossaries}
\setacronymstyle{long-sc-short}
\glsdisablehyper
\makeglossaries
\usepackage{multirow}
\usepackage{tabularx}
\usepackage{mleftright}
\usepackage{xr-hyper}
\usepackage{adjustbox}
\usepackage{footmisc}
\usepackage{arydshln}
\usepackage{amssymb}

\usepackage{listings}

\definecolor{codegreen}{rgb}{0,0.6,0}
\definecolor{codegray}{rgb}{0.5,0.5,0.5}
\definecolor{codepurple}{rgb}{0.58,0,0.82}
\definecolor{backcolour}{rgb}{0.95,0.95,0.92}
\lstdefinestyle{mystyle}{
    backgroundcolor=\color{white},
    numberstyle=\tiny\color{codegray},
    stringstyle=\color{codepurple},
    basicstyle=\fontsize{7.5pt}{7.5pt}\ttfamily\selectfont,
    commentstyle=\fontsize{7.5pt}{7.5pt}\color{codegreen},
    keywordstyle=\fontsize{7.5pt}{7.5pt}\color{magenta},
    breakatwhitespace=false,
    breaklines=true,
    captionpos=b,
    keepspaces=true,
    numbers=left,
    numbersep=5pt,
    showspaces=false,
    showstringspaces=false,
    showtabs=false,
    tabsize=2
}
\usepackage{epstopdf,caption,tablefootnote}
\usepackage{tikz,tikzscale,subcaption}
 \usetikzlibrary{arrows.meta,calc,decorations.markings,math,arrows.meta,shapes,arrows}
 \tikzset{%
             base/.style = {rectangle, draw=black,
                            minimum width=4cm, minimum height=1cm,
                            text centered, font=\rmfamily},
   binary/.style = {base, minimum width=1cm},
        startstop/.style = {base, fill=red!30, minimum width=2cm},
     activityRuns/.style = {base, fill=green!30},
          process/.style = {base, minimum width=2cm, fill=gray!15,
                            font=\ttfamily},
          sum/.style      = {draw, circle, node distance = 1.5cm},
 }
 \usetikzlibrary{positioning}
 \usetikzlibrary{shapes.misc}
 \usetikzlibrary{automata,arrows,positioning,calc}

\usepackage{pgfplots}
\pgfplotsset{compat=1.6}
\usepgfplotslibrary{groupplots}
\tikzstyle{every picture}+=[font=\rmfamily]
\tikzstyle{optimized} = [circle,fill=white,draw=black, dashed,inner sep=1pt, minimum size=20pt, font=\fontsize{10}{10}\selectfont, node distance=1]
\pgfkeys{/pgf/number format/.cd,1000 sep={}}
\pgfplotsset{
	tick label style = {font=\rmfamily},
	every axis label/.append style={font=\rmfamily},
	typeset ticklabels with strut,
}
\pgfplotsset{every axis/.append style={
			every x tick label/.append style={font=\fontsize{6pt}{6pt}\rmfamily, yshift=.5ex,},
			every y tick label/.append style={font=\fontsize{6pt}{6pt}\rmfamily, xshift=.5ex},
			every y label/.append style={xshift=10ex, font=\rmfamily},
			every x label/.append style={yshift=3ex, font=\rmfamily},
			every title/.append style={font=\rmfamily}
		},
}
\pgfplotsset{
  xticklabel={$\mathsf{\pgfmathprintnumber{\tick}}$},
  yticklabel={$\mathsf{\pgfmathprintnumber{\tick}}$},
}
\pgfplotsset{every axis title/.append style={yshift=-1.5ex}}
\newlength\figureheight
\newlength\figurewidth
\usepgfplotslibrary{external}
\usepackage{todonotes}

\definecolor{color_binarynet}{HTML}{bcbd22}
\definecolor{color_binaryconnect}{HTML}{17becf}
\definecolor{color_xnornet}{HTML}{2ca02c}
\definecolor{color_birealnet}{HTML}{e377c2}
\definecolor{color_real2binary}{HTML}{8c564b}
\definecolor{color_reactnet}{HTML}{9467bd}
\definecolor{color_meliusnet}{HTML}{7f7f7f}
\definecolor{color_bnext}{HTML}{ff7f0e}
\definecolor{color_pokebnn}{HTML}{1f77b4}
\definecolor{color_booldl}{HTML}{d62728}

\usepackage{tcolorbox}
\definecolor{lightblue}{HTML}{ff7f2a}
\definecolor{lighterblue}{HTML}{ffe6d5}%
\newtcolorbox{mybox}{colback=mylightgray2,colframe=mylightgray2,top=1.2pt,bottom=1.2pt,right=1.8pt,left=1.8pt}
\newtcolorbox{myboxcontrib}{colback=mylightgray2,colframe=mylightgray2,top=1.2pt,bottom=1.2pt,right=1.8pt,left=1.8pt}

\definecolor{asparagus}{rgb}{0.53, 0.66, 0.42}
\definecolor{b_df_c}{rgb}{0, 0.69, 0.31}
\definecolor{b_block_c}{rgb}{0.57, 0.81, 0.31}
\definecolor{fp_df_c}{rgb}{0.78, 0, 0.04}
\definecolor{fp_block_c}{rgb}{0.78, 0, 0.04}
\definecolor{int_df_c}{rgb}{0, 0.2, 0.8}

\usepackage{setspace}

\usepackage{algorithmic}

\usepackage{amsmath}
\usepackage{mathtools, nccmath}
\usepackage{color}

\definecolor{color_blue}{HTML}{1f77b4}
\definecolor{color_orange}{HTML}{ff7f0e}
\definecolor{color_green}{HTML}{2ca02c}
\definecolor{color_red}{HTML}{d62728}
\definecolor{color_purle}{HTML}{9467bd}
\definecolor{color_brown}{HTML}{8c564b}
\definecolor{color_olive}{RGB}{188,189,34}

\newacronym{KD}{kd}{knowledge distillation}
\newacronym{BNN}{bnn}{binarized neural network}
\newacronym{STE}{ste}{straight-through-estimator}
\newacronym{SE}{se}{squeeze-and-excitation}
\newacronym{CNN}{cnn}{convolutional neural network}
\newacronym{FP}{fp}{full-precision}
\newacronym{SOTA}{sota}{state-of-the-art}
\newacronym{LLM}{llm}{large language model}
\newacronym{OP}{op}{compute operation}
\newacronym{NN}{nn}{neural network}
\newacronym{BN}{bn}{batch normalization}
\newacronym{FC}{fc}{fully-connected}
\newacronym{ASPP}{aspp}{atrous pyramid pooling}
\newacronym{MSE}{mse}{mean-squared error}
\newacronym{SVID}{svid}{sign-value-independent decomposition}
\newacronym{SVD}{svd}{singular value decomposition}
\newacronym{PWCCA}{pwcca}{projection weighted canonical correlation analysis}
\newacronym{KL}{kl}{Kullback–Leibler}
\newacronym{PTQ}{ptq}{Post-Training Quantization}
\newacronym{QAT}{qat}{Quantization-Aware Training}

\newacronym{MBOK}{mbok}{Multiple Boolean Kernels}

\newacronym{CCA}{cca}{Canonical Correlation Analysis}

\newcommand{\lora}{{\textsc{l}{\small{o}}\textsc{ra}}\xspace}
\newcommand{\billm}{{\textsc{b}{\small{i}}\textsc{llm}}\xspace}
\newcommand{\pbllm}{{\textsc{pb}{\small{-}}\textsc{llm}}\xspace}
\newcommand{\stbllm}{{\textsc{stbllm}}\xspace}
\newcommand{\qptq}{{\textsc{optq}}\xspace}
\newcommand{\qbb}{{\textsc{qbb}}\xspace}
\newcommand{\dbllm}{{\textsc{db}{\small{-}}\textsc{llm}}\xspace}
\newcommand{\arbllm}{{\textsc{arb}{\small{-}}\textsc{llm}}\xspace}
\newcommand{\bitstack}{{\textsc{b}{\small{it}}\textsc{s}{\small{tack}}}\xspace}
\newcommand{\onebit}{{\textsc{o}{\small{ne}}\textsc{b}{\small{it}}}\xspace}
\newcommand{\bitnet}{{\textsc{b}{\small{it}}\textsc{n}{\small{et}}}\xspace}
\newcommand{\bitnetternary}{{\textsc{b}{\small{it}}\textsc{n}{\small{et-b1.58}}}\xspace}
\newcommand{\bitlinear}{{\textsc{b}{\small{it}}\textsc{l}{\small{inear}}}\xspace}
\newcommand{\mos}{{\textsc{m}{\small{o}}\textsc{s}}\xspace}
\newcommand{\omniquant}{{\textsc{o}{\small{mni}}\textsc{q}{\small{uant}}}\xspace}
\newcommand{\llmqat}{\textsc{llm}{\small{-}}\textsc{qat}\xspace}
\newcommand{\ours}{\textsc{mbok}\xspace}
\newcommand{\shiftaddllm}{{\textsc{s}{\small{hift}}\textsc{a}{\small{dd}}\textsc{llm}}\xspace}
\newcommand{\quip}{{\textsc{q}{\small{u}}\textsc{ip}}\xspace}
\newcommand{\awq}{{\textsc{awq}}\xspace}

\newcommand{\quipsharp}{{\textsc{quip\#}}\xspace}
\newcommand{\qtip}{{\textsc{qtip}}\xspace}

\newcommand{\optsmall}{\textsc{opt}{\small{-125}}\textsc{m}\xspace}
\newcommand{\optmedium}{\textsc{opt}{\small{-350}}\textsc{m}\xspace}
\newcommand{\optbig}{\textsc{opt}{\small{-1.3}}\textsc{b}\xspace}
\newcommand{\opthuge}{\textsc{opt}{\small{-6.7}}\textsc{b}\xspace}

\newcommand{\llamasmall}{\textsc{ll}{\small{a}}{\textsc{ma}}{\small{-7}}\textsc{b}\xspace}
\newcommand{\llamabig}{\textsc{ll}{\small{a}}{\textsc{ma}}{\small{-13}}\textsc{b}\xspace}
\newcommand{\halffp}{\textsc{fp}{\small{16}}\xspace}
\newcommand{\llamatwo}{\textsc{ll}{\small{a}}{\textsc{ma}}{\small{2}}\xspace}
\newcommand{\llamatwosmall}{\textsc{ll}{\small{a}}{\textsc{ma}}{\small{2-7}}\textsc{b}\xspace}
\newcommand{\llamatwobig}{\textsc{ll}{\small{a}}{\textsc{ma}}{\small{2-13}}\textsc{b}\xspace}

\newcommand{\booll}{\textsc{bool}\xspace}

\newcommand{\galore}{\textsc{g}{\small{a}}\textsc{l}{\small{ore}}\xspace}

\newcommand{\true}{\text{\textsc{true}\xspace}}
\newcommand{\false}{\text{\textsc{false}\xspace}}

\theoremstyle{plain}
\newtheorem{theorem}{Theorem}[section]
\newtheorem{proposition}[theorem]{Proposition}
\newtheorem{lemma}[theorem]{Lemma}

\newtheorem{definition}[theorem]{\textit{Definition}}

\theoremstyle{remark}
\newtheorem{remark}[theorem]{\textit{Remark}}%
\newtheorem{example}[theorem]{\textit{Example}}%
\newtheorem{notation}[theorem]{\textit{Notation}}%

\newcommand{\True}{\mathrm{True}}
\newcommand{\False}{\mathrm{False}}
\newcommand{\xor}{\mathbf{xor}}
\newcommand{\xnor}{\mathbf{xnor}}
\newcommand{\andd}{\mathbf{and}}

\newcommand{\logic}{\mathrm{logic}}

\newcommand{\pren}[1]{\mleft(#1\mright)}

\usepackage{xargs}

\newcommand{\bvar}[1]{\delta#1} %

\newcommand{\trans}[1]{#1^{\mathrm{T}}}

\def\sign{\operatorname{sign}}

\def\proj{\operatorname{p}}
\def\emb{\operatorname{e}}

\makeatletter

\def\env@sqcases{%
	\let\@ifnextchar\new@ifnextchar
	\left\lbrack
	\def\arraystretch{1.2}%
	\array{@{}l@{\quad}l@{}}%
}
\makeatother

\newcommand{\mathbold}[1]{{\boldsymbol{\mathbf{#1}}}}

\newcommand{\nestedmathbold}[1]{{\mathbold{#1}}}

\newcommand{\mba}{\nestedmathbold{a}}
\newcommand{\mbb}{\nestedmathbold{b}}
\newcommand{\mbc}{\nestedmathbold{c}}
\newcommand{\mbd}{\nestedmathbold{d}}

\newcommand{\mbf}{\nestedmathbold{f}}
\newcommand{\mbg}{\nestedmathbold{g}}
\newcommand{\mbh}{\nestedmathbold{h}}

\newcommand{\mbk}{\nestedmathbold{k}}
\newcommand{\mbl}{\nestedmathbold{l}}

\newcommand{\mbp}{\nestedmathbold{p}}

\newcommand{\mbs}{\nestedmathbold{s}}

\newcommand{\mbw}{\nestedmathbold{w}}
\newcommand{\mbx}{\nestedmathbold{x}}
\newcommand{\mby}{\nestedmathbold{y}}

\newcommand{\mbA}{\nestedmathbold{A}}

\newcommand{\mbC}{\nestedmathbold{C}}
\newcommand{\mbD}{\nestedmathbold{D}}
\newcommand{\mbE}{\nestedmathbold{E}}

\newcommand{\mbG}{\nestedmathbold{G}}

\newcommand{\mbM}{\nestedmathbold{M}}

\newcommand{\mbP}{\nestedmathbold{P}}
\newcommand{\mbQ}{\nestedmathbold{Q}}

\newcommand{\mbS}{\nestedmathbold{S}}

\newcommand{\mbU}{\nestedmathbold{U}}
\newcommand{\mbV}{\nestedmathbold{V}}
\newcommand{\mbW}{\nestedmathbold{W}}
\newcommand{\mbX}{\nestedmathbold{X}}
\newcommand{\mbY}{\nestedmathbold{Y}}
\newcommand{\mbZ}{\nestedmathbold{Z}}

\newcommand{\mbSigma}{\nestedmathbold{\Sigma}}

\DeclarePairedDelimiterX{\infdivx}[2]{[}{]}{%
  #1\;\delimsize\|\;#2%
}

\DeclareMathOperator*{\argmin}{arg\,min}

\newcommand{\cE}{\mathcal{E}}
\newcommand{\cL}{\mathcal{L}}

\newcommand{\cF}{\mathcal{F}}

\newcommand{\cK}{\mathcal{K}}

\newcommand{\bbR}{\mathbb{R}}
\newcommand{\bbD}{\mathbb{D}}
\newcommand{\bbL}{\mathbb{L}}
\newcommand{\bbM}{\mathbb{M}}
\newcommand{\bbN}{\mathbb{N}}
\newcommand{\bbZ}{\mathbb{Z}}
\newcommand{\bbB}{\mathbb{B}}

\def\Lb{\mathbb{L}}
\def\Lm{\mathrm{L}}

\newcommand{\wbool}{\mbW_{\mathrm{bool}}}
\newcommand{\wfp}{\mbW_{\mathrm{FP}}}
\newcommand{\ssin}{\mbs_{\mathrm{in}}}
\newcommand{\ssout}{\mbs_{\mathrm{out}}}

\newcommand{\ufp}{\mathrm{FP}}
\newcommand{\uout}{\mathrm{out}}
\newcommand{\uin}{\mathrm{in}}

\usepackage{url}

\title{Highly Efficient and Effective LLMs with\\Multi-Boolean Architectures}

\author{%
Ba-Hien Tran \& Van Minh Nguyen \\ 
Huawei Paris Research Center \\
Paris, France \\
\texttt{ba.hien.tran@huawei.com} \small \textit{(corresponding author)} \\
See \cref{sec:author_contributions} for detailed contributions.
}

\iclrfinalcopy %
\begin{document}

\addtocontents{toc}{\protect\setcounter{tocdepth}{0}}

\vspace{-6ex}

\maketitle

\vspace{-5ex}

\begin{abstract}
\vspace{-1.5ex}

Weight binarization has emerged as a promising strategy to reduce the complexity of large language models (LLMs).
Existing approaches fall into post-training binarization, which is simple but causes severe performance loss, and training-aware methods, which depend on full-precision latent weights, adding complexity and limiting efficiency.
We propose a novel framework that represents LLMs with multi-kernel Boolean parameters and, for the first time, enables direct finetuning LMMs in the Boolean domain, eliminating the need for latent weights.
This enhances representational capacity and dramatically reduces complexity during both finetuning and inference.
Extensive experiments across diverse LLMs show our method outperforms recent ultra low-bit quantization and binarization techniques.

\end{abstract}

\vspace{-0.5ex}

\vspace{-2.1ex}

\section{Introduction} \label{sec:introduction}

\vspace{-1.5ex}

\begin{wrapfigure}[16]{r}{0.3\textwidth}
    \centering

        \vspace{-2ex}

        \tikzexternaldisable
        \centering
        \scriptsize
        \setlength{\figurewidth}{4.75cm}
        \setlength{\figureheight}{4.2cm}
        \begin{tikzpicture}

  \definecolor{darkgray176}{RGB}{176,176,176}
  \definecolor{darkorange25512714}{RGB}{255,127,14}
  \definecolor{forestgreen4416044}{RGB}{44,160,44}
  \definecolor{steelblue31119180}{RGB}{31,119,180}

  \begin{groupplot}[group style={group size=1 by 1, horizontal sep=5pt}]

    \nextgroupplot[
      height=\figureheight,
      major tick length=1ex,
      tick align=outside,
      tick pos=left,
      width=\figurewidth,
      x grid style={darkgray176},
      xlabel={Model Size (GB)},
      xmajorgrids,
      xmin=-0.489051273465157, xmax=13.0162516742945,
      xtick style={color=black},
      xtick={-2.5,0,2.5,5,7.5,10,12.5,15},
      xticklabels={
          \(\displaystyle {\ensuremath{-}2.5}\),
          \(\displaystyle {0.0}\),
          \(\displaystyle {2.5}\),
          \(\displaystyle {5.0}\),
          \(\displaystyle {7.5}\),
          \(\displaystyle {10.0}\),
          \(\displaystyle {12.5}\),
          \(\displaystyle {15.0}\)
        },
      y grid style={darkgray176},
      ylabel={Perplexity on C4 (\(\displaystyle \leftarrow\))},
      ymajorgrids,
      ymin=10.826, ymax=43.914,
      ytick style={color=black},
      ytick={10,15,20,25,30,35,40,45},
      yticklabels={
          \(\displaystyle {10}\),
          \(\displaystyle {15}\),
          \(\displaystyle {20}\),
          \(\displaystyle {25}\),
          \(\displaystyle {30}\),
          \(\displaystyle {35}\),
          \(\displaystyle {40}\),
          \(\displaystyle {45}\)
        }
    ]
    \addplot [thick, steelblue31119180, opacity=0.8, mark=pentagon*, mark size=1.75, mark options={solid}]
    table {%
        0.2332763671875 26.56
        0.616901397705078 22.59
        2.45079040527344 16.07
        4.93898391723633 14.34
        12.4023742675781 12.71
      };
    \addplot [thick, darkorange25512714, opacity=0.8, mark=*, mark size=1.75, mark options={solid}]
    table {%
        0.12482613325119 42.41
        0.234117269515991 31.33
        0.77751350402832 21.63
        1.32354736328125 18.17
        2.96142578125 17.14
      };
    \addplot [thick, forestgreen4416044, opacity=0.8, mark=triangle*, mark size=1.75, mark options={solid}]
    table {%
        0.12482613325119 28.62
        0.234117269515991 22.1
        0.77751350402832 15.68
        1.32354736328125 14
        2.96142578125 12.33
      };
  \end{groupplot}

\end{tikzpicture}
        \tikzexternalenable

        \tikzexternaldisable

        \vspace{-4.3ex}

        \captionof{figure}{\small Finetuning \textsc{opt} models \citep{zhang2022opt} using our 3 Boolean kernels
            ({\protect\tikz[baseline=-.65ex] \protect\draw[fill=color_green, draw=color_green, line width=1.4pt] plot[mark size=1.6, mark=triangle*, only marks] (0, 0) --+(.18, 0) --+ (-.18, 0);}),
            compared to \qptq \citep{frantar2023optq}
            ({\protect\tikz[baseline=-.65ex] \protect\draw[fill=color_orange, draw=color_orange, line width=1.4pt] plot[mark size=1.6, mark=triangle*, only marks] (0, 0) --+(.18, 0) --+ (-.18, 0);}),
            which quantizes the models to 3 bits, and the \halffp baseline
            ({\protect\tikz[baseline=-.65ex] \protect\draw[fill=color_blue, draw=color_blue, line width=1.4pt] plot[mark size=1.6, mark=pentagon*, only marks] (0, 0) --+(.18, 0) --+ (-.18, 0);})
            on the C4 dataset. \label{fig:opt_scaling}}

        \tikzexternalenable
\end{wrapfigure}

Large language models \citep{Brown2020,touvron2023llama,liu2024deepseek} have demonstrated unprecedented capabilities, largely due to the continuous growth in both model and dataset sizes.
A key area of focus in optimizing these models is lower-precision computation, which offers substantial benefits in terms of memory and computational efficiency.
One prominent approach to achieving this is through the quantization of weight parameters, which reduces the model size by lowering the precision of the weight values.
Recent studies on scaling laws \citep{dettmers2023thecase, kumar2025scaling} have highlighted the potential of using low-precision techniques for \glspl{LLM}.

\vspace{-0.3ex}

Binarization represents one of the most extreme forms of quantization for \glspl{LLM}.
While significant progress has been made, challenges remain \citep{yuan2024pbllm,huang2024billm,li2025arbllm}.
Even with advanced techniques like \gls{QAT}, which fine-tunes the model extensively after binarization \citep{xu2024onebit,jo2024mixture}, or trains it from scratch \citep{wang2023bitnet}, performance still lags behind that of \gls{FP} models.
This performance gap can be attributed to the limited representation capacity of binary weights and the heavy reliance on \gls{FP} latent weights for binarization.
This reliance not only makes the approach computationally expensive but also suboptimal, as it requires gradient approximation.
Meanwhile, recent advances in 4-bit quantization have achieved remarkable compression with minimal accuracy loss, but further compression or applying these methods to smaller models has yielded unsatisfactory results \citep{frantar2023optq,Lin2024AWQ}.

\vspace{-0.3ex}

In this paper, we aim to push the boundary of low-precision \glspl{LLM} by proposing a novel method named as \gls{MBOK}.
We extend the work in \cite{nguyen2024bold}, which proposes training neural networks with native Boolean weights directly in the Boolean domain,
However, effectively applying this approach to \glspl{LLM} remains a key challenge.
In particular, our contributions are:

\vspace{-1.5ex}

{
    \setlist[itemize]{leftmargin=2mm}
    \begin{itemize}
        \item We propose the framework \ours, which employs multiple Boolean kernels, each using distinct Boolean weights (\cref{sec:multi_kernel_boolean}).
        This allows for flexibly representing \glspl{LLM} with low bits, while approaching to \gls{FP} performance with minimal \emph{both} finetuning and inference cost.
        The Boolean weights are directly trained in Boolean domain, avoiding the need for \gls{FP} latent weights and gradient approximations.
        
        \vspace{-0.5ex}
        
        \item We propose a novel successive method that effectively transfers knowledge from an \gls{FP} \gls{LLM} into the Boolean model (\cref{sec:knowledge_transfer}), followed by further fine-tuning using knowledge distillation (\cref{sec:knowledge_distillation}).
        
        \vspace{-0.5ex}
        
        \item We introduce a method for automatically allocating the number of kernels for each weight (\cref{sec:kernel_allocation}), supporting any average bit-width, including fractional values.
        
        \vspace{-0.5ex}

        \item We provide a comprehensive empirical analysis and benchmarks, demonstrating our method's superior performance over recent binarization and quantization approaches (see \cref{sec:experiment}) with much lower memory and computational overhead.
        For example, \cref{fig:opt_scaling} shows that our method achieves the best accuracy-compression trade-off, outperforming \gls{FP} and existing quantization techniques.
    \end{itemize}

}

\vspace{-2.5ex}

\section{Related Works} \label{sec:related_work}

\vspace{-1.5ex}

\paragraph{LLMs quantization.}

Quantization techniques are commonly used to reduce the memory and latency of \glspl{LLM}.
They fall into two categories: \gls{QAT}, which involves retraining or finetuning in a quantized form, and \gls{PTQ}, which can be applied directly without retraining.
Due to the difficulty of retraining such large models, most work focuses on \gls{PTQ} \citep{frantar2023optq, Sheng2023FlexGen, Lin2024AWQ, lee2024owq}, though recent efforts also explore \gls{QAT} via data-free methods (\llmqat \citep{liu2024llmqat}), or parameter-efficient fine-tuning like \lora \citep{Dettmers2023QLoRA}.
A promient \gls{PTQ} method is \qptq \citep{frantar2023optq}, which introduces one-shot low-bit weight quantization using approximate second-order information.
Follow-up work refines this by addressing outliers \citep{Kim2024SqueezeLLM, dettmers2024spqr}, accounting for activation effects \citep{Lin2024AWQ, lee2024owq}, and optimizing quantization parameters (\omniquant \citep{shao2024omniquant}).
However, effective \glspl{LLM} quantization is still challenging \citep{xu2025understanding}.

\vspace{-1.5ex}

\paragraph{Binarization.} 
This represents the most extreme form of quantization, typically using the $\mathbf{sign}(\cdot)$ function with gradients estimated via the \gls{STE} \citep{bengio2013ste}.
Early work focused on small Transformer models \citep{vaswani2017attention} trained or fine-tuned on labeled data \citep{bai2021binarybert, qin2022bibert, Liu2022BiT, liu2023binary}.
Recent efforts have extended binarization to \glspl{LLM}.
Methods like \billm \citep{huang2024billm}, \pbllm \citep{yuan2024pbllm}, \stbllm \citep{dong2025stbllm}, and \arbllm \citep{li2025arbllm} adopt hybrid \gls{PTQ} approaches, binarizing non-salient weights while using higher precision for important ones, with calibration data used to adjust scaling factors. 
\bitstack \citep{wang2025bitstack}, \qbb \citep{bulat2024qbb}, \dbllm \citep{chen2024dbllm} further improve this with multiple binary bases, either through a training-free method or via knowledge distillation.
In contrast, \bitnet \citep{wang2023bitnet} replaces linear layers with a custom 1-bit weight structure, \bitlinear, and trains the model from scratch. 
\onebit \citep{xu2024onebit}, which decomposes weights into 1-bit components and scaling vectors for \gls{QAT}, further enhanced by \mos \citep{jo2024mixture} using a mixture of scalings. 
Despite progress, these methods remain costly due to their dependence on \gls{FP} latent weights during training.
\cref{tab:comparison} summarizes the key characteristics of these methods in comparison to ours.

\vspace{-1.8ex}

\begin{table}[H]
    \setlength{\tabcolsep}{2.8pt}
    \centering
    \caption{\small A summary of \textsc{sota} binarization methods for \glspl{LLM} compared to our method.}
    \label{tab:comparison}

    \vspace{-1.5ex}
    \renewcommand{\arraystretch}{1.0}

    \scalebox{.7}{
        \begin{tabular}{rccccccc}
            \toprule
            Method                             & \begin{tabular}[c]{@{}c@{}}Train\\ from Scratch\end{tabular} & \begin{tabular}[c]{@{}c@{}}Post-training\\ Binarization\end{tabular} & \begin{tabular}[c]{@{}c@{}}Finetune from\\ FP Model\end{tabular} & \begin{tabular}[c]{@{}c@{}}Calibration\\ Data\end{tabular} & \begin{tabular}[c]{@{}c@{}}Weight\\ Update\end{tabular} & \begin{tabular}[c]{@{}c@{}}Multiple\\ Binary Bases\end{tabular} & \begin{tabular}[c]{@{}c@{}}Higher-bit \\ Salient Weights\end{tabular} \\
            \midrule
            \bitnet \citep{wang2023bitnet}     & \cmark                                                       & \xmark                                                               & \xmark                                                           & \textsc{NA}                                                & \small{FP latent-weights}                               & \xmark                                                          & \xmark                                                                \\
            \billm \citep{huang2024billm}      & \xmark                                                       & \cmark                                                               & \xmark                                                           & \cmark                                                     & \textsc{NA}                                             & \cmark                                                          & \cmark                                                                \\
            \pbllm \citep{yuan2024pbllm}       & \xmark                                                       & \cmark                                                               & \xmark                                                           & \cmark                                                     & \textsc{NA}                                             & \xmark                                                          & \cmark                                                                \\
            \stbllm \citep{dong2025stbllm}     & \xmark                                                       & \cmark                                                               & \xmark                                                           & \cmark                                                     & \textsc{NA}                                             & \cmark                                                          & \cmark                                                                \\
            \arbllm \citep{li2025arbllm}       & \xmark                                                       & \cmark                                                               & \xmark                                                           & \cmark                                                     & \textsc{NA}                                             & \cmark                                                          & \cmark                                                                \\
            \bitstack \citep{wang2025bitstack} & \xmark                                                       & \cmark                                                               & \xmark                                                           & \xmark                                                     & \small{NA}                                              & \cmark                                                          & \xmark                                                                \\
            \dbllm \citep{chen2024dbllm}       & \xmark                                                       & \cmark                                                               & \cmark                                                           & \cmark                                                     & \small{FP latent-weights}                               & \cmark                                                          & \xmark                                                                \\
            \qbb \citep{bulat2024qbb}          & \xmark                                                       & \cmark                                                               & \cmark                                                           & \cmark                                                     & \small{FP latent-weights}                               & \cmark                                                          & \xmark                                                                \\
            \onebit \citep{xu2024onebit}       & \xmark                                                       & \xmark                                                               & \cmark                                                           & \cmark                                                     & \small{FP latent-weights}                               & \xmark                                                          & \xmark                                                                \\
            \mos \citep{jo2024mixture}         & \xmark                                                       & \xmark                                                               & \cmark                                                           & \cmark                                                     & \small{FP latent-weights}                               & \xmark                                                          & \xmark                                                                \\
            \midrule
            \ours [Ours]                       & \xmark                                                       & \xmark                                                               & \cmark                                                           & \cmark                                                     & \small{Native Boolean weights}                          & \cmark                                                          & \xmark                                                                \\
            \bottomrule
        \end{tabular}
    }

    \vspace{-2.7ex}
\end{table}

\section{Preliminaries}

\vspace{-1.2ex}

\paragraph{Notations.}
We use a standard notation for vectors ($\mba$), matrices ($\mbA$), and scalars ($a$).
The $i$-th element of a vector $\mba$ is $\mba_{[i]}$, and the element at the $i$-th row and $j$-th column of a matrix $\mbA$ is $\mbA_{[i,j]}$.
The symbol $\odot$ denotes element-wise multiplication, with broadcasting if needed.

\vspace{-1.2ex}

\subsection{Pitfalls of Full-Precision Latent Weights for Binarization}

\vspace{-0.5ex}

Binarization is an effective technique for reducing both the size and computation of deep learning models by converting high-precision weight parameters into 1-bit values \citep{Hubara2016, Courbariaux2015, Rastegari2016}.
For a linear layer, $\mbY = \mbX \mbW_{\mathrm{FP}}^\top + \mbb$, where $\mbX_{\mathrm{FP}} \in \mathbb{R}^{b \times n}$ is the input data, and $\mbW \in \mathbb{R}^{m \times n}$ with the input size $n$ and output size $m$, and $\mbb \in \mathbb{R}^{m}$ are the \gls{FP} weights and bias.
Binarization results in $\mbY = \alpha \cdot \mbX \mbW_{\mathrm{bin}}^\top  + \mbb$, with $\mbW_{\mathrm{bin}} = \mathbf{sign}(\mbW_{\mathrm{FP}})$ and $\alpha$ as a scaling factor (e.g., $\alpha = \frac{\|\mbW_{\mathrm{FP}}\|_1}{m \times n}$) \citep{Rastegari2016}.

\vspace{-0.5ex}

During training, the \gls{FP} weights must be retained for learning the binarized weights.
In vanilla gradient descent, binarized weights are updated as $\mbW_{\mathrm{bin}} = \mathbf{sign}(\mbW_{\mathrm{FP}} - \eta \cdot \mbG_{\mbW_{\mathrm{FP}}})$, where $\eta$ is the learning rate and $\mbG_{\mbW_{\mathrm{FP}}}$ is the gradient of the \gls{FP} weights.
This leads to high memory usage, especially with optimizers like Adam \citep{Kingma2014}, which require storing two additional \gls{FP} momenta for each parameter.
Moreover, the gradient approximation for binarized weights often uses a differentiable proxy, like the \gls{STE} \citep{bengio2013ste}, but this introduces performance drops due to proxy gradient noise.
This noise can cause oscillations and instability during training. %

\vspace{-1.9ex}

\subsection{Native Boolean Framework for Neural Networks}

\vspace{-1.4ex}

To address the issues associated with latent-weight-based approaches, \cite{nguyen2024bold} recently proposed a principled framework for directly training Boolean neural networks in the Boolean domain.
Consider the $l$-th Boolean linear layer; in the forward pass, the output of the next layer is defined as:

\vspace{-2.1ex}
\begin{align}
    \mbY_{[k,j]}^{(l)} =  \mbb_{[j]}^{(l)} + \sum_{i=1}^{n} \mathrm{L}(\mbX_{[k,i]}^{(l)}, \mbW_{[i,j]}^{(l)}), \qquad 1 \leq j \leq m,
\end{align}
\vspace{-2.1ex}

where $k$ denotes the sample index in the batch, and $\mathrm{L}$ is a logic gate such as $\mathbf{and}, \mathbf{or}, \mathbf{xor}$, or $\mathbf{xnor}$;
Hereafter, for clarity, we consider $\mathrm{L} = \mathbf{xnor}$ as a concrete example.
The weights $\mbW_{[i,j]}^{(l)}$ are Boolean values $\{\textsc{true}, \textsc{false}\}$ or $\{-1, +1\}$, as typically used in practical implementations. 

\vspace{-0.5ex}

The logic gate $\textrm{L}$ can be extended to handle mixed-type data.
In this paper, we focus on the case where the input data is real-valued, and the weights are Boolean.
Specifically, for an input element $x \in \mathbb{R}$, we define $x_{\textrm{bool}}=\textsc{true} \Leftrightarrow x \geq 0$, and $x_{\textrm{bool}}=\textsc{false} \Leftrightarrow x < 0$, and $|x|$ its magnitude.
The logic operation between a real input $x \in \mathbb{R}$ and a Boolean weight $w \in \mathbb{B}$ is defined as $\textbf{xnor}(w, x) \triangleq s$ such that $s_{\mathrm{bool}} = \mathbf{xnor}(w_{\mathrm{bool}}, x)$ and $|s| = |x|$.

\vspace{-1.5ex}

\paragraph{Backward pass.}
This layer receives the backpropagation signal from the downstream layer.
Specifically, $\mbZ^{(l)}_{[k,j]} \triangleq \frac{\delta \cL}{\delta \mbY_{[k,j]}^{(l)}}$ denotes the variation of the loss function $\cL$ w.r.t. the output at layer $l$.
To optimize the Boolean weights, we need to compute the corresponding loss signal, denoted as $\mbQ^{(l)}_{[i,j]} \triangleq \frac{\delta \cL}{\delta \mbW_{[i,j]}^{(l)}}$, which is aggregated over the batch dimension $k$ as:

\vspace{-2.9ex}
\begin{align}
   \hspace{-5ex} \mbQ^{(l)}_{[i,j]} = \sum_{k=1}^{b} \mathbf{1} ( \mbQ_{[k,i,j]}^{(l)} = \textsc{true} ) |\mbQ_{[k,i,j]}^{(l)}| - \sum_{k=1}^{b} \mathbf{1} ( \mbQ_{[k,i,j]}^{(l)} = \textsc{false} ) |\mbQ_{[k,i,j]}^{(l)}|, %
\end{align}

\vspace{-1.8ex}

where $\mbQ_{[i,j,k]}^{(l)} = \mathbf{xnor}(\mbZ_{[k,j]}^{(l)}, \mbX_{[k,i]}^{(l)})$, and $\mathbf{1}(\cdot)$ is the indicator function.
The backpropagation signal for the upstream layer, $\mbP^{(l)}_{[k,j]} \triangleq \frac{\delta \cL}{\delta \mbX_{[k,j]}^{(l)}}$, can be computed in a similar manner. 

\vspace{-2.7ex}

\paragraph{Boolean optimizer.}
Given the loss signal, the rule for updating the Boolean weight $\mbW_{[i,j]}^{(l)}$ to minimize the loss function $\cL$ is as  $\mbW_{[i,j]}^{(l)} = \lnot \mbW_{[i,j]}^{(l)}$ if $\mathbf{xnor}(\mbQ^{(l)}_{[i,j]}, \mbW^{(l)}_{[i,j]}) = \textsc{true}$.
Based on this update rule, we can develop an optimizer that accumulates the signal $\mbQ_{[i,j]}^{(l)}$ over training iterations.
Specifically, let $\mbW_{[i,j]}^{(l),t}$ denotes the weight at iteration $t$, and $\mbM_{[i,j]}^{(l),t}$ represents its accumulator, initialized as $\mbM_{[i,j]}^{(l),0} = 0$.
The update rule for the accumulator is then defined as:

\vspace{-1.5ex}
\begin{align}
    \mbM_{[i,j]}^{(l),t+1} \leftarrow \beta^{t} \mbM_{[i,j]}^{(l),t} + \eta \mbQ_{[i,j]}^{(l),t}, \label{eq:bool_optimizer_main}
\end{align}

\vspace{-1.1ex}

where $\eta$ is the accumulation factor acting as a learning rate, and $\beta^{t}$ is a regularizing factor that reflects the system's state at time $t$. 
In our work, we use brain plasticity \citep{Fuchs2014} and Hebbian theory \citep{Hebb2005} to adaptively set $\beta^{t}$.
We encourage the reader check \cref{sec:app_bool_nn} for details.

\vspace{-1.8ex}

\paragraph{Remarks on complexity and applicability to LLMs.}
This Boolean framework optimizes Boolean parameters $\mbW^{(l)}_{[i,j]}$ directly in the Boolean space, eliminating the need for \gls{FP} latent weights.
As shown in \cref{eq:bool_optimizer_main}, the Boolean optimizer is more lightweight than common \gls{LLM} optimizers like Adam, requiring only one \gls{FP} momentum per parameter.
This reduces both training and inference complexity and avoids gradient approximation induced from \gls{STE}.
As shown in \cref{prop:XNORAlgebra} in Appendix, $\mathbf{xnor}(w, s) = w \times s$, mathematically enabling direct application to existing linear algebra operations.
Practically, native logic operations are much faster than multiplication.

\section{Multiple Boolean Kernels} \label{sec:multi_kernel}

\vspace{-1.3ex}

\subsection{Boolean Reformulation for Linear Layers} \label{sec:bool_linear}

\vspace{-1.1ex}

\begin{wrapfigure}[7]{r}{0.45\textwidth}
    \centering
    \vspace{-3.5ex}

    \includegraphics[width=0.4\textwidth]{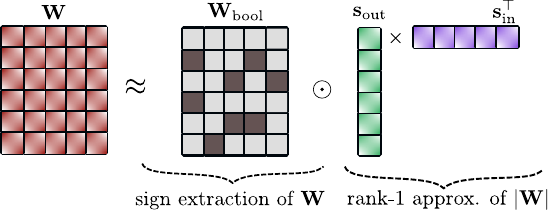}
    
    \vspace{-0.5ex}
    
    \caption{\small Illustration of \textsc{svid}.}
    \label{fig:rank_1_approx}
\end{wrapfigure}

\glspl{LLM} \citep{Brown2020} are mostly based on the Transformer architecture \citep{vaswani2017attention}, in which linear layers are the core elements.
Inpsired by \cite{xu2024onebit}, we employ \gls{SVID} such that an \gls{FP} input matrix $\mbW \in \mathbb{R}^{m\times n} $ of linear layers is decomposed into one Boolean matrix $\mbW_{\mathrm{bool}} \triangleq \mathbf{sign}(\mbW)$ and two \gls{FP} vectors $\mbs_{\mathrm{in}}$ and $\mbs_{\mathrm{out}}$.
Precisely, let $|\mbW|$ be the element-wise absolute value of $\mbW$, write $|\mbW| = \mbU \mbSigma \mbV^{\top}$ its \gls{SVD} \citep{sulle1990}.
Using rank-1 approximation of $|\mbW|$, $\mbs_{\mathrm{in}}$ and $\mbs_{\mathrm{out}}$ are given as: $\mbs_{\mathrm{in}} = \sqrt{\sigma_1} \mbV_{[:,1]}$, and $\mbs_{\mathrm{out}} = \sqrt{\sigma_1} \mbU_{[:,1]}$.
Then, the input matrix is approximated as $\mbW = \mbW_{\mathrm{bool}} \odot |\mbW| \approx \mbW_{\mathrm{bool}} \odot \left( \mbs_{\mathrm{out}} \mbs_{\mathrm{in}}^{\top} \right)$.
This procedure is illustrated in \cref{fig:rank_1_approx}.

\vspace{-0.5ex}

\begin{proposition}\citep{xu2024onebit}
    \label{pro:svid}
    For $\mbW \in \mathbb{R}^{m\times n}$, write $\mbW = \widetilde{\mbU} \widetilde{\mbSigma} \widetilde{\mbV}^{\top}$ its \gls{SVD}.
    Let $\mba = \sqrt{\tilde{\sigma}_1} \widetilde{\mbU}_{[:,1]}$, and $\mbb = \sqrt{\tilde{\sigma}_1} \widetilde{\mbV}_{[:,1]}$.
    With the notations as described above, we have:

    \vspace{-1.5ex}

    \begin{align}
        \left\| \mbW - \mbW_{\mathrm{bool}} \odot \mbs_{\mathrm{out}} \mbs_{\mathrm{in}}^{\top}   \right\|_{F}^{2} \leq \left\| \mbW - \mba \mbb^{\top} \right\|_{F}^{2}.
    \end{align}
\end{proposition}

\vspace{-0.5ex}

\begin{remark}
    \cref{pro:svid} re-states Proposition 2 of \cite{xu2024onebit} with its precise assumption of vectors $\mba$ and $\mbb$ which is necessary for its proof provided in Appendix therein.
\end{remark}

\vspace{-0.5ex}

\cref{pro:svid} shows that using $\mbW_{\mathrm{bool}}$ together with value matrix approximation is better than a direct rank-1 approximation of $\mbW$ in terms of Frobenius-norm. 
This emphasizes the important role of $\mbW_{\mathrm{bool}}$ in approximating the original \gls{FP} matrix.
Moreover, our following \cref{pro:svid_rank_1} shows that the \gls{SVID} approximation as described above is optimal for approximating the original matrix $\mbW_{\mathrm{bool}}$.
\begin{proposition}
    \label{pro:svid_rank_1}
    For $\mbW \in \mathbb{R}^{m\times n}$ and the notations as described above, we have:
    \begin{align}
        \left\| \mbW - \mbW_{\mathrm{bool}} \odot \mbs_{\mathrm{out}} \mbs_{\mathrm{in}}^{\top}   \right\|_{F}^{2} \leq \left\| \mbW - \mbW_{\mathrm{bool}} \odot \mbc \mbd^{\top}   \right\|_{F}^{2}, \quad \forall \mbc \in \mathbb{R}^{m \times 1}, \forall \mbd \in \mathbb{R}^{n \times 1}.
    \end{align}
\end{proposition}

\vspace{-1ex}

The proof is given in \cref{proof:svid_rank_1}.
The linear layer can be then reformulated as \citep{xu2024onebit}:
\begin{align}
    \mbX \mbW_{\mathrm{FP}}^{\top} \approx \left[ \left( \mbX \odot \mbs_{\mathrm{in}}^{\top} \right) \mbW_{\mathrm{bool}} \right] \odot \mbs_{\mathrm{out}}^{\top}.
\end{align}

\vspace{-2.0ex}

\subsection{Enhanced Expressivity with Multiple Boolean Kernels } \label{sec:multi_kernel_boolean}

\vspace{-1ex}

\begin{wrapfigure}[13]{r}{0.48\textwidth}
    \centering
    \vspace{-3.5ex}
    \includegraphics[width=0.42\textwidth]{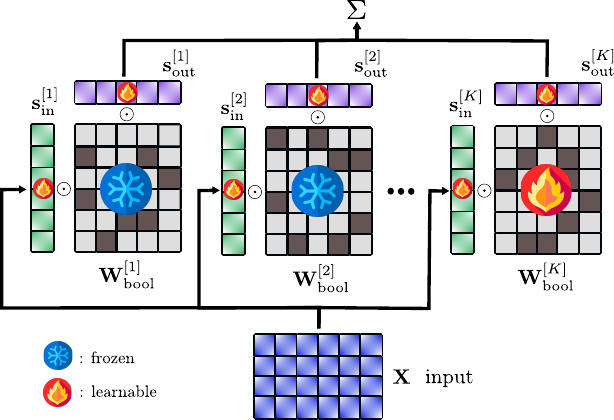}
    
    \vspace{-1ex}
    
    \caption{\small The computation of a linear layer approximated using multi kernels of Boolean.}
    \label{fig:multi_kernel}
\end{wrapfigure}

We have shown that \gls{SVID} provides a good approximation of the original weights, its expressivity can be still limited to capture well the original \gls{FP} parameters of complicated models, which were trained on large-scale datasets over extended periods of time.
To overcome this limitation, we propose the use of a multi-Boolean kernel structure for the weights. Specifically, we employ $K$ kernels, where each kernel utilizes distinct Boolean weights and scaling factors, to better represent the original weight parameters.
This leads to the approximation: $\mbW_{\mathrm{FP}} \approx \mbW_{\mathrm{approx}} \triangleq \sum_{k=1}^{K} \mbW_{\mathrm{bool}}^{[k]} \odot ( \mbs_{\mathrm{out}}^{[k]} {\mbs_{\mathrm{in}}^{[k]}}^\top ) $.
The computation of a linear layer can then be approximated as follows (see \cref{fig:multi_kernel} for an illustration):

\vspace{-1.5ex}
\begin{align}
    \mbX \mbW_{\mathrm{FP}}^{\top} \approx \sum_{k=1}^{K} \left[ \left( \mbX \odot {\mbs_{\mathrm{in}}^{[k]}}^{\top} \right) \mbW_{\mathrm{bool}}^{[k]} \right] \odot {\mbs_{\mathrm{out}}^{[k]}}^{\top}.
\end{align}

\vspace{-1.5ex}

Here, the computational costs associated with the \gls{FP} scaling factors, $\mbs_{\mathrm{in}}$ and $\mbs_{\mathrm{out}}$, are small because they only involve element-wise multiplications. 
The dominant computational cost arises from the matrix multiplication between the scaled input data, $\mbX \odot \mbs_{\mathrm{in}}$, and the weights.
However, thanks to the use of Boolean weights, the complexity is significantly reduced, as these multiplications can be replaced by additions.
Moreover, as we will demonstrate in \cref{sec:exp_num_kernels}, only a small number of kernels are required to achieve a reasonable result. 
Additionally, we find that, after the successive extraction process from the \gls{FP} model (\cref{sec:successive_extraction}), we only need to train the Boolean weights for the last kernel and the scaling factors, further significantly reducing the overall complexity.

\vspace{-1ex}

\subsection{Effective Knowledge Transfer into Boolean Models} \label{sec:knowledge_transfer}

\vspace{-1ex}

We have introduced our proposed multi-Boolean kernel structure for effectively representing the linear layers of \glspl{LLM}.
In this section, we outline the process for transferring knowledge from a source \gls{FP} model to a Boolean model.
This process consists of two steps: (1) data-free initialization to maximize information retention from the source, and (2) data-dependent finetuning, where the Boolean model is further trained on a target dataset with guidance from the \gls{FP} model.

\vspace{-1.3ex}

\subsubsection{Successive Extraction using SVID} \label{sec:successive_extraction}

\vspace{-0.8ex}

For each linear layer, to initialize the values of the Boolean weights and scaling factors for all kernels, we successively apply \gls{SVID} to the given \gls{FP} weights. 
The goal here is to further proceed to \gls{SVID} process to approximate the residual error introduced by the previous step. %
Specifically, after each step of decomposing the weight matrix using \gls{SVID}, we obtain a residual matrix, which is defined as:

\vspace{-1.8ex}
\begin{align}
    \mbW_{\mathrm{res}}^{[k]} = \mbW_{\mathrm{input}}^{[k]} - \mbW_{\mathrm{bool}}^{[k]} \odot \left( \mbs_{\mathrm{out}}^{[k]} {\mbs_{\mathrm{in}}^{[k]}}^\top \right). \label{eq:Wres}
\end{align}

\vspace{-2.3ex}

Here, $\mbW_{\mathrm{res}}^{[k]}$ is the residual matrix, and $\mbW_{\mathrm{bool}}^{[k]}$,  $\mbs_{\mathrm{out}}^{[k]}$ and $\mbs_{\mathrm{in}}^{[k]}$ are the extracted parameters for the $k$-th kernel, while $\mbW_{\mathrm{input}}^{[k]}$ represents the input \gls{FP} matrix for step $k$.
For the first step, this is the original weight matrix, and for subsequent steps, it is the residual matrix obtained from the previous step.

\begin{figure}[H]
    \centering
    \includegraphics[width=0.95\textwidth]{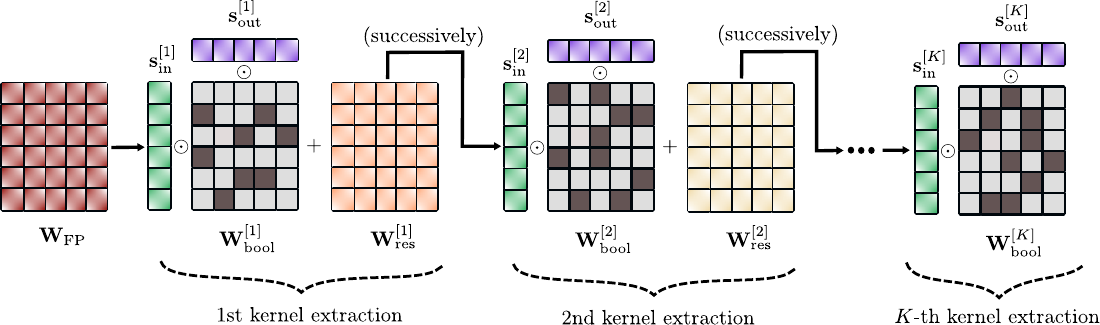}
    \vspace{-1ex}
    \caption{Illustration of successive extractions of Boolean kernels from a given \gls{FP} weight matrix. \label{fig:overview}}
\end{figure}

\cref{fig:overview} illustrate this process.
Although using multiple kernels effectively captures the original weight matrix, a residual error still remains at the end of the process.
While this residual error is small, it can accumulate as it propagates through the layers, finally leading to predictions that diverge from those of the original \gls{FP} model.
To address this issue, it is necessary to further finetune the resulting model to compensate for these errors and make it better suited to the target task.
We will discuss this in \cref{sec:optimization_strategy}.
In the following section, we will introduce knowledge distillation to achieve this goal.

\vspace{-2ex}

\vspace{-2.5ex}

\subsubsection{Finetuning with Knowledge Distillation} \label{sec:knowledge_distillation}

Knowledge distillation (\textsc{kd}) \citep{hinton2015kd} trains a student network to mimic a more powerful teacher, usually with greater efficiency.
The student learns from the teacher’s output distribution and/or intermediate states as ``soft targets''.
Here, the \gls{FP} model is the teacher and the Boolean model the student.
Specifically, the output probability distribution of an \gls{LLM} for a token $\mbX_{[i]}$ is:

\vspace{-2.3ex}
\begin{align}
    p(\mbX_{[i]}; \tau) = \frac{\exp (\mbX_{[i]}/ \tau)}{\sum_{j=1}^{N_V} \exp (\mbX_{[j]} / \tau)},
\end{align}

\vspace{-1.8ex}

where $N_V$ is the vocabulary size and $\tau$ is the softmax temperature.
The logit-based \gls{KD} loss across the sequence of all output tokens is defined as follows:

\vspace{-2.6ex}
\begin{align}
    \cL_{\mathrm{logits}} = \frac{1}{L} \sum_{j=1}^{L} \mathrm{D}_{\textrm{logits}} \left(p_{\mathrm{FP}} (\mbX_{[j]}; \tau), p_{\mathrm{bool}} (\mbX_{[j]}; \tau) \right). \label{eq:loss_logits}
\end{align}
\vspace{-3.5ex}

Here, $p_{\mathrm{FP}} (\mbX_{[j]}; \tau)$ and $p_{\mathrm{bool}} (\mbX_{[j]}; \tau)$ denote the distributions over the $j$-th token from the \gls{FP} and Boolean models, respectively, with $L$ as the sequence length.
We find that $\tau=1$ works best in practice. Among possible measures for $\mathrm{D}_{\textrm{logits}}$ \citep{ko2024distillm}, the forward \gls{KL} divergence gives the strongest results; further discussion is in \cref{sec:kd_loss_choice}.

To further reduce distributional discrepancies in intermediate layers, we additionally employ an intermediate state–based \gls{KD} loss over a sequence of hidden states:

\vspace{-2.1ex}
\begin{align}
    \cL_{\mathrm{is}} = \frac{1}{L} \sum_{h \in H} \sum_{j=1}^{L} \left\|  \mbQ_{\mathrm{FP}}^{j,h}  - \mbQ_{\mathrm{bool}}^{j,h} \right\|_2^2,
\end{align}

\vspace{-1.8ex}

where $\mbQ_{\mathrm{FP}}^{j,h}$ and $\mbQ_{\mathrm{bool}}^{j,h}$ represent the $h$-th hidden states of the \gls{FP} and Boolean models for the $j$-th token, repsectively; $H$ is the set of chosen intermediate states.
Finally, the overall loss is then computed as $\cL = \cL_{\mathrm{logits}} + \gamma \cL_{\mathrm{is}}$,
where $\gamma$ is a weighted factor that balances the contribution of the two losses.
We empirically found that $\gamma=10$ works best.

\section{Kernel Allocation} \label{sec:kernel_allocation}

Using more kernels enhances the Boolean model’s representational capacity but also increases its size.
We propose a method to automatically allocate kernels per weight under a fixed budget.
Let $N_{\mbW}$ be the number of weights in the \gls{FP} teacher model, and $K_l$ for \( l \in [1, N_{\mbW}] \) the number of Boolean kernels for the $l$-th weight.
Our goal is to determine $\mbk \triangleq \{K_l\}_{l \in [1, N_{\mbW}]}$ subject to design constraints. Key factors include:

\vspace{-0.8ex}

\textit{(1) Residual error}: Let $e_l^{[k]} \in \bbR$ denote the approximation error from applying the successive \gls{SVID} extraction to the $k$-th kernel of the $l$-th weight, measured by the Frobenius norm of $\mbW_{\mathrm{res}}^{[k]}$ (\cref{eq:Wres}).

\vspace{-0.8ex}

\textit{(2) Weight importance}: Let $h_l$ denote the importance of the $l$-th weight in the \gls{FP} teacher model.
Higher scores indicate the need for more Boolean kernels.
We propose estimating $h_l$ using \gls{PWCCA} \citep{Morcos2018}, a reliable method for analyzing deep model representations. Details are provided in \cref{sub:weight_importance}.

\vspace{-0.8ex}

\textit{(3) Weight size}: The size of the $l$-th weight is denoted by $s_l$ and $p_l \triangleq s_l / \sum_{k=1}^{N_{\mbW}} s_k$ represents its relative size in the model.

For a given $\mbk$, the size of the target Boolean model, in terms of the number of weights, is $\sum_{l=1}^{N_{\mbW}} K_l s_l$.
Relative to the source \gls{FP} model, this repersents an expansion ratio, defined as:

\vspace{-1ex}
\begin{equation}
    \rho(\mbk) \triangleq \frac{\sum_{l=1}^{N_{\mbW}} K_l s_l}{\sum_{l=1}^{N_{\mbW}} s_l} = \sum_{l=1}^{N_{\mbW}} K_l p_l.
\end{equation}

\vspace{-2ex}

\paragraph{Optimization objective.}
To control model size, we constrain the expansion ratio to a target $T \geq 1$ and limit the kernel size by $K_{\max}$, with $T \leq K_{\max} \leq \infty$.
The optimization space is thus $\cK \triangleq [1, K_{\max}]^{N_{\mbW}}$, and the problem is formulated as:

\vspace{-2ex}
\begin{align}
    \mbk^{\ast} = \argmin_{\mbk \in \cK} \cE(\mbk), \quad \textrm{s.t.} \quad \rho(\mbk) \le T, \quad \text{where } \cE(\mbk) \triangleq \sum_{l=1}^{N_{\mbW}} h_l  e_l^{[K_l]} f(p_l).
\end{align}

\vspace{-2ex}

Here, $\cE(\mbk)$ is the objective (energy) function, and $f(\cdot)$ is a monotonically decreasing function. In practice, we use $f(p_l) = (1/p_l)\log(1/p_l)$.
Intuitively, the goal is to minimize residual error while prioritizing weights with higher importance and smaller size, balancing accurate knowledge transfer with model efficiency.

\vspace{-2ex}

\paragraph{Optimization algorithm.}

The problem has complexity $\mathcal{O}(K_{\max}^{N_{\mbW}})$, which is prohibitive for \glspl{LLM}.
To tackle this NP-hard problem efficiently, we note that $e_l^{[k]}$ decreases with $k$ for all $l$, and $\cE(\mbk)$ is maximized at $\mbk = \mathbf{1}$, with any increase in $k_l$ reducing $\cE(\mbk)$.
This motivates a heuristic iterative approach: at each step, increment the $K_l$ that yields the largest reduction in $\cE(\mbk)$.
The full algorithm is given in \Cref{algo:KernelAllocation} in the Appendix.
We will demonstrate in \cref{sec:exp_kernel_alloc} the practicality of our method.

\vspace{-1ex}

\section{Experiments} \label{sec:experiment}

\vspace{-1.3ex}

\paragraph{Setups.}
In all experiments, we follow the protocol from \cite{jo2024mixture}, without quantizing activations.
The training set combines WikiText2 \citep{merity2017pointer} and a selected partition of C4 \citep{raffel2020c4} data, using sequences of length 2048.
We apply a cosine decay learning rate with a 3\% warm-up over 3 epochs and batch size 8.
Boolean parameters use a maximum learning rate of $5\times10^{-3}$, while remaining \gls{FP} parameters are optimized with AdamW \citep{loshchilov2018decoupled} at a maximum learning rate of $2\times10^{-5}$, with $\beta_1=0.9$ and $\beta_2=0.999$.
Following standard practice \citep{jo2024mixture}, performance is evaluated via perplexity on WikiText2 and C4 (lower is better).

\subsection{Ablation Studies and Analysis}

\subsubsection{Effect of the number of kernels} \label{sec:exp_num_kernels}

\vspace{-3.5ex}

\begin{figure}[H]
    \centering
    \begin{subfigure}[c]{0.6\textwidth}

        \tikzexternaldisable
        \centering
        \scriptsize
        \setlength{\figurewidth}{4.3cm}
        \setlength{\figureheight}{3.5cm}
        \input{figures/approx_errors_init.tikz}
        \tikzexternalenable

    \end{subfigure}
    \begin{subfigure}[c]{0.33\textwidth}
        \caption*{{\scriptsize{Perplexity ($\downarrow$)}}}
        \vspace{-0.5ex}
        \scalebox{.7}{
    \setlength{\tabcolsep}{3.9pt}
    \renewcommand{\arraystretch}{1.2}
    \begin{tabular}{rcccc}
        \toprule
        \#Kernels & 1     & 2     & 3     & 4     \\
        \midrule
        Wiki2     & 39.38 & 31.47 & 29.10 & 28.52 \\
        C4        & 35.44 & 28.62 & 26.48 & 25.90 \\
        \midrule
        \midrule
        \#Kernels & 5     & 6     & 7     & 8     \\
        \midrule
        Wiki2     & 28.19 & 28.16 & 28.13 & 28.08 \\
        C4        & 25.65 & 25.50 & 25.40 & 25.31 \\
        \bottomrule
    \end{tabular}
}
    
    \end{subfigure}

    \vspace{-1.2ex}

    \caption{Normalized L1 norm difference between the approximated weights at initialization and after finetuning against the \gls{FP} weights ($\|\mbW_{\text{approx}} - \mbW_{\text{FP}}\|_{1} / \|\mbW_{\text{FP}}\|_{1}$) \label{fig:approx_erros_init}, and the final results.}
\end{figure}

\vspace{-3.5ex}

We begin by examining the effect of the number of Boolean kernels on \optsmall model \citep{zhang2022opt}.
\cref{fig:approx_erros_init} shows the normalized difference between weights approximated via successive \gls{SVID} and the original \gls{FP} weights, both at initialization and after finetuning.
Increasing the number of kernels reduces approximation error and improves perplexity, unlike \mos \citep{jo2024mixture}, where adding more experts does not always help and can even hurt performance.
Using 3–4 kernels yields a good approximation, with diminishing improvements beyond that.
Interestingly, the normalized difference relative to the full \gls{FP} weights is larger after \gls{KD} finetuning.
We hypothesize that \gls{KD} compensates the errors due to the lower expressiveness of a small number of kernels, further emphasizing its role in adapting the model to approximate the \gls{FP} model rather than exactly replicating each weight.

\subsubsection{Optimization strategy} \label{sec:optimization_strategy}

\vspace{-1.5ex}

\begin{figure}[H]
    \begin{subfigure}[c]{0.65\textwidth}
        \centering
        \tikzexternaldisable
        \centering
        \scriptsize
        \setlength{\figurewidth}{4.3cm}
        \setlength{\figureheight}{3.2cm}
        \input{figures/optimization_strategy.tikz}
        \tikzexternalenable
    \end{subfigure}
    \begin{subfigure}[c]{0.33\textwidth}
        \centering
        \tikzexternaldisable

\scalebox{.7}{
    \setlength{\tabcolsep}{3.9pt}
    \renewcommand{\arraystretch}{1.2}
    \begin{tabular}{lcc}
        \toprule
        Optim.                                                                                                                                                              & \multirow{2}{*}{Wiki2} & \multirow{2}{*}{C4} \\
        Kernel                                                                                                                                                              &                        &                     \\
        \midrule
        {\protect\tikz[baseline=-1ex]\protect\draw[color=color_blue, fill=color_blue, opacity=0.99, mark size=1.7pt, line width=1.7pt] plot[] (-0.0,0)--(-0.45,0);} 1st     & 33.90                  & 30.70               \\
        {\protect\tikz[baseline=-1ex]\protect\draw[color=color_orange, fill=color_orange, opacity=0.99, mark size=1.7pt, line width=1.7pt] plot[] (-0.0,0)--(-0.45,0);} 2nd & 30.29                  & 27.55               \\
        {\protect\tikz[baseline=-1ex]\protect\draw[color=color_green, fill=color_green, opacity=0.99, mark size=1.7pt, line width=1.7pt] plot[] (-0.0,0)--(-0.45,0);} 3rd   & 29.00                  & 26.36               \\
        {\protect\tikz[baseline=-1ex]\protect\draw[color=color_red, fill=color_red, opacity=0.99, mark size=1.7pt, line width=1.7pt] plot[] (-0.0,0)--(-0.45,0);} 4th       & \textbf{28.60}         & \textbf{25.93}      \\
        {\protect\tikz[baseline=-1ex]\protect\draw[color=black, fill=black, opacity=0.99, mark size=1.7pt, line width=1.7pt] plot[] (-0.0,0)--(-0.45,0);} All               & 32.04                  & 29.08               \\
        \bottomrule
    \end{tabular}
}

\tikzexternalenable
    
    \end{subfigure}

    \vspace{-1.1ex}
    
    \caption{
        The progression of training losses, number of flips, and perplexity of the resulting models (\optsmall) is examined with respect to the optimization of different kernel configurations.
    \label{fig:optim_strategy}}
\end{figure}

\vspace{-1.5ex}

Next, we study the effect of optimizing kernels on the \optsmall model. 
We consider four Boolean kernels but train only one at a time, keeping the others frozen.
\cref{fig:optim_strategy} shows the loss convergence.
Training the first kernel converges slowest, while higher-order kernels improve progressively.
As shown in \cref{pro:svid} and \cref{pro:svid_rank_1}, the \gls{SVID} effectively extracts optimal Boolean weights and scaling factors.
In our successive \gls{SVID} framework, the first kernel is well extracted and captures the most important information, and higher-order kernels approximate residuals.
Since the kernels are related in a successive manner, modifying lower-order kernels affects higher-order ones.
We observe that training only the first kernel results in many weight flips, indicating optimization difficulty, whereas fine-tuning only the last kernel efficiently compensates for residual errors, showing the lowest flip rates and best performance.
This is in line with the observation by \cite{liu2024bitdelta}, where they compress ``delta'' induced by the finetuning process by using 1-bit weights.
This further highlights the advantage of our approach, as training complexity is significantly reduced by only optimizing the last kernel.
Thus, we apply this strategy in all our experiments.

\vspace{-1.5ex}

\subsection{Main Benchmark Results}

\vspace{-1.5ex}

\cref{tab:benchmark_main} compares our method with recent baselines in binarization and 2-bit quantization, evaluating perplexity and accuracy on zero-shot tasks including Winogrande \citep{sakaguchi2021winogrande}, HellaSwag \citep{zellers2019hellaswag}, PIQA \citep{bisk2020piqa}, BoolQ \citep{clark2019boolq}, and ARC \citep{clark2018arc}. 
For our method, we use 2 Boolean kernels, an ultra low-bit setting. 
Due to space constraints, the results for \llamatwosmall and \llamatwobig \citep{touvron2023llama2} and different number of Boolean kernels are provided in \cref{sec:bencmark_llama2} and \cref{sec:benchmark_more_kernels}.
We note that our method is close to scalar quantization while being completely orthogonal to vector quantization (VQ) which adds substantial overhead \citep{gray1984vector}.
For completeness, we encourage the reader refer to \cref{sec:latency_vq} for VQ comparisons, and \cref{sec:addtional_baselines} for further baselines.

Our method consistently and significantly outperforms the baselines in both perplexity and zero-shot accuracy, achieving results close to the \halffp baseline despite using only a budget of 2 bits for weight.
As expected, \gls{QAT} methods like \onebit and \mos perform better than \gls{PTQ} methods, but this comes at the cost of extensive finetuning. 
In contrast, our approach efficiently address this problem by optimizing parameters directly in Boolean space, avoiding the need for optimizing in \gls{FP} latent sapce.

\vspace{-1.6ex}

\begin{table}[H]
    \setlength{\tabcolsep}{3.8pt}
      \centering
      \caption{Perplexity and zero-shot accuracy results of Float16, quantized and binarized \glspl{LLM}.}
      \label{tab:benchmark_main}

      \vspace{-1.5ex}

      \renewcommand{\arraystretch}{1.00}
      \scalebox{.81}{
        \begin{tabular}{llcccccccccc}
        \toprule
        \multirow{2}{*}{\textbf{Model}} & \multirow{2}{*}{\textbf{Method}} & \multirow{2}{*}{\textbf{Wbits}} & \multicolumn{2}{c}{\textbf{Perplexity ($\downarrow$})} & \multicolumn{7}{c}{\textbf{Zero-shot Accuracy ($\uparrow$})} \\
          &&& \textbf{Wiki2} & \textbf{C4} & BoolQ & PIQA & Hella. & WinoG. & ARC-e & ARC-c & \textbf{Average} \\ 
        \midrule
        \midrule

        \multicolumn{1}{l}{\multirow{9}{*}{\optbig}}  & \halffp  & \small{16} & \small{14.62} & \small{14.72} & \small{57.82} & \small{72.42} & \small{53.70} & \small{59.51} & \small{50.97} & \small{29.52} & \small{53.99} \\ 
        \cmidrule(l){2-12} 
        \multicolumn{1}{c}{} & \pbllm  & \small{1.7} & \small{272.83} & \small{175.42} & \small{62.17} & \small{54.24} & \small{27.25} & \small{50.27} & \small{27.98} & \small{23.72} & \small{40.94} \\
        \multicolumn{1}{c}{} & \billm  & \small{1.11} & \small{69.45} & \small{63.92} & \small{61.92} & \small{59.52} & \small{33.81} & \small{49.32} & \small{34.38} & \small{22.35} & \small{43.55} \\
        \multicolumn{1}{c}{} & \onebit  & \small{1} & \small{20.36} & \small{20.76} & \small{57.85} & \small{66.53} & \small{39.21} & \small{54.61} & \small{42.80} & \small{23.97} & \small{47.50} \\
        \multicolumn{1}{c}{} & \mos  & \small{1} & \small{18.45} & \small{18.83} & \small{60.34} & \small{68.66} & \small{41.99} & \small{53.99} & \small{44.87} & \small{26.19} & \small{49.34} \\ 
        \cmidrule(l){2-12}
        \multicolumn{1}{c}{} & \qptq  & \small{2} & \small{9.5e3} & \small{3.8e3} & \small{39.60} & \small{52.07} & \small{25.57} & \small{49.33} & \small{26.68} & \small{23.63} & \small{35.15} \\
        \multicolumn{1}{c}{} & \llmqat  & \small{2} & \small{4.9e3} & \small{2.1e3} & \small{37.83} & \small{50.05} & \small{25.72} & \small{49.72} & \small{25.76} & \small{25.09} & \small{34.07} \\
        \multicolumn{1}{c}{} & \omniquant & \small{2} & \small{42.43} & \small{55.64} & \small{56.45} & \small{60.94} & \small{33.39} & \small{51.85} & \small{38.76} & \small{23.38} & \small{44.13} \\
        \cmidrule(l){2-12}
        \multicolumn{1}{c}{} & \ours [Ours] & \small{2$\times$1} & \cellcolor[HTML]{d6e9c9}\small{\textbf{16.13}} & \cellcolor[HTML]{d6e9c9}\small{\textbf{16.61}} & \cellcolor[HTML]{d6e9c9}\small{58.53} & \cellcolor[HTML]{d6e9c9}\small{70.67} & \cellcolor[HTML]{d6e9c9}\small{48.11} & \cellcolor[HTML]{d6e9c9}\small{56.75} & \cellcolor[HTML]{d6e9c9}\small{48.19} & \cellcolor[HTML]{d6e9c9}\small{27.90} & \cellcolor[HTML]{d6e9c9}\small{\textbf{51.69}} \\

        \midrule
        \midrule 
        
        \multicolumn{1}{l}{\multirow{9}{*}{\llamasmall }}  & \halffp & \small{16} & \small{5.68} & \small{7.08} & \small{73.21} & \small{77.42} & \small{72.99} & \small{66.85} & \small{52.53} & \small{41.38} & \small{64.06} \\ 
        \cmidrule(l){2-12} 
        \multicolumn{1}{c}{} & \pbllm  & \small{1.7} & \small{198.37} & \small{157.35} & \small{60.51} & \small{53.53} & \small{27.23} & \small{49.17} & \small{27.48} & \small{26.02} & \small{40.66} \\
        \multicolumn{1}{c}{} & \billm  & \small{1.11} & \small{41.66} & \small{48.15} & \small{62.23} & \small{58.65} & \small{34.64} & \small{51.14} & \small{33.08} & \small{25.68} & \small{44.24} \\
        \multicolumn{1}{c}{} & \onebit & \small{1} & \small{8.48} & \small{10.49} & \small{62.50} & \small{70.40} & \small{54.03} & \small{55.32} & \small{41.07} & \small{30.88} & \small{52.36} \\
        \multicolumn{1}{c}{} & \mos  & \small{1} & \small{7.97} & \small{9.72} & \small{64.59} & \small{71.82} & \small{58.18} & \small{58.88} & \small{42.09} & \small{31.31} & \small{54.48} \\ 
        \cmidrule(l){2-12}
        \multicolumn{1}{c}{} & \qptq  & \small{2} & \small{1.9e3} & \small{7.8e2} & \small{43.79} & \small{49.95} & \small{25.63} & \small{49.41} & \small{25.84} & \small{27.47} & \small{37.02} \\
        \multicolumn{1}{c}{} & \llmqat  & \small{2} & \small{7.1e2} & \small{3.0e2} & \small{37.83} & \small{50.87} & \small{24.76} & \small{51.78} & \small{26.26} & \small{25.51} & \small{36.17} \\
        \multicolumn{1}{c}{} & \omniquant & \small{2} & \small{15.34} & \small{26.21} & \small{58.69} & \small{62.79} & \small{43.68} & \small{52.96} & \small{41.54} & \small{29.35} & \small{48.17} \\
        \cmidrule(l){2-12}
        \multicolumn{1}{c}{} & \ours [Ours] & \small{2$\times$1} & \cellcolor[HTML]{d6e9c9}\small{\textbf{6.83}} & \cellcolor[HTML]{d6e9c9}\small{\textbf{8.53}} & \cellcolor[HTML]{d6e9c9}\small{69.20} & \cellcolor[HTML]{d6e9c9}\small{74.32} & \cellcolor[HTML]{d6e9c9}\small{64.80} & \cellcolor[HTML]{d6e9c9}\small{60.30} & \cellcolor[HTML]{d6e9c9}\small{49.05} & \cellcolor[HTML]{d6e9c9}\small{34.90} & \cellcolor[HTML]{d6e9c9}\small{\textbf{58.76}} \\

        \midrule
        \midrule
      
        \multicolumn{1}{l}{\multirow{9}{*}{\llamabig }}  & \halffp & \small{16} & \small{5.09} & \small{6.61} & \small{68.47} & \small{79.05} & \small{76.24} & \small{70.17} & \small{59.85} & \small{44.54} & \small{66.39} \\ 
        \cmidrule(l){2-12} 
        \multicolumn{1}{c}{} & \pbllm & \small{1.7} & \small{35.83} & \small{39.79} & \small{62.17} & \small{58.70} & \small{33.97} & \small{52.17} & \small{31.86} & \small{23.63} & \small{43.75} \\
        \multicolumn{1}{c}{} & \billm & \small{1.11} & \small{14.56} & \small{16.67} & \small{62.53} & \small{68.17} & \small{52.24} & \small{59.43} & \small{41.91} & \small{29.94} & \small{52.37} \\
        \multicolumn{1}{c}{} & \onebit & \small{1} & \small{7.65} & \small{9.56} & \small{63.30} & \small{71.98} & \small{60.61} & \small{59.43} & \small{42.85} & \small{32.42} & \small{55.10} \\
        \multicolumn{1}{c}{} & \mos & \small{1} & \small{7.16} & \small{8.81} & \small{63.82} & \small{73.88} & \small{64.05} & \small{60.93} & \small{44.28} & \small{33.11} & \small{56.68} \\ 
        \cmidrule(l){2-12}
        \multicolumn{1}{c}{} & \qptq & \small{2} & \small{3.2e3} & \small{9.9e2}  & \small{42.39} & \small{50.00} & \small{25.27} & \small{50.67} & \small{26.14} & \small{27.39} & \small{36.98} \\
        \multicolumn{1}{c}{} & \llmqat & \small{2} & \small{1.8e3} & \small{1.2e3} & \small{37.83} & \small{50.33} & \small{25.40} & \small{51.62} & \small{27.02} & \small{26.87} & \small{36.51} \\
        \multicolumn{1}{c}{} & \omniquant & \small{2} & \small{13.43} & \small{19.33} & \small{62.20} & \small{68.99} & \small{54.16} & \small{53.83} & \small{45.50} & \small{30.38} & \small{52.51} \\
        \cmidrule(l){2-12}
        \multicolumn{1}{c}{} & \ours [Ours] & \small{2$\times$1} & \cellcolor[HTML]{d6e9c9} \textbf{\small{6.17}} & \cellcolor[HTML]{d6e9c9} \textbf{\small{7.88}} & \cellcolor[HTML]{d6e9c9}\small{68.10} & \cellcolor[HTML]{d6e9c9}\small{76.33} & \cellcolor[HTML]{d6e9c9}\small{69.88} & \cellcolor[HTML]{d6e9c9}\small{64.17} & \cellcolor[HTML]{d6e9c9}\small{52.34} & \cellcolor[HTML]{d6e9c9}\small{37.88} & \cellcolor[HTML]{d6e9c9}\textbf{\small{61.45}} \\

        \bottomrule
      \end{tabular}}
      \vspace{-3ex}
    \end{table}

\vspace{-0.8ex}

\subsection{Accuracy-Compression Trade-offs}

\vspace{-0.5ex}

We further investigate the accuracy-compression trade-offs of our method, quantization methods, and the \gls{FP} model. 
Specifically, we compare 3-bit quantization using round-to-nearest (\textsc{rtn}) \citep{yao2022zeroquant,dettmers2022gptint} and \qptq \citep{frantar2023optq} methods against our approach using 3 Boolean kernels. 
We evaluate these methods on \textsc{opt} models of varying sizes. 
The results, presented in \cref{tab:opt_scaling} and \cref{fig:opt_scaling}, show that with 3 kernels, our method closely approaches the performance of the \gls{FP} model. 
Given the same weight budget, our method clearly sits on the Pareto frontier, delivering the best performance for the same model size.

\vspace{-2ex}

\begin{table}[H]
    \centering
    \caption{ \textsc{opt} perplexity results (\textit{lower is better}) on WikiText2 and C4.
        The results of \gls{FP}, rount-to-nearest (\textsc{rtn}) and \textsc{optq} are taken from \citep{frantar2023optq}.
    }
    \label{tab:opt_scaling}

    \vspace{-1.6ex}

    \renewcommand{\arraystretch}{1.01}

    \resizebox{.99\columnwidth}{!}{
        \begin{tabular}{lc|ccccc|ccccc}
            \toprule
            \multirow{2}{*}{\textbf{OPT Model}}                        & \multicolumn{1}{c}{\multirow{2}{*}{\textbf{WBits}}} & \multicolumn{5}{c|}{\textbf{Wiki2}}    & \multicolumn{5}{c}{\textbf{C4}}                                                                                                                                                                                                                                                                                                                                                \\ %
                                                                       & \multicolumn{1}{c|}{}                               & \multicolumn{1}{c}{125M}               & \multicolumn{1}{c}{350M}               & \multicolumn{1}{c}{1.3B}               & \multicolumn{1}{c}{2.7B}               & \multicolumn{1}{c|}{6.7B}              & \multicolumn{1}{c}{125M}               & \multicolumn{1}{c}{350M}               & \multicolumn{1}{c}{1.3B}               & \multicolumn{1}{c}{2.7B}               & \multicolumn{1}{c}{6.7B}               \\ \midrule

            \textsc{full-precision}                                    & 16                                                  & 27.65                                  & 22.00                                  & 14.63                                  & 12.47                                  & 10.86                                  & 26.56                                  & 22.59                                  & 16.07                                  & 14.34                                  & 12.71                                  \\ \midrule
            \textsc{rtn}   \citep{yao2022zeroquant,dettmers2022gptint} & 3                                                   & 1.3e3                                  & 64.57                                  & 1.3e4                                  & 1.6e4                                  & 5.8e3                                  & 834                                  & 55.49                                  & 5.2e3                                  & 1.1e4                                 & 5.3e3                                  \\
            \textsc{optq} \citep{frantar2023optq}                      & 3                                                   & 53.85                                  & 33.79                                  & 20.97                                  & 16.88                                  & 14.86                                  & 42.41                                  & 31.33                                  & 21.63                                  & 18.17                                  & 17.14                                  \\ \midrule
            \ours [Ours]                                               & 3$\times$1                                          & \cellcolor[HTML]{d6e9c9}\textbf{29.10} & \cellcolor[HTML]{d6e9c9}\textbf{23.12} & \cellcolor[HTML]{d6e9c9}\textbf{15.30} & \cellcolor[HTML]{d6e9c9}\textbf{13.09} & \cellcolor[HTML]{d6e9c9}\textbf{11.03} & \cellcolor[HTML]{d6e9c9}\textbf{28.62} & \cellcolor[HTML]{d6e9c9}\textbf{22.10} & \cellcolor[HTML]{d6e9c9}\textbf{15.68} & \cellcolor[HTML]{d6e9c9}\textbf{14.00} & \cellcolor[HTML]{d6e9c9}\textbf{12.33} \\
            \bottomrule
        \end{tabular}
    }

    \renewcommand{\arraystretch}{1}
    \vspace{-1.1ex}
\end{table}

\subsection{Comparison with Latent-weight Approaches}

We compare our method with latent-weight approaches on \textsc{opt} models, using \mos with 3 experts and our method with 3 Boolean kernels.
We also introduce a baseline using our \gls{SVID} framework to construct 3 binary weights that rely on \gls{FP} latent weights for training.
Results in \cref{fig:latent_weights} show that our method converges much faster, as it directly optimizes Boolean parameters without the need for \gls{STE} to approximate gradient signals.
Both our approach and the latent-weight method outperform \mos, demonstrating the benefit of using additional Boolean kernels and our successive \gls{SVID} framework. 
Our method is also more efficient, avoiding the need for \gls{FP} latent weights and extra momentum.

\vspace{-1.9ex}

\begin{figure}[H]
    \centering
    \begin{subfigure}[c]{0.58\textwidth}
        \tikzexternaldisable
        \centering
        \scriptsize
        \setlength{\figurewidth}{4.6cm}
        \setlength{\figureheight}{3.15cm}
        \input{figures/latent_weights.tikz}
        \tikzexternalenable
    \end{subfigure}
    \begin{subfigure}[c]{0.4\textwidth}
        \tikzexternaldisable

\scalebox{.68}{
    \setlength{\tabcolsep}{3.9pt}
    \renewcommand{\arraystretch}{1.1}
    \begin{tabular}{llcc}
        \toprule
        OPT                   & Method                                                                                                                                                                                            & Wiki2          & C4             \\
        \midrule
        \midrule
        \multirow{3}{*}{125M} & {\protect\tikz[baseline=-1ex]\protect\draw[color=color_blue, fill=color_blue, opacity=0.99, mark size=1.7pt, line width=1.7pt] plot[] (-0.0,0)--(-0.4,0);} \mos (3 experts) & 38.62          & 34.72          \\
                              & {\protect\tikz[baseline=-1ex]\protect\draw[color=color_orange, fill=color_orange, opacity=0.99, mark size=1.7pt, line width=1.7pt] plot[] (-0.0,0)--(-0.4,0);} 3 Latent weights                   & 29.47          & 27.18          \\
                              & {\protect\tikz[baseline=-1ex]\protect\draw[color=color_green, fill=color_green, opacity=0.99, mark size=1.7pt, line width=1.7pt] plot[] (-0.0,0)--(-0.4,0);} \ours (3 kernels) [Ours]             & \cellcolor[HTML]{d6e9c9}\textbf{29.10} & \cellcolor[HTML]{d6e9c9}\textbf{26.48} \\
        \midrule
        \multirow{3}{*}{350M} & {\protect\tikz[baseline=-1ex]\protect\draw[color=color_blue, fill=color_blue, opacity=0.99, mark size=1.7pt, line width=1.7pt] plot[] (-0.0,0)--(-0.4,0);} \mos (3 experts)  & 29.93          & 28.25          \\
                              & {\protect\tikz[baseline=-1ex]\protect\draw[color=color_orange, fill=color_orange, opacity=0.99, mark size=1.7pt, line width=1.7pt] plot[] (-0.0,0)--(-0.4,0);}  3 Latent weights                  & 23.58          & 22.65          \\
                              & {\protect\tikz[baseline=-1ex]\protect\draw[color=color_green, fill=color_green, opacity=0.99, mark size=1.7pt, line width=1.7pt] plot[] (-0.0,0)--(-0.4,0);} \ours (3 kernels) [Ours]             & \cellcolor[HTML]{d6e9c9}\textbf{23.12} & \cellcolor[HTML]{d6e9c9}\textbf{22.10} \\
        \bottomrule
    \end{tabular}
}

\tikzexternalenable

    \end{subfigure}

    \vspace{-3.2ex}

    \caption{\small Comparions between our method and latent-weight approaches.}
    \label{fig:latent_weights}

    \vspace{-0.5ex}
\end{figure}

\vspace{-1.1ex}

\subsection{Kernel Allocation and Comparison to BitNet b1.58} \label{sec:exp_kernel_alloc}

\begin{wrapfigure}[10]{r}{0.3\textwidth}
    \centering
        
        \vspace{-3ex}

        \begin{subfigure}[c]{0.17\textwidth}
        \tikzexternaldisable
        \centering
        \scriptsize
        \setlength{\figurewidth}{3.8cm}
        \setlength{\figureheight}{3.3cm}
        \begin{tikzpicture}

\definecolor{crimson2143940}{RGB}{214,39,40}
\definecolor{darkgray176}{RGB}{176,176,176}
\definecolor{darkorange25512714}{RGB}{255,127,14}
\definecolor{forestgreen4416044}{RGB}{44,160,44}
\definecolor{mediumpurple148103189}{RGB}{148,103,189}
\definecolor{sienna1408675}{RGB}{140,86,75}
\definecolor{steelblue31119180}{RGB}{31,119,180}

\begin{axis}[
height=\figureheight,
major tick length=1ex,
tick align=outside,
tick pos=left,
width=\figurewidth,
x grid style={darkgray176},
xlabel={Transfomer Block},
xmin=0.0099999999999999, xmax=12.99,
xtick style={color=black},
xtick={5,10},
xticklabels={
  \(\displaystyle {5}\),
  \(\displaystyle {10}\),
},
ytick={0,5,10,15,20},
yticklabels={
  \(\displaystyle {0}\),
  \(\displaystyle {5}\),
  \(\displaystyle {10}\),
  \(\displaystyle {15}\),
  \(\displaystyle {20}\),
},
y grid style={darkgray176},
ylabel={\#Kernels},
ymin=0, ymax=22.05,
ytick style={color=black}
]
\draw[draw=none,fill=steelblue31119180] (axis cs:0.6,0) rectangle (axis cs:1.4,5);

\draw[draw=none,fill=steelblue31119180] (axis cs:1.6,0) rectangle (axis cs:2.4,5);
\draw[draw=none,fill=steelblue31119180] (axis cs:2.6,0) rectangle (axis cs:3.4,5);
\draw[draw=none,fill=steelblue31119180] (axis cs:3.6,0) rectangle (axis cs:4.4,5);
\draw[draw=none,fill=steelblue31119180] (axis cs:4.6,0) rectangle (axis cs:5.4,3);
\draw[draw=none,fill=steelblue31119180] (axis cs:5.6,0) rectangle (axis cs:6.4,5);
\draw[draw=none,fill=steelblue31119180] (axis cs:6.6,0) rectangle (axis cs:7.4,5);
\draw[draw=none,fill=steelblue31119180] (axis cs:7.6,0) rectangle (axis cs:8.4,4);
\draw[draw=none,fill=steelblue31119180] (axis cs:8.6,0) rectangle (axis cs:9.4,2);
\draw[draw=none,fill=steelblue31119180] (axis cs:9.6,0) rectangle (axis cs:10.4,2);
\draw[draw=none,fill=steelblue31119180] (axis cs:10.6,0) rectangle (axis cs:11.4,2);
\draw[draw=none,fill=steelblue31119180] (axis cs:11.6,0) rectangle (axis cs:12.4,2);
\draw[draw=none,fill=darkorange25512714] (axis cs:0.6,5) rectangle (axis cs:1.4,7);

\draw[draw=none,fill=darkorange25512714] (axis cs:1.6,5) rectangle (axis cs:2.4,7);
\draw[draw=none,fill=darkorange25512714] (axis cs:2.6,5) rectangle (axis cs:3.4,7);
\draw[draw=none,fill=darkorange25512714] (axis cs:3.6,5) rectangle (axis cs:4.4,7);
\draw[draw=none,fill=darkorange25512714] (axis cs:4.6,3) rectangle (axis cs:5.4,5);
\draw[draw=none,fill=darkorange25512714] (axis cs:5.6,5) rectangle (axis cs:6.4,7);
\draw[draw=none,fill=darkorange25512714] (axis cs:6.6,5) rectangle (axis cs:7.4,7);
\draw[draw=none,fill=darkorange25512714] (axis cs:7.6,4) rectangle (axis cs:8.4,6);
\draw[draw=none,fill=darkorange25512714] (axis cs:8.6,2) rectangle (axis cs:9.4,4);
\draw[draw=none,fill=darkorange25512714] (axis cs:9.6,2) rectangle (axis cs:10.4,4);
\draw[draw=none,fill=darkorange25512714] (axis cs:10.6,2) rectangle (axis cs:11.4,4);
\draw[draw=none,fill=darkorange25512714] (axis cs:11.6,2) rectangle (axis cs:12.4,4);
\draw[draw=none,fill=forestgreen4416044] (axis cs:0.6,7) rectangle (axis cs:1.4,12);

\draw[draw=none,fill=forestgreen4416044] (axis cs:1.6,7) rectangle (axis cs:2.4,12);
\draw[draw=none,fill=forestgreen4416044] (axis cs:2.6,7) rectangle (axis cs:3.4,12);
\draw[draw=none,fill=forestgreen4416044] (axis cs:3.6,7) rectangle (axis cs:4.4,12);
\draw[draw=none,fill=forestgreen4416044] (axis cs:4.6,5) rectangle (axis cs:5.4,7);
\draw[draw=none,fill=forestgreen4416044] (axis cs:5.6,7) rectangle (axis cs:6.4,9);
\draw[draw=none,fill=forestgreen4416044] (axis cs:6.6,7) rectangle (axis cs:7.4,9);
\draw[draw=none,fill=forestgreen4416044] (axis cs:7.6,6) rectangle (axis cs:8.4,8);
\draw[draw=none,fill=forestgreen4416044] (axis cs:8.6,4) rectangle (axis cs:9.4,6);
\draw[draw=none,fill=forestgreen4416044] (axis cs:9.6,4) rectangle (axis cs:10.4,6);
\draw[draw=none,fill=forestgreen4416044] (axis cs:10.6,4) rectangle (axis cs:11.4,6);
\draw[draw=none,fill=forestgreen4416044] (axis cs:11.6,4) rectangle (axis cs:12.4,6);
\draw[draw=none,fill=crimson2143940] (axis cs:0.6,12) rectangle (axis cs:1.4,14);

\draw[draw=none,fill=crimson2143940] (axis cs:1.6,12) rectangle (axis cs:2.4,14);
\draw[draw=none,fill=crimson2143940] (axis cs:2.6,12) rectangle (axis cs:3.4,14);
\draw[draw=none,fill=crimson2143940] (axis cs:3.6,12) rectangle (axis cs:4.4,16);
\draw[draw=none,fill=crimson2143940] (axis cs:4.6,7) rectangle (axis cs:5.4,10);
\draw[draw=none,fill=crimson2143940] (axis cs:5.6,9) rectangle (axis cs:6.4,14);
\draw[draw=none,fill=crimson2143940] (axis cs:6.6,9) rectangle (axis cs:7.4,14);
\draw[draw=none,fill=crimson2143940] (axis cs:7.6,8) rectangle (axis cs:8.4,13);
\draw[draw=none,fill=crimson2143940] (axis cs:8.6,6) rectangle (axis cs:9.4,11);
\draw[draw=none,fill=crimson2143940] (axis cs:9.6,6) rectangle (axis cs:10.4,11);
\draw[draw=none,fill=crimson2143940] (axis cs:10.6,6) rectangle (axis cs:11.4,11);
\draw[draw=none,fill=crimson2143940] (axis cs:11.6,6) rectangle (axis cs:12.4,11);
\draw[draw=none,fill=mediumpurple148103189] (axis cs:0.6,14) rectangle (axis cs:1.4,16);

\draw[draw=none,fill=mediumpurple148103189] (axis cs:1.6,14) rectangle (axis cs:2.4,16);
\draw[draw=none,fill=mediumpurple148103189] (axis cs:2.6,14) rectangle (axis cs:3.4,16);
\draw[draw=none,fill=mediumpurple148103189] (axis cs:3.6,16) rectangle (axis cs:4.4,18);
\draw[draw=none,fill=mediumpurple148103189] (axis cs:4.6,10) rectangle (axis cs:5.4,12);
\draw[draw=none,fill=mediumpurple148103189] (axis cs:5.6,14) rectangle (axis cs:6.4,16);
\draw[draw=none,fill=mediumpurple148103189] (axis cs:6.6,14) rectangle (axis cs:7.4,16);
\draw[draw=none,fill=mediumpurple148103189] (axis cs:7.6,13) rectangle (axis cs:8.4,15);
\draw[draw=none,fill=mediumpurple148103189] (axis cs:8.6,11) rectangle (axis cs:9.4,13);
\draw[draw=none,fill=mediumpurple148103189] (axis cs:9.6,11) rectangle (axis cs:10.4,13);
\draw[draw=none,fill=mediumpurple148103189] (axis cs:10.6,11) rectangle (axis cs:11.4,13);
\draw[draw=none,fill=mediumpurple148103189] (axis cs:11.6,11) rectangle (axis cs:12.4,13);
\draw[draw=none,fill=sienna1408675] (axis cs:0.6,16) rectangle (axis cs:1.4,18);

\draw[draw=none,fill=sienna1408675] (axis cs:1.6,16) rectangle (axis cs:2.4,18);
\draw[draw=none,fill=sienna1408675] (axis cs:2.6,16) rectangle (axis cs:3.4,19);
\draw[draw=none,fill=sienna1408675] (axis cs:3.6,18) rectangle (axis cs:4.4,21);
\draw[draw=none,fill=sienna1408675] (axis cs:4.6,12) rectangle (axis cs:5.4,16);
\draw[draw=none,fill=sienna1408675] (axis cs:5.6,16) rectangle (axis cs:6.4,18);
\draw[draw=none,fill=sienna1408675] (axis cs:6.6,16) rectangle (axis cs:7.4,21);
\draw[draw=none,fill=sienna1408675] (axis cs:7.6,15) rectangle (axis cs:8.4,20);
\draw[draw=none,fill=sienna1408675] (axis cs:8.6,13) rectangle (axis cs:9.4,18);
\draw[draw=none,fill=sienna1408675] (axis cs:9.6,13) rectangle (axis cs:10.4,18);
\draw[draw=none,fill=sienna1408675] (axis cs:10.6,13) rectangle (axis cs:11.4,18);
\draw[draw=none,fill=sienna1408675] (axis cs:11.6,13) rectangle (axis cs:12.4,18);
\end{axis}

\end{tikzpicture}
        \tikzexternalenable
        \end{subfigure}
        \hfill
        \begin{subfigure}[c]{0.12\textwidth}
            \vspace{-6ex}
            \tikzexternaldisable

\scalebox{.63}{
    \setlength{\tabcolsep}{3.9pt}
    \renewcommand{\arraystretch}{1.1}
    \begin{tabular}{l}
        \hspace{3.5ex} {\protect\tikz[baseline=-.65ex] \protect\draw[fill=color_brown, draw=color_brown, line width=1.4pt] plot[mark size=2.0, only marks, mark=square*, mark options={xscale=1.5, yscale=1}] (0, 0);} FC 2 \\
        \hspace{3.5ex} {\protect\tikz[baseline=-.65ex] \protect\draw[fill=color_purle, draw=color_purle, line width=1.4pt] plot[mark size=2.0, only marks, mark=square*, mark options={xscale=1.5, yscale=1}] (0, 0);} FC 1 \\
        \hspace{3.5ex} {\protect\tikz[baseline=-.65ex] \protect\draw[fill=color_red, draw=color_red, line width=1.4pt] plot[mark size=2.0, only marks, mark=square*, mark options={xscale=1.5, yscale=1}] (0, 0);} Out proj \\
        \hspace{3.5ex} {\protect\tikz[baseline=-.65ex] \protect\draw[fill=color_green, draw=color_green, line width=1.4pt] plot[mark size=2.0, only marks, mark=square*, mark options={xscale=1.5, yscale=1}] (0, 0);} Q proj \\
        \hspace{3.5ex} {\protect\tikz[baseline=-.65ex] \protect\draw[fill=color_orange, draw=color_orange, line width=1.4pt] plot[mark size=2.0, only marks, mark=square*, mark options={xscale=1.5, yscale=1}] (0, 0);} V proj\\
        \hspace{3.5ex} {\protect\tikz[baseline=-.65ex] \protect\draw[fill=color_blue, draw=color_blue, line width=1.4pt] plot[mark size=2.0, only marks, mark=square*, mark options={xscale=1.5, yscale=1}] (0, 0);} K proj \\
    \end{tabular}
}

\tikzexternalenable
        \end{subfigure}

        \vspace{-3.5ex}

        \captionof{figure}{\small{Allocated kernels for \optsmall. \label{fig:allocated_kernels}}}
\end{wrapfigure}

\vspace{-0.5ex}

We next evaluate our kernel allocation method on the \optsmall model.
It supports bit allocation at any granularity, including fractional averages, providing practitioners with a flexible model selection tool under deployment constraints.
\cref{fig:kernel_allocation} reports results for varying average bit budgets, showing consistent improvements as the budget increases.
\cref{fig:allocated_kernels} illustrates kernel allocation with a 3.5-bit average, where more kernels are assigned to FC2 and output projection layers in the final blocks.
This aligns with prior observations \citep{bondarenko2023quantizable,frantar2023optq} that these layers are particularly important and sensitive to compression.

\vspace{-0.8ex}

\begin{wrapfigure}[8]{r}{0.25\textwidth}
    \centering

        \vspace{-5.5ex}

        \tikzexternaldisable
        \centering
        \scriptsize
        \setlength{\figurewidth}{3.8cm}
        \setlength{\figureheight}{3.1cm}
        \begin{tikzpicture}

  \definecolor{crimson2143940}{RGB}{214,39,40}
  \definecolor{darkgray176}{RGB}{176,176,176}
  \definecolor{steelblue31119180}{RGB}{31,119,180}

  \begin{axis}[
      height=\figureheight,
      major tick length=1ex,
      tick align=outside,
      tick pos=left,
      width=\figurewidth,
      x grid style={darkgray176},
      xlabel={Average bit budget},
      xmin=1.434, xmax=4.646,
      y grid style={darkgray176},
      ylabel={C4 Perplexity},
      ymin=25.481, ymax=32.279,
      ytick style={color=black},
      xtick={1.58,2,2.5,3,3.5,4,4.5},
      xticklabels={
          \(\displaystyle {1.58}\),
          \(\displaystyle {2}\),
          \(\displaystyle {2.5}\),
          \(\displaystyle {3}\),
          \(\displaystyle {3.5}\),
          \(\displaystyle {4}\),
          \(\displaystyle {4.5}\)
        },
      ytick={24,26,28,30,32,34},
      yticklabels={
          \(\displaystyle {24}\),
          \(\displaystyle {26}\),
          \(\displaystyle {28}\),
          \(\displaystyle {30}\),
          \(\displaystyle {32}\),
          \(\displaystyle {34}\)
        }
    ]
    \addplot [thick, crimson2143940, opacity=0.8, mark=pentagon*, mark size=1.75, mark options={solid}]
    table {%
        1.58 31.97
        2 28.62
        2.5 27.73
        3 26.48
        3.5 26.26
        4 25.9
        4.5 25.79
      };
  \end{axis}

\end{tikzpicture}
        \tikzexternalenable

        \tikzexternaldisable

        \vspace{-2.0ex}

        \captionof{figure}{\small
            \optsmall performance w.r.t. bit budget.%
            \label{fig:kernel_allocation}
        }

        \tikzexternalenable
\end{wrapfigure}

In addition, our framework’s flexibility enables direct comparison with \bitnetternary \citep{ma2024era}, which employs ternary weights.
With a 1.58-bit budget, our model achieves reasonable results, whereas \bitnetternary reaches a C4 perplexity of 10199.89 due to finetuning instability, consistent with \cite{xu2024onebit}.
We also compare against \shiftaddllm \citep{haoran2024addshiftllm}, a \gls{PTQ} method supporting bit allocation.
Our approach performs substantially better (32.23 with a 2-bit budget vs. 435.84 for their mixed 2.2-bit allocation, see Table 17 in \shiftaddllm).

\vspace{-0.7ex}

\subsection{Discussion on Complexity and Latency}

\begin{wrapfigure}[8]{r}{0.29\textwidth}
    \centering

        \vspace{-5ex}

        \tikzexternaldisable
        \centering
        \scriptsize
        \setlength{\figurewidth}{4.8cm}
        \setlength{\figureheight}{2.4cm}
        \begin{tikzpicture}

\definecolor{darkgray176}{RGB}{176,176,176}
\definecolor{darkorange25512714}{RGB}{255,127,14}
\definecolor{goldenrod18818934}{RGB}{188,189,34}

\begin{axis}[
height=\figureheight,
major tick length=1ex,
tick align=outside,
tick pos=left,
title={OPT-6.7B},
width=\figurewidth,
x grid style={darkgray176},
xlabel={Memory (GB)},
xmin=0, xmax=36.601831056,
xtick style={color=black},
xtick={0,10,20,30,40},
xticklabels={
  \(\displaystyle {0}\),
  \(\displaystyle {10}\),
  \(\displaystyle {20}\),
  \(\displaystyle {30}\),
  \(\displaystyle {40}\)
},
y grid style={darkgray176},
ymin=-0.49, ymax=1.49,
ytick style={color=black},
ytick={0,1},
yticklabels={MBOK,MoS}
]
\draw[draw=none,fill=darkorange25512714] (axis cs:0,-0.4) rectangle (axis cs:2.17868042,0.4);
\draw[draw=none,fill=darkorange25512714] (axis cs:0,0.6) rectangle (axis cs:11.61962891,1.4);
\draw[draw=none,fill=goldenrod18818934] (axis cs:2.17868042,-0.4) rectangle (axis cs:13.79830933,0.4);
\draw[draw=none,fill=goldenrod18818934] (axis cs:11.61962891,0.6) rectangle (axis cs:34.85888672,1.4);
\end{axis}

\end{tikzpicture}
        \tikzexternalenable

        \tikzexternaldisable

        \vspace{-3.5ex}

        \captionof{figure}{\small
            Estimated memory for finetuning for weights ( {\protect\tikz[baseline=-.65ex] \protect\draw[fill=color_orange, draw=color_orange, line width=1.4pt] plot[mark size=2.0, only marks, mark=square*, mark options={xscale=1.5, yscale=1}] (0, 0);} ) and optimizer states ( {\protect\tikz[baseline=-.65ex] \protect\draw[fill=color_olive, draw=color_olive, line width=1.4pt] plot[mark size=2.0, only marks, mark=square*, mark options={xscale=1.5, yscale=1}] (0, 0);} ).
            \label{fig:opt_train_memory}
        }

        \tikzexternalenable
\end{wrapfigure}

\vspace{-1ex}

We emphasize the efficiency of our method during finetuning by comparing \mos \citep{jo2024mixture} with our approach using 3 Boolean kernels on the \opthuge model.
Because we optimize directly in the Boolean domain, each weight requires only 1 bit, whereas \mos relies on 16-bit latent weights.
Moreover, we finetune only the last Boolean kernel, with the optimizer storing a single 16-bit momentum per weight.
In contrast, Adam \citep{Kingma2014} for latent weights needs two 16-bit momenta per weight.
\cref{fig:opt_train_memory} shows the estimated memory for a minibatch of one, highlighting the substantial memory savings of our method.
These gains could be further amplified by incorporating optimizer state compression techniques such as \galore \citep{zhao2024galore}.

Beyond our theoretical analysis of training complexity (see \cref{sec:training_complexity}), we provide empirical evidence of practical GPU latency gains (see \cref{sec:latency_vq} for detailed analyses). 
Leveraging the \textrm{BitBLAS} library \citep{ladderosdi24} for 1-bit matrix multiplications (INT1 weights, FP16 activations) on an A100 GPU, \ours achieves up to an $8.7\times$ speedup for \llamabig linear layers compared to \halffp baselines at a batch size of 1 (\cref{tab:latency_llama_13b_main}).
Our method significantly outpaces existing binarization and scalar quantization techniques, and runs much faster than SOTA 2-bit vector quantization baselines like \quipsharp \citep{tseng2024quipsharp} and \qtip \citep{tseng2024qtip} while delivering comparable performance. Taken together, our native Boolean approach is a highly efficient alternative to VQ methods, with even greater performance expected on dedicated Boolean hardware.

\begin{table}[H]
    \setlength{\tabcolsep}{3.8pt}
    \centering
    \caption{Measured latency (ms) of linear layers in  \llamabig, with values in parentheses denoting speed-up relative to the \halffp baseline.}
    \label{tab:latency_llama_13b_main}

    \renewcommand{\arraystretch}{1.2}

    \scalebox{.9}{
        \begin{tabular}{cccccccc}
            \toprule
            \textsc{weight size}          & \halffp           & \quipsharp  \citep{tseng2024quipsharp} & \qtip \citep{tseng2024qtip}      & \ours (Ours)                                                                 \\
            \midrule
            \midrule
            \small{$5120  \times 5120 $}  & \small{$0.16540$} & \small{$0.62260$} ($0.27\times$)       & \small{$1.96368$} ($0.08\times$) & \cellcolor[HTML]{d6e9c9} \small{$\mathbf{0.05074}$ ($\mathbf{3.25}\times$) } \\
            \small{$5120  \times 13824$}  & \small{$0.42830$} & \small{$0.62836$} ($0.68\times$)       & \small{$5.23681$} ($0.09\times$) & \cellcolor[HTML]{d6e9c9} \small{$\mathbf{0.05098}$ ($\mathbf{8.40}\times$) } \\
            \small{$13824  \times  5120$} & \small{$0.43411$} & \small{$0.62840$} ($0.69\times$)       & \small{$5.21193$} ($0.08\times$) & \cellcolor[HTML]{d6e9c9} \small{$\mathbf{0.04987}$ ($\mathbf{8.70}\times$) } \\
            \bottomrule
        \end{tabular}
    }

\end{table}

\section{Conclusions} \label{sec:conclusion}

We introduced Multiple Boolean Kernels (\ours), a novel framework for low-bit finetuning \glspl{LLM}. 
By utilizing Boolean weights and optimizing them directly in the Boolean domain, our framework significantly reduces both memory and computation costs during \emph{both} finetuning and inference. 
The flexible multi-Boolean structure, along with the proposed successive \gls{SVID}, effectively transfers knowledge from a source \gls{FP} model. 
Through extensive experiments on \glspl{LLM} of various sizes, we demonstrate that our method approaches \gls{FP} performance while achieving the best accuracy-compression trade-off compared to existing quantization and binarization methods.

\paragraph{Limitations.}
Our method, like other binarized neural networks, could not be assessed on native Boolean accelerators due to hardware being optimized for real arithmetic.
Nevertheless, we demonstrated strong results even on modern hardware, underscoring the promise of our approach and motivating future development of accelerators tailored to Boolean computation.

\section*{Acknowledgements}

We thank Jean-Claude Belfiore and Li Rong for their feedback during patent review meetings, and Yun Yaw Chu for providing infrastructure support.

\section*{Author Contributions Statement}
\label{sec:author_contributions}

List of Authors: Ba-Hien Tran (B.H.T.), Van Minh Nguyen (V.M.N.).

B.H.T. initiated the low-complexity multi-kernel research direction, defined the narrative of the paper, formulated the core algorithm, proved the mathematical optimality of successive kernel extraction, and proposed both the efficient optimization strategies and the method for estimating weight importance. B.H.T. also designed and conducted all experiments. V.M.N. proposed the heuristic algorithm for kernel allocation. Both authors contributed to the writing and editing of the manuscript, focusing on their respective areas of contribution.

\bibliography{bib/main_biblio.bib}

@inproceedings{xu2024onebit,
  title     = {{OneBit: Towards Extremely Low-bit Large Language Models}},
  author    = {Yuzhuang Xu and Xu Han and Zonghan Yang and Shuo Wang and Qingfu Zhu and Zhiyuan Liu and Weidong Liu and Wanxiang Che},
  booktitle = {The Thirty-eighth Annual Conference on Neural Information Processing Systems},
  year      = {2024},
  url       = {https://openreview.net/forum?id=ZwiG9KjfHV}
}

@inproceedings{jo2024mixture,
  title     = {{Mixture of Scales: Memory-Efficient Token-Adaptive Binarization for Large Language Models}},
  author    = {Dongwon Jo and Taesu Kim and Yulhwa Kim and Jae-Joon Kim},
  booktitle = {The Thirty-eighth Annual Conference on Neural Information Processing Systems},
  year      = {2024},
  url       = {https://openreview.net/forum?id=pGOBEYcXzs}
}

@inproceedings{bulat2024qbb,
  title     = {{{QBB}: Quantization with Binary Bases for {LLM}s}},
  author    = {Adrian Bulat and Yassine Ouali and Georgios Tzimiropoulos},
  booktitle = {The Thirty-eighth Annual Conference on Neural Information Processing Systems},
  year      = {2024},
  url       = {https://openreview.net/forum?id=Kw6MRGFx0R}
}

@article{wang2023bitnet,
  title   = {{BitNet: Scaling 1-bit Transformers for Large Language Models}},
  author  = {Wang, Hongyu and Ma, Shuming and Dong, Li and Huang, Shaohan and Wang, Huaijie and Ma, Lingxiao and Yang, Fan and Wang, Ruiping and Wu, Yi and Wei, Furu},
  journal = {arXiv preprint arXiv:2310.11453},
  year    = {2023}
}

@article{ma2024era,
  title     = {{The Era of 1-bit LLMs: All Large Language Models are in 1.58 Bits}},
  author    = {Ma, Shuming and Wang, Hongyu and Ma, Lingxiao and Wang, Lei and Wang, Wenhui and Huang, Shaohan and Dong, Lifeng and Wang, Ruiping and Xue, Jilong and Wei, Furu},
  journal   = {arXiv preprint arXiv:2402.17764},
  volume    = {1},
  year      = {2024},
  publisher = {arXivpreprint}
}

@article{hinton2015kd,
  title     = {{Distilling the Knowledge in a Neural Network}},
  author    = {Hinton, Geoffrey and Oriol, Vinyals and Jeff Dean},
  journal   = {arXiv preprint arXiv:1503.02531},
  volume    = {1},
  year      = {2015},
  publisher = {arXivpreprint}
}

@inproceedings{frantar2023optq,
  title     = {{{OPTQ}: Accurate Quantization for Generative Pre-trained Transformers}},
  author    = {Elias Frantar and Saleh Ashkboos and Torsten Hoefler and Dan Alistarh},
  booktitle = {The Eleventh International Conference on Learning Representations },
  year      = {2023},
  url       = {https://openreview.net/forum?id=tcbBPnfwxS}
}

@article{zhang2022opt,
  title   = {{OPT: Open Pre-trained Transformer Language Models}},
  author  = {Zhang, Susan and Roller, Stephen and Goyal, Naman and Artetxe, Mikel and Chen, Moya and Chen, Shuohui and Dewan, Christopher and Diab, Mona and Li, Xian and Lin, Xi Victoria and others},
  journal = {arXiv preprint arXiv:2205.01068},
  year    = {2022}
}

@inproceedings{yao2022zeroquant,
  title     = {{ZeroQuant: Efficient and Affordable Post-Training Quantization for Large-Scale Transformers}},
  author    = {Zhewei Yao and Reza Yazdani Aminabadi and Minjia Zhang and Xiaoxia Wu and Conglong Li and Yuxiong He},
  booktitle = {Advances in Neural Information Processing Systems},
  year      = {2022},
  url       = {https://openreview.net/forum?id=f-fVCElZ-G1}
}

@inproceedings{dettmers2022gptint,
  title     = {{{GPT}3.int8(): 8-bit Matrix Multiplication for Transformers at Scale}},
  author    = {Tim Dettmers and Mike Lewis and Younes Belkada and Luke Zettlemoyer},
  booktitle = {Advances in Neural Information Processing Systems},
  year      = {2022},
  url       = {https://openreview.net/forum?id=dXiGWqBoxaD}
}

@inproceedings{huang2024billm,
  title     = {{{B}i{LLM}: Pushing the Limit of Post-Training Quantization for {LLM}s}},
  author    = {Huang, Wei and Liu, Yangdong and Qin, Haotong and Li, Ying and Zhang, Shiming and Liu, Xianglong and Magno, Michele and Qi, Xiaojuan},
  booktitle = {Proceedings of the 41st International Conference on Machine Learning},
  pages     = {20023--20042},
  year      = {2024},
  volume    = {235},
  series    = {Proceedings of Machine Learning Research},
  month     = {21--27 Jul},
  publisher = {PMLR},
  url       = {https://proceedings.mlr.press/v235/huang24q.html}
}

@inproceedings{yuan2024pbllm,
  title     = {{{PB}-{LLM}: Partially Binarized Large Language Models}},
  author    = {Zhihang Yuan and Yuzhang Shang and Zhen Dong},
  booktitle = {The Twelfth International Conference on Learning Representations},
  year      = {2024},
  url       = {https://openreview.net/forum?id=BifeBRhikU}
}

@inproceedings{dong2025stbllm,
  title     = {{{STBLLM}: Breaking the 1-Bit Barrier with Structured Binary {LLM}s}},
  author    = {Peijie Dong and Lujun Li and Yuedong Zhong and DaYou Du and Ruibo FAN and Yuhan Chen and Zhenheng Tang and Qiang Wang and Wei Xue and Yike Guo and Xiaowen Chu},
  booktitle = {The Thirteenth International Conference on Learning Representations},
  year      = {2025},
  url       = {https://openreview.net/forum?id=6XUSDvBFkV}
}

@inproceedings{li2025arbllm,
  title     = {{{ARB}-{LLM}: Alternating Refined Binarizations for Large Language Models}},
  author    = {Zhiteng Li and Xianglong Yan and Tianao Zhang and Haotong Qin and Dong Xie and Jiang Tian and Zhongchao Shi and Linghe Kong and Yulun Zhang and Xiaokang Yang},
  booktitle = {The Thirteenth International Conference on Learning Representations},
  year      = {2025},
  url       = {https://openreview.net/forum?id=ZU8OdDLTts}
}

@techreport{sulle1990,
  title       = {Sulle funzioni bilineari, Giomale di Mathematiche ad Uso studenti Delle Uninersita. 11, 98--106.(An English translation by D Boley is available as University of Minnesota, Department of Computer Science)},
  author      = {Beltrami, E},
  year        = {1990},
  institution = {Technical Report 90--37}
}

@article{eckart1936approx,
  author  = {Eckart, C. and Young, G.},
  journal = {Psychometrika},
  title   = {{The Approximation of One Matrix by Another of Lower Rank}},
  year    = 1936
}

@inproceedings{vaswani2017attention,
  author    = {Vaswani, Ashish and Shazeer, Noam and Parmar, Niki and Uszkoreit, Jakob and Jones, Llion and Gomez, Aidan N and Kaiser, \L ukasz and Polosukhin, Illia},
  booktitle = {Advances in Neural Information Processing Systems},
  publisher = {Curran Associates, Inc.},
  title     = {{Attention is All you Need}},
  url       = {https://proceedings.neurips.cc/paper_files/paper/2017/file/3f5ee243547dee91fbd053c1c4a845aa-Paper.pdf},
  volume    = {30},
  year      = {2017}
}

@article{touvron2023llama,
  title   = {{LLaMA: Open and Efficient Foundation Language Models}},
  author  = {Touvron, Hugo and Lavril, Thibaut and Izacard, Gautier and Martinet, Xavier and Lachaux, Marie-Anne and Lacroix, Timoth{\'e}e and Rozi{\`e}re, Baptiste and Goyal, Naman and Hambro, Eric and Azhar, Faisal and others},
  journal = {arXiv preprint arXiv:2302.13971},
  year    = {2023}
}

@inproceedings{Brown2020,
  author    = {Brown, Tom and Mann, Benjamin and Ryder, Nick and Subbiah, Melanie and Kaplan, Jared D and Dhariwal, Prafulla and Neelakantan, Arvind and Shyam, Pranav and Sastry, Girish and Askell, Amanda and Agarwal, Sandhini and Herbert-Voss, Ariel and Krueger, Gretchen and Henighan, Tom and Child, Rewon and Ramesh, Aditya and Ziegler, Daniel and Wu, Jeffrey and Winter, Clemens and Hesse, Chris and Chen, Mark and Sigler, Eric and Litwin, Mateusz and Gray, Scott and Chess, Benjamin and Clark, Jack and Berner, Christopher and McCandlish, Sam and Radford, Alec and Sutskever, Ilya and Amodei, Dario},
  booktitle = {Advances in Neural Information Processing Systems},
  pages     = {1877--1901},
  publisher = {Curran Associates, Inc.},
  title     = {{Language Models are Few-Shot Learners}},
  url       = {https://proceedings.neurips.cc/paper_files/paper/2020/file/1457c0d6bfcb4967418bfb8ac142f64a-Paper.pdf},
  volume    = {33},
  year      = {2020}
}

@article{liu2024deepseek,
  title   = {{DeepSeek-V3 Technical Report}},
  author  = {Liu, Aixin and Feng, Bei and Xue, Bing and Wang, Bingxuan and Wu, Bochao and Lu, Chengda and Zhao, Chenggang and Deng, Chengqi and Zhang, Chenyu and Ruan, Chong and others},
  journal = {arXiv preprint arXiv:2412.19437},
  year    = {2024}
}

@inproceedings{Rastegari2016,
  author    = {Rastegari, Mohammad
               and Ordonez, Vicente
               and Redmon, Joseph
               and Farhadi, Ali},
  title     = {{XNOR-Net: ImageNet Classification Using Binary Convolutional Neural Networks}},
  booktitle = {Proceedings of the European Conference on Computer Vision (ECCV)},
  month     = {October},
  year      = {2016}
}

@inproceedings{Kingma2014,
  title     = {{Adam: A Method for Stochastic Optimization}},
  author    = {Kingma, Diederik P and Ba, Jimmy},
  booktitle = {International Conference on Learning Representations},
  year      = {2015}
}

@inproceedings{Hubara2016,
  title     = {Binarized neural networks},
  author    = {Hubara, Itay and Courbariaux, Matthieu and Soudry, Daniel and El-Yaniv, Ran and Bengio, Yoshua},
  booktitle = {Advances in neural information processing systems},
  year      = {2016},
  pages     = {4107--4115}
}

@inproceedings{Courbariaux2015,
  author    = {Courbariaux, Matthieu and Bengio, Yoshua and David, Jean-Pierre},
  booktitle = {Advances in Neural Information Processing Systems},
  pages     = {},
  publisher = {Curran Associates, Inc.},
  title     = {{BinaryConnect: Training Deep Neural Networks with Binary Weights during Propagations}},
  volume    = {28},
  year      = {2015}
}

@article{bengio2013ste,
  title   = {{Estimating or Propagating Gradients Through Stochastic Neurons for Conditional Computation}},
  author  = {Bengio, Yoshua and L{\'e}onard, Nicholas and Courville, Aaron},
  journal = {arXiv preprint arXiv:1308.3432},
  year    = {2013}
}

@inproceedings{nguyen2024bold,
  title     = {{BOLD: Boolean Logic Deep Learning}},
  author    = {Van Minh Nguyen and Cristian Ocampo and Aymen Askri and Louis Leconte and Ba-Hien Tran},
  booktitle = {The Thirty-eighth Annual Conference on Neural Information Processing Systems},
  year      = {2024},
  url       = {https://openreview.net/forum?id=DO9wPZOPjk}
}

@inproceedings{ko2024distillm,
  title     = {{{D}isti{LLM}: Towards Streamlined Distillation for Large Language Models}},
  author    = {Ko, Jongwoo and Kim, Sungnyun and Chen, Tianyi and Yun, Se-Young},
  booktitle = {Proceedings of the 41st International Conference on Machine Learning},
  pages     = {24872--24895},
  year      = {2024},
  volume    = {235},
  series    = {Proceedings of Machine Learning Research},
  month     = {21--27 Jul},
  publisher = {PMLR},
  url       = {https://proceedings.mlr.press/v235/ko24c.html}
}

@article{Fuchs2014,
  author    = {Fuchs, Eberhard and Fl{\"u}gge, Gabriele and others},
  journal   = {Neural plasticity},
  title     = {{Adult Neuroplasticity: More than 40 Years of Research}},
  year      = {2014},
  volume    = {2014},
  publisher = {Hindawi}
}

@book{Hebb2005,
  author    = {Hebb, Donald Olding},
  publisher = {Psychology press},
  title     = {{The Organization of Behavior: A Neuropsychological Theory}},
  year      = {2005}
}

@inproceedings{Sheng2023FlexGen,
  title     = {{{F}lex{G}en: High-Throughput Generative Inference of Large Language Models with a Single {GPU}}},
  author    = {Sheng, Ying and Zheng, Lianmin and Yuan, Binhang and Li, Zhuohan and Ryabinin, Max and Chen, Beidi and Liang, Percy and Re, Christopher and Stoica, Ion and Zhang, Ce},
  booktitle = {Proceedings of the 40th International Conference on Machine Learning},
  pages     = {31094--31116},
  year      = {2023},
  volume    = {202},
  series    = {Proceedings of Machine Learning Research},
  month     = {23--29 Jul},
  publisher = {PMLR}
}

@inproceedings{Lin2024AWQ,
  author    = {Lin, Ji and Tang, Jiaming and Tang, Haotian and Yang, Shang and Chen, Wei-Ming and Wang, Wei-Chen and Xiao, Guangxuan and Dang, Xingyu and Gan, Chuang and Han, Song},
  booktitle = {Proceedings of Machine Learning and Systems},
  pages     = {87--100},
  title     = {{AWQ: Activation-aware Weight Quantization for On-Device LLM Compression and Acceleration}},
  url       = {https://proceedings.mlsys.org/paper_files/paper/2024/file/42a452cbafa9dd64e9ba4aa95cc1ef21-Paper-Conference.pdf},
  volume    = {6},
  year      = {2024}
}

@inproceedings{lee2024owq,
  title     = {{OWQ: Outlier-aware Weight Quantization for Efficient Fine-tuning and Inference of Large Language Models}},
  author    = {Lee, Changhun and Jin, Jungyu and Kim, Taesu and Kim, Hyungjun and Park, Eunhyeok},
  booktitle = {Proceedings of the AAAI Conference on Artificial Intelligence},
  volume    = {38},
  number    = {12},
  pages     = {13355--13364},
  year      = {2024}
}

@inproceedings{liu2024llmqat,
  title     = {{{{LLM}-{QAT}: Data-Free Quantization Aware Training for Large Language Models}}},
  author    = {Liu, Zechun  and
               Oguz, Barlas  and
               Zhao, Changsheng  and
               Chang, Ernie  and
               Stock, Pierre  and
               Mehdad, Yashar  and
               Shi, Yangyang  and
               Krishnamoorthi, Raghuraman  and
               Chandra, Vikas},
  booktitle = {Findings of the Association for Computational Linguistics: ACL 2024},
  month     = August,
  year      = {2024},
  address   = {Bangkok, Thailand},
  publisher = {Association for Computational Linguistics},
  url       = {https://aclanthology.org/2024.findings-acl.26/},
  pages     = {467--484}
}

@inproceedings{Dettmers2023QLoRA,
  author    = {Dettmers, Tim and Pagnoni, Artidoro and Holtzman, Ari and Zettlemoyer, Luke},
  booktitle = {Advances in Neural Information Processing Systems},
  pages     = {10088--10115},
  publisher = {Curran Associates, Inc.},
  title     = {{QLoRA: Efficient Finetuning of Quantized LLMs}},
  url       = {https://proceedings.neurips.cc/paper_files/paper/2023/file/1feb87871436031bdc0f2beaa62a049b-Paper-Conference.pdf},
  volume    = {36},
  year      = {2023}
}

@inproceedings{Kim2024SqueezeLLM,
  title     = {{{S}queeze{LLM}: Dense-and-Sparse Quantization}},
  author    = {Kim, Sehoon and Hooper, Coleman Richard Charles and Gholami, Amir and Dong, Zhen and Li, Xiuyu and Shen, Sheng and Mahoney, Michael W. and Keutzer, Kurt},
  booktitle = {Proceedings of the 41st International Conference on Machine Learning},
  pages     = {23901--23923},
  year      = {2024},
  volume    = {235},
  series    = {Proceedings of Machine Learning Research},
  month     = {21--27 Jul},
  publisher = {PMLR},
  url       = {https://proceedings.mlr.press/v235/kim24f.html}
}

@inproceedings{dettmers2024spqr,
  title     = {{Sp{QR}: A Sparse-Quantized Representation for Near-Lossless {LLM} Weight Compression}},
  author    = {Tim Dettmers and Ruslan A. Svirschevski and Vage Egiazarian and Denis Kuznedelev and Elias Frantar and Saleh Ashkboos and Alexander Borzunov and Torsten Hoefler and Dan Alistarh},
  booktitle = {The Twelfth International Conference on Learning Representations},
  year      = {2024},
  url       = {https://openreview.net/forum?id=Q1u25ahSuy}
}

@inproceedings{shao2024omniquant,
  title     = {{OmniQuant: Omnidirectionally Calibrated Quantization for Large Language Models}},
  author    = {Wenqi Shao and Mengzhao Chen and Zhaoyang Zhang and Peng Xu and Lirui Zhao and Zhiqian Li and Kaipeng Zhang and Peng Gao and Yu Qiao and Ping Luo},
  booktitle = {The Twelfth International Conference on Learning Representations},
  year      = {2024},
  url       = {https://openreview.net/forum?id=8Wuvhh0LYW}
}

@inproceedings{bai2021binarybert,
  title     = {{{{B}inary{BERT}: Pushing the Limit of {BERT} Quantization}}},
  author    = {Bai, Haoli  and
               Zhang, Wei  and
               Hou, Lu  and
               Shang, Lifeng  and
               Jin, Jin  and
               Jiang, Xin  and
               Liu, Qun  and
               Lyu, Michael  and
               King, Irwin},
  booktitle = {Proceedings of the 59th Annual Meeting of the Association for Computational Linguistics and the 11th International Joint Conference on Natural Language Processing (Volume 1: Long Papers)},
  month     = August,
  year      = {2021},
  address   = {Online},
  publisher = {Association for Computational Linguistics},
  url       = {https://aclanthology.org/2021.acl-long.334/},
  pages     = {4334--4348}
}

@inproceedings{qin2022bibert,
  title     = {{Bi{BERT}: Accurate Fully Binarized {BERT}}},
  author    = {Haotong Qin and Yifu Ding and Mingyuan Zhang and Qinghua YAN and Aishan Liu and Qingqing Dang and Ziwei Liu and Xianglong Liu},
  booktitle = {International Conference on Learning Representations},
  year      = {2022},
  url       = {https://openreview.net/forum?id=5xEgrl_5FAJ}
}

@inproceedings{Liu2022BiT,
  author    = {Liu, Zechun and Oguz, Barlas and Pappu, Aasish and Xiao, Lin and Yih, Scott and Li, Meng and Krishnamoorthi, Raghuraman and Mehdad, Yashar},
  booktitle = {Advances in Neural Information Processing Systems},
  pages     = {14303--14316},
  publisher = {Curran Associates, Inc.},
  title     = {{BiT: Robustly Binarized Multi-distilled Transformer}},
  url       = {https://proceedings.neurips.cc/paper_files/paper/2022/file/5c1863f711c721648387ac2ef745facb-Paper-Conference.pdf},
  volume    = {35},
  year      = {2022}
}

@inproceedings{liu2023binary,
  title     = {{{Binary and Ternary Natural Language Generation}}},
  author    = {Liu, Zechun  and
               Oguz, Barlas  and
               Pappu, Aasish  and
               Shi, Yangyang  and
               Krishnamoorthi, Raghuraman},
  booktitle = {Proceedings of the 61st Annual Meeting of the Association for Computational Linguistics (Volume 1: Long Papers)},
  month     = July,
  year      = {2023},
  address   = {Toronto, Canada},
  publisher = {Association for Computational Linguistics},
  url       = {https://aclanthology.org/2023.acl-long.5/},
  doi       = {10.18653/v1/2023.acl-long.5},
  pages     = {65--77}
}

@inproceedings{Morcos2018,
  author    = {Morcos, Ari and Raghu, Maithra and Bengio, Samy},
  booktitle = {Advances in Neural Information Processing Systems},
  pages     = {},
  publisher = {Curran Associates, Inc.},
  title     = {{Insights on Representational Similarity in Neural Networks with Canonical Correlation}},
  url       = {https://proceedings.neurips.cc/paper_files/paper/2018/file/a7a3d70c6d17a73140918996d03c014f-Paper.pdf},
  volume    = {31},
  year      = {2018}
}

@inproceedings{merity2017pointer,
  title     = {{Pointer Sentinel Mixture Models}},
  author    = {Stephen Merity and Caiming Xiong and James Bradbury and Richard Socher},
  booktitle = {International Conference on Learning Representations},
  year      = {2017},
  url       = {https://openreview.net/forum?id=Byj72udxe}
}

@article{raffel2020c4,
  author  = {Colin Raffel and Noam Shazeer and Adam Roberts and Katherine Lee and Sharan Narang and Michael Matena and Yanqi Zhou and Wei Li and Peter J. Liu},
  title   = {{Exploring the Limits of Transfer Learning with a Unified Text-to-Text Transformer}},
  journal = {Journal of Machine Learning Research},
  year    = {2020},
  volume  = {21},
  number  = {140},
  pages   = {1--67},
  url     = {http://jmlr.org/papers/v21/20-074.html}
}

@inproceedings{loshchilov2018decoupled,
  title     = {{Decoupled Weight Decay Regularization}},
  author    = {Ilya Loshchilov and Frank Hutter},
  booktitle = {International Conference on Learning Representations},
  year      = {2019},
  url       = {https://openreview.net/forum?id=Bkg6RiCqY7}
}

@inproceedings{wen2023fdivergence,
  title     = {{{f-Divergence Minimization for Sequence-Level Knowledge Distillation}}},
  author    = {Wen, Yuqiao  and
               Li, Zichao  and
               Du, Wenyu  and
               Mou, Lili},
  booktitle = {Proceedings of the 61st Annual Meeting of the Association for Computational Linguistics (Volume 1: Long Papers)},
  month     = July,
  year      = {2023},
  address   = {Toronto, Canada},
  publisher = {Association for Computational Linguistics},
  url       = {https://aclanthology.org/2023.acl-long.605/},
  doi       = {10.18653/v1/2023.acl-long.605},
  pages     = {10817--10834}
}

@inproceedings{agarwal2024onpolicy,
  title     = {{On-Policy Distillation of Language Models: Learning from Self-Generated Mistakes}},
  author    = {Rishabh Agarwal and Nino Vieillard and Yongchao Zhou and Piotr Stanczyk and Sabela Ramos Garea and Matthieu Geist and Olivier Bachem},
  booktitle = {The Twelfth International Conference on Learning Representations},
  year      = {2024},
  url       = {https://openreview.net/forum?id=3zKtaqxLhW}
}

@article{sakaguchi2021winogrande,
  title     = {{Winogrande: An Adversarial Winograd Schema Challenge at Scale}},
  author    = {Sakaguchi, Keisuke and Bras, Ronan Le and Bhagavatula, Chandra and Choi, Yejin},
  journal   = {Communications of the ACM},
  volume    = {64},
  number    = {9},
  pages     = {99--106},
  year      = {2021},
  publisher = {ACM New York, NY, USA}
}

@inproceedings{zellers2019hellaswag,
  title     = {{{{H}ella{S}wag: Can a Machine Really Finish Your Sentence?}}},
  author    = {Zellers, Rowan  and
               Holtzman, Ari  and
               Bisk, Yonatan  and
               Farhadi, Ali  and
               Choi, Yejin},
  booktitle = {Proceedings of the 57th Annual Meeting of the Association for Computational Linguistics},
  month     = July,
  year      = {2019},
  address   = {Florence, Italy},
  publisher = {Association for Computational Linguistics},
  url       = {https://aclanthology.org/P19-1472/},
  doi       = {10.18653/v1/P19-1472},
  pages     = {4791--4800}
}

@inproceedings{bisk2020piqa,
  title     = {{PIQA: Reasoning about Physical Commonsense in Natural Language}},
  author    = {Bisk, Yonatan and Zellers, Rowan and Gao, Jianfeng and Choi, Yejin and others},
  booktitle = {Proceedings of the AAAI conference on artificial intelligence},
  volume    = {34},
  number    = {05},
  pages     = {7432--7439},
  year      = {2020}
}

@inproceedings{clark2019boolq,
  title     = {{{{B}ool{Q}: Exploring the Surprising Difficulty of Natural Yes/No Questions}}},
  author    = {Clark, Christopher  and
               Lee, Kenton  and
               Chang, Ming-Wei  and
               Kwiatkowski, Tom  and
               Collins, Michael  and
               Toutanova, Kristina},
  booktitle = {Proceedings of the 2019 Conference of the North {A}merican Chapter of the Association for Computational Linguistics: Human Language Technologies, Volume 1 (Long and Short Papers)},
  month     = jun,
  year      = {2019},
  address   = {Minneapolis, Minnesota},
  publisher = {Association for Computational Linguistics},
  url       = {https://aclanthology.org/N19-1300/},
  doi       = {10.18653/v1/N19-1300},
  pages     = {2924--2936}
}

@article{clark2018arc,
  title   = {{Think you have Solved Question Answering? Try ARC, the AI2 Reasoning Challenge}},
  author  = {Clark, Peter and Cowhey, Isaac and Etzioni, Oren and Khot, Tushar and Sabharwal, Ashish and Schoenick, Carissa and Tafjord, Oyvind},
  journal = {arXiv preprint arXiv:1803.05457},
  year    = {2018}
}

@article{touvron2023llama2,
  title   = {{Llama 2: Open Foundation and Fine-Tuned Chat Models}},
  author  = {Touvron, Hugo and Martin, Louis and Stone, Kevin and Albert, Peter and Almahairi, Amjad and Babaei, Yasmine and Bashlykov, Nikolay and Batra, Soumya and Bhargava, Prajjwal and Bhosale, Shruti and others},
  journal = {arXiv preprint arXiv:2307.09288},
  year    = {2023}
}

@inproceedings{dettmers2023thecase,
  title     = {{The case for 4-bit precision: k-bit Inference Scaling Laws}},
  author    = {Dettmers, Tim and Zettlemoyer, Luke},
  booktitle = {Proceedings of the 40th International Conference on Machine Learning},
  pages     = {7750--7774},
  year      = {2023},
  volume    = {202},
  series    = {Proceedings of Machine Learning Research},
  month     = {23--29 Jul},
  publisher = {PMLR},
  url       = {https://proceedings.mlr.press/v202/dettmers23a.html}
}

@inproceedings{kumar2025scaling,
  title     = {{Scaling Laws for Precision}},
  author    = {Tanishq Kumar and Zachary Ankner and Benjamin Frederick Spector and Blake Bordelon and Niklas Muennighoff and Mansheej Paul and Cengiz Pehlevan and Christopher Re and Aditi Raghunathan},
  booktitle = {The Thirteenth International Conference on Learning Representations},
  year      = {2025},
  url       = {https://openreview.net/forum?id=wg1PCg3CUP}
}

@inproceedings{liu2024bitdelta,
  author    = {Liu, James and Xiao, Guangxuan and Li, Kai and Lee, Jason D. and Han, Song and Dao, Tri and Cai, Tianle},
  booktitle = {Advances in Neural Information Processing Systems},
  pages     = {13579--13600},
  publisher = {Curran Associates, Inc.},
  title     = {{BitDelta: Your Fine-Tune May Only Be Worth One Bit}},
  url       = {https://proceedings.neurips.cc/paper_files/paper/2024/file/187d94b3c93343f0e925b5cf729eadd5-Paper-Conference.pdf},
  volume    = {37},
  year      = {2024}
}

@inproceedings{zhao2024galore,
  title     = {{{G}a{L}ore: Memory-Efficient {LLM} Training by Gradient Low-Rank Projection}},
  author    = {Zhao, Jiawei and Zhang, Zhenyu and Chen, Beidi and Wang, Zhangyang and Anandkumar, Anima and Tian, Yuandong},
  booktitle = {Proceedings of the 41st International Conference on Machine Learning},
  pages     = {61121--61143},
  year      = {2024},
  volume    = {235},
  series    = {Proceedings of Machine Learning Research},
  month     = {21--27 Jul},
  publisher = {PMLR},
  url       = {https://proceedings.mlr.press/v235/zhao24s.html}
}

@book{Jordan1950,
  author    = {Charles Jordan},
  publisher = {Chelsea Publishing Company},
  title     = {{Calculus of Finite Differences}},
  year      = {1950},
  address   = {New York},
  edition   = {2nd},
  doi       = {https://doi.org/10.1017/S0025557200230271}
}

@inproceedings{Paszke2019,
  author    = {Paszke, Adam and Gross, Sam and Massa, Francisco and Lerer, Adam and Bradbury, James and Chanan, Gregory and Killeen, Trevor and Lin, Zeming and Gimelshein, Natalia and Antiga, Luca and Desmaison, Alban and Kopf, Andreas and Yang, Edward and DeVito, Zachary and Raison, Martin and Tejani, Alykhan and Chilamkurthy, Sasank and Steiner, Benoit and Fang, Lu and Bai, Junjie and Chintala, Soumith},
  booktitle = {Advances in Neural Information Processing Systems},
  pages     = {},
  publisher = {Curran Associates, Inc.},
  title     = {{PyTorch: An Imperative Style, High-Performance Deep Learning Library}},
  volume    = {32},
  year      = {2019}
}

@inproceedings{gromov2025the,
  title     = {{The Unreasonable Ineffectiveness of the Deeper Layers}},
  author    = {Andrey Gromov and Kushal Tirumala and Hassan Shapourian and Paolo Glorioso and Dan Roberts},
  booktitle = {The Thirteenth International Conference on Learning Representations},
  year      = {2025},
  url       = {https://openreview.net/forum?id=ngmEcEer8a}
}

@inproceedings{xu2025understanding,
  title     = {{Understanding the Difficulty of Low-Precision Post-Training Quantization for {LLM}s}},
  author    = {Zifei Xu and Sayeh Sharify and Wanzin Yazar and Tristan J Webb and Xin Wang},
  booktitle = {ICLR 2025 Workshop on Sparsity in LLMs},
  year      = {2025},
  url       = {https://openreview.net/forum?id=fx9eAKwZKk}
}

@article{nguyen2023boolean,
  title   = {{Variation and Boolean Logic BackPropagation}},
  author  = {Nguyen, Van Minh},
  journal = {arXiv preprint arXiv:2311.07427},
  year    = {2023}
}

@inproceedings{chen2024dbllm,
  title     = {{{{DB}-{LLM}: Accurate Dual-Binarization for Efficient {LLM}s}}},
  author    = {Chen, Hong  and
               Lv, Chengtao  and
               Ding, Liang  and
               Qin, Haotong  and
               Zhou, Xiabin  and
               Ding, Yifu  and
               Liu, Xuebo  and
               Zhang, Min  and
               Guo, Jinyang  and
               Liu, Xianglong  and
               Tao, Dacheng},
  booktitle = {Findings of the Association for Computational Linguistics: ACL 2024},
  month     = aug,
  year      = {2024},
  address   = {Bangkok, Thailand},
  publisher = {Association for Computational Linguistics},
  url       = {https://aclanthology.org/2024.findings-acl.516/},
  doi       = {10.18653/v1/2024.findings-acl.516},
  pages     = {8719--8730}
}

@inproceedings{wang2025bitstack,
  title     = {{BitStack: Any-Size Compression of Large Language Models in Variable Memory Environments}},
  author    = {Xinghao Wang and Pengyu Wang and Bo Wang and Dong Zhang and Yunhua Zhou and Xipeng Qiu},
  booktitle = {The Thirteenth International Conference on Learning Representations},
  year      = {2025},
  url       = {https://openreview.net/forum?id=lBntjGbyv0}
}

@inproceedings{tseng2024qtip,
  author    = {Tseng, Albert and Sun, Qingyao and Hou, David and De, Christopher},
  booktitle = {Advances in Neural Information Processing Systems},
  pages     = {59597--59620},
  publisher = {Curran Associates, Inc.},
  title     = {{QTIP: Quantization with Trellises and Incoherence Processing}},
  url       = {https://proceedings.neurips.cc/paper_files/paper/2024/file/6de2e84b8da47bb2eb5e2ac96c63d2b0-Paper-Conference.pdf},
  volume    = {37},
  year      = {2024}
}

@inproceedings{tseng2024quipsharp,
  title     = {{{Q}u{IP}\#: Even Better {LLM} Quantization with Hadamard Incoherence and Lattice Codebooks}},
  author    = {Tseng, Albert and Chee, Jerry and Sun, Qingyao and Kuleshov, Volodymyr and De Sa, Christopher},
  booktitle = {Proceedings of the 41st International Conference on Machine Learning},
  pages     = {48630--48656},
  year      = {2024},
  volume    = {235},
  series    = {Proceedings of Machine Learning Research},
  month     = {21--27 Jul},
  publisher = {PMLR}
}

@inproceedings{haoran2024addshiftllm,
  author    = {You, Haoran and Guo, Yipin and Fu, Yichao and Zhou, Wei and Shi, Huihong and Zhang, Xiaofan and Kundu, Souvik and Yazdanbakhsh, Amir and Lin, Yingyan (Celine)},
  booktitle = {Advances in Neural Information Processing Systems},
  pages     = {24822--24848},
  publisher = {Curran Associates, Inc.},
  title     = {{ShiftAddLLM: Accelerating Pretrained LLMs via Post-Training Multiplication-Less Reparameterization}},
  url       = {https://proceedings.neurips.cc/paper_files/paper/2024/file/2c30a37c75f062e0bf79297c73db8c6c-Paper-Conference.pdf},
  volume    = {37},
  year      = {2024}
}

@inproceedings{chee2023quip,
  author    = {Chee, Jerry and Cai, Yaohui and Kuleshov, Volodymyr and De Sa, Christopher M},
  booktitle = {Advances in Neural Information Processing Systems},
  pages     = {4396--4429},
  publisher = {Curran Associates, Inc.},
  title     = {{QuIP: 2-Bit Quantization of Large Language Models With Guarantees}},
  url       = {https://proceedings.neurips.cc/paper_files/paper/2023/file/0df38cd13520747e1e64e5b123a78ef8-Paper-Conference.pdf},
  volume    = {36},
  year      = {2023}
}

@inproceedings{bondarenko2023quantizable,
  author    = {Bondarenko, Yelysei and Nagel, Markus and Blankevoort, Tijmen},
  booktitle = {Advances in Neural Information Processing Systems},
  pages     = {75067--75096},
  publisher = {Curran Associates, Inc.},
  title     = {{Quantizable Transformers: Removing Outliers by Helping Attention Heads Do Nothing}},
  url       = {https://proceedings.neurips.cc/paper_files/paper/2023/file/edbcb7583fd8921dad78adecfe06a99b-Paper-Conference.pdf},
  volume    = {36},
  year      = {2023}
}

@article{gray1984vector,
  title     = {{Vector Quantization}},
  author    = {Gray, Robert},
  journal   = {IEEE ASSP Magazine},
  volume    = {1},
  number    = {2},
  pages     = {4--29},
  year      = {1984},
  publisher = {IEEE}
}

@inproceedings{ladderosdi24,
  author    = {Lei Wang and Lingxiao Ma and Shijie Cao and Quanlu Zhang and Jilong Xue and Yining Shi and Ningxin Zheng and Ziming Miao and Fan Yang and Ting Cao and Yuqing Yang and Mao Yang},
  title     = {{Ladder: Enabling Efficient Low-Precision Deep Learning Computing through Hardware-aware Tensor Transformation}},
  booktitle = {18th USENIX Symposium on Operating Systems Design and Implementation (OSDI 24)},
  year      = {2024},
  isbn      = {978-1-939133-40-3},
  address   = {Santa Clara, CA},
  pages     = {307--323},
  url       = {https://www.usenix.org/conference/osdi24/presentation/wang-lei},
  publisher = {USENIX Association}
}
\bibliographystyle{iclr2026_conference}

\newpage

\appendix

\vspace{-5ex}

{\bf \Large{Appendix}}

\vspace{-1.8ex}

\begin{spacing}{0.2}
    \addtocontents{toc}{\protect\setcounter{tocdepth}{2}}
    \renewcommand{\contentsname}{\textbf{Table of Contents}\vskip1pt\hrule}
    \tableofcontents
    \vskip0.6pt\hrule
\end{spacing}

\section{Primer on Boolean Neural Networks} \label{sec:app_bool_nn}

For completeness, this section reviews the concepts and methodology of Boolean neural networks as proposed by \cite{nguyen2023boolean,nguyen2024bold}.

\subsection{Neuron Design}

\paragraph{Boolean Neuron.}
Consider the $l$-th Boolean linear layer; in the forward pass, the output of the next layer is defined as \cite{nguyen2024bold}:

\begin{align}
    \mbY_{[k,j]}^{(l)} =  \mbb_{[j]}^{(l)} + \sum_{i=1}^{n} \mathrm{L}(\mbX_{[k,i]}^{(l), \mbW_{[i,j]}^{(l)}}), \qquad 1 \leq j \leq m, \label{eq:Preactivation}
\end{align}

where $k$ denotes the sample index in the batch, and $\mathrm{L}$ is a logic gate such as $\mathbf{and}, \mathbf{or}, \mathbf{xor}$, or $\mathbf{xnor}$;
The weights $\mbW_{[i,j]}^{(l)}$ are Boolean values $\{\textsc{true}, \textsc{false}\}$ or $\{-1, +1\}$, as typically used in practical implementations.
$n$ and $m$ are the number of input and output neurons, respectively.
As the most extreme use case, the input data are also Boolean values.
The above summation is understood as the counting of \textsc{true} values.
We emphasize that the framework is flexible, as it allows Boolean linear layers to be connected through activation layers, layer normalization, arithmetic layers, or other types of layers.

\paragraph{Mixed Boolean-Real Neuron.}
To enable flexible integration and coexistence of Boolean designs with real-valued components in deep models, we consider two cases of mixed-type data: (i) Boolean weights with real-valued inputs, and (ii) real-valued weights with Boolean inputs.
This paper focuses on the first case.
These scenarios are addressed through an extension of Boolean logic to accommodate mixed-type data.
To proceed, we introduce the essential notations and definitions. Specifically, we define $\bbB \triangleq \{\true, \false\}$ as the Boolean domain, equipped with standard Boolean logic operations.

\begin{mybox}
    \begin{definition}[Three-valued logic]\label{def:ThreeValueLogic}
        We define the mixed logic domain as $\bbM \triangleq \bbB \cup \{0\}$, where $0$ represents an undefined or neutral value.
        The logic connectives in $\bbM$ are defined in alignment with standard Boolean logic, as follows.
        First, the negation operator is extended as: $\neg \true = \false$, $\neg \false = \true$, and $\neg 0 = 0$.
        Next, let $\Lm$ denote a generic logic connective (e.g., \textsc{and}, \textsc{or}).
        We distinguish its use in $\bbM$ and $\bbB$ by writing $\Lm_{\bbM}$ and $\Lm_{\bbB}$, respectively.
        The extended connective $\Lm_{\bbM}$ is defined by:
        \[
            \Lm_{\bbM}(a,b) =
            \begin{cases}
                \Lm_{\bbB}(a,b) & \text{for $a, b \in \bbB$}, \\
                0               & \text{otherwise}.
            \end{cases}
        \]
    \end{definition}
\end{mybox}

\begin{notation}
    Denote by $\bbL$ a logic set (e.g., $\bbB$ or $\bbM$), $\bbR$ the real set, $\bbZ$ the set of integers, $\bbN$ a numeric set (e.g., $\bbR$ or $\bbZ$), and $\bbD$ a certain set of $\bbL$ or $\bbN$.
\end{notation}

\begin{mybox}
    \begin{definition}\label{def:Real2Bool}
        For $x \in \bbN$, its logic value denoted by $x_{\logic}$ is given as $x_{\logic} = \true \Leftrightarrow x > 0$,
        $x_{\logic} = \false \Leftrightarrow x < 0$, and $x_{\logic} = 0 \Leftrightarrow x = 0$.
    \end{definition}
\end{mybox}

\begin{mybox}
    \begin{definition}
        The magnitude of a variable $x$, denoted by $|x|$, is defined as follows. If $x \in \bbN$, then $|x|$ is the standard absolute value. For $x \in \bbL$, the magnitude is given by:
        \[
            |x| =
            \begin{cases}
                0 & \text{if } x = 0, \\
                1 & \text{otherwise}.
            \end{cases}
        \]

    \end{definition}
\end{mybox}

\begin{mybox}
    \begin{definition}[Mixed-type logic]\label{def:MixedLogic}
        For $\Lm$ a logic connective of $\bbL$ and variables $a$, $b$, operation $c = \Lm(a, b)$ is defined such that $|c| = |a||b|$ and $c_{\logic} = \Lm(a_{\logic}, b_{\logic})$.
    \end{definition}
\end{mybox}

\subsection{Mathematical Foundation
 of Boolean Variation}

In this section, we present the mathematical foundation of Boolean variation which is the corner stone of the method for training Boolean weights directly within the Boolean domain, without relying on \gls{FP} latent weights \citep{nguyen2024bold}.

\subsubsection{Boolean Variation}

\begin{mybox}
    \begin{definition}\label{def:BoolOrder}
        Order relations `$<$' and `$>$' in $\bbB$ are defined as follows:
        \begin{equation}
            \false < \true, \quad \true > \false.
        \end{equation}
    \end{definition}

\end{mybox}

\begin{mybox}
    \begin{definition}\label{def:BoolVariation}
        For $a, b \in \bbB$, the variation from $a$ to $b$, denoted $\bvar(a \to b)$, is defined as: %
        \begin{equation}
            \bvar(a \to b) \triangleq \begin{cases}
                \true,  & \textrm{if } b > a, \\
                0,      & \textrm{if } b = a, \\
                \false, & \textrm{if } b < a.
            \end{cases}
        \end{equation}
    \end{definition}

\end{mybox}

\begin{mybox}
    \begin{definition}[Type conversion]\label{def:Conversion}
        Define:
        \begin{align}
            \proj \colon & \bbN \to \bbL \nonumber                              \\
                         & x \mapsto \proj(x) = \begin{cases}
                                                    \true,  & \textrm{if } x > 0, \\
                                                    0,      & \textrm{if } x = 0, \\
                                                    \false, & \textrm{if } x < 0.
                                                \end{cases}\label{eq:Projector}
        \end{align}
    \end{definition}
\end{mybox}

\begin{mybox}
    \begin{proposition} \citep{nguyen2023boolean,nguyen2024bold} \label{prop:Conversion}
        The following properties hold:
        \begin{enumerate}
            \item  $\forall x, y \in \bbN$: $\proj(xy) = \xnor(\proj(x), \proj(y))$.
            \item  $\forall a, b \in \bbL$: $\emb(\xnor(a,b)) = \emb(a)\emb(b)$.
            \item  $\forall x, y \in \bbN$: $x = y \Leftrightarrow |x| = |y| \textrm{ and } \proj(x) = \proj(y)$.
        \end{enumerate}
    \end{proposition}
\end{mybox}

In particular, property \cref{prop:Conversion}(2) implies that by the embedding map $\emb(\cdot)$, we have:
\begin{align}
    (\{\true, \false\}, \xor)  & \cong (\{\pm 1\}, -\times), \\
    (\{\true, \false\}, \xnor) & \cong (\{\pm 1\}, \times),
\end{align}
where $\cong$ and $\times$ stand for isomorphic relation, and the real multiplication, resp.
A consequence is that by $\emb(\cdot)$, a computing sequence of pointwise \textsc{xor} or \textsc{xnor}, counting, and majority vote is equivalent to a sequence of pointwise multiplications and accumulation performed on the embedded data.

\begin{mybox}
    \begin{proposition}\label{prop:XNORAlgebra}
        The following properties hold:
        \begin{enumerate}
            \item $a \in \bbL$, $x \in \bbN$: $\xnor(a,x) = \emb(a)x$.
            \item $x, y \in \bbN$: $\xnor(x,y) = xy$. %
            \item $x \in \{\bbL, \bbN\}$, $y, z \in \bbN$: $\xnor(x,y+z) = \xnor(x,y) + \xnor(x,z)$. %
            \item $x \in \{\bbL, \bbN\}$, $y, \lambda \in \bbN$: $\xnor(x, \lambda y) = \lambda \xnor(x,y)$.
            \item $x \in \{\bbL, \bbN\}$, $y \in \bbN$: $\xor(x,y) = -\xnor(x,y)$.
        \end{enumerate}
    \end{proposition}
\end{mybox}

\begin{proof}
    The proof follows definitions \ref{def:MixedLogic} and \ref{def:Conversion}.
    \begin{itemize}
        \item Following \cref{def:ThreeValueLogic} we have $\forall t \in \bbM$, $\xnor(\true, t) = t$, $\xnor(\false, t) = \neg t$, and $\xnor(0, t) = 0$. Put $v = \xnor(a, x)$. We have $|v| = |x|$ and $\proj(v) = \xnor(a, \proj(x))$. Hence, $a = 0 \Rightarrow \proj(v) = 0 \Rightarrow v = 0$; $a = \true \Rightarrow \proj(v) = \proj(x) \Rightarrow v = x$; $a = \false \Rightarrow \proj(v) = \neg \proj(x) \Rightarrow v = -x$. Hence (1).
        \item The result is trivial if $x=0$ or $y=0$. For $x, y \neq 0$, put $v = \xnor(x,y)$, we have $|v| = |x||y|$ and $\proj(v) = \xnor(\proj(x), \proj(y))$. According to \cref{def:Conversion}, if $\sign(x)=\sign(y)$, we have $\proj(v) = \true \Rightarrow v = |x||y| = xy$. Otherwise, i.e., $\sign(x)=-\sign(y)$, $\proj(v) = \false \Rightarrow v = -|x||y| = xy$. Hence (2).
        \item (3) and (4) follow (1) for $x \in \Lb$ and follow (2) for $x \in \bbN$.
        \item For (5), write $u = \xor(x,y)$ and $v = \xnor(x,y)$, we have $|u| = |v|$ and $\proj(u) = \xor(\proj(x), \proj(y)) = \neg \xnor(\proj(x), \proj(y)) = \neg \proj(v)$. Thus, $\sign(u) = -\sign(v) \Rightarrow u = -v$. \qedhere
    \end{itemize}
\end{proof}

\begin{notation}
    We denote $\cF(\mathbb{S},\mathbb{T})$  the set of all functions from source $\mathbb{S}$ to image $\mathbb{T}$.
\end{notation}

\begin{mybox}
    \begin{definition}\label{def:BoolFuncVar}
        For $f \in \cF(\bbB, \bbD)$, $\forall x \in \bbB$, write $\bvar f(x \to \neg x) := \bvar(f(x) \to f(\neg x))$.
        The variation of $f$ w.r.t. $x$, denoted $f'(x)$, is defined as: %
        \begin{equation*}
            f'(x) \triangleq \xnor(\bvar(x \to \neg x), \bvar f(x \to \neg x)).
        \end{equation*}
    \end{definition}

\end{mybox}

\begin{remark}
    For convenience and consistency of notation, we intentionally adopt the standard symbol for the continuous derivative, $f'$, to also denote Boolean variation
    The intended meaning --- whether it represents a continuous derivative or a Boolean variation --- can be inferred from the context in which the function $f$ is defined.
    Intuitively, the variation of $f$ w.r.t $x$ is $\true$ if $f$ varies in the same direction with $x$.
\end{remark}

\begin{example}\label{ex:XORVariation}
    Let $a \in \bbB$, $f(x) = \xor(x,a)$ for $x \in \bbB$, the variation of $f$ w.r.t. $x$ can be derived by establishing a truth table (see \cref{tab:VariationXOR}) from which we obtain $f'(x) = \neg a$.
\end{example}

{\renewcommand{\arraystretch}{1.0}
\begin{table}[H]
    \centering
    \caption{Variation truth table of $f(x) = \xor(a,x)$, $a, x \in \bbB$.}
    \label{tab:VariationXOR}
    \vspace{1ex}
    \begin{tabular}{cccccccc}
        \toprule %
        \multirow{2}{*}{$a$}
                 & \multirow{2}{*}{$x$}
                 & \multirow{2}{*}{$\neg x$}
                 & \multirow{2}{*}{$\bvar(x \to \neg x)$}
                 & \multirow{2}{*}{$f(a,x)$}
                 & \multirow{2}{*}{$f(a,\neg x)$}
                 & \multirow{2}{*}{$\bvar f(x \to \neg x)$}
                 & \multirow{2}{*}{$f'(x)$}                                                                                   \\
                 &                                          &          &          &          &                                \\
        \midrule %
        $\true$  & $\true$                                  & $\false$ & $\false$ & $\false$ & $\true$  & $\true$  & $\false$ \\

        $\true$  & $\false$                                 & $\true$  & $\true$  & $\true$  & $\false$ & $\false$ & $\false$ \\

        $\false$ & $\true$                                  & $\false$ & $\false$ & $\true$  & $\false$ & $\false$ & $\true$  \\

        $\false$ & $\false$                                 & $\true$  & $\true$  & $\false$ & $\true$  & $\true$  & $\true$  \\
        \bottomrule
    \end{tabular}
\end{table}
}

\subsubsection{Boolean Variation Calculus}

Below are some rules of Boolean variation which are necessary for training Boolean neural networks.

\begin{mybox}
    \begin{proposition} \citep{nguyen2023boolean,nguyen2024bold} \label{prop:B2BVariation}
        For $f, g \in \cF(\bbB, \bbB)$, $\forall x, y \in \bbB$ the following properties hold:
        \begin{enumerate}
            \item $\bvar f(x \to y) = \xnor(\bvar(x \to y), f'(x)).$
            \item $(\neg f(x))' = \neg f'(x)$.
            \item $(g \circ f)'(x) = \xnor(g'(f(x)), f'(x))$.
        \end{enumerate}
    \end{proposition}
\end{mybox}

\begin{proof} The proof is by definition:
    \begin{enumerate}
        \item $\forall x, y \in \bbB$, there are two cases. If $y = x$, then the result is trivial. Otherwise, i.e., $y = \neg x$, by definition we have:
              \begin{align*}
                  f'(x)                                       & = \xnor(\bvar(x \to \neg x), \bvar f(x \to \neg x)) \\
                  \Leftrightarrow \quad \bvar f(x \to \neg x) & = \xnor(\bvar(x \to \neg x), f'(x)).
              \end{align*}
              Hence the result.
        \item $\forall x, y \in \bbB$, it is easy to verify by truth table that $\bvar(\neg f(x \to y)) = \neg \bvar{f(x \to y)}$. Hence, by definition,
              \begin{align*}
                  (\neg f)'(x) & = \xnor(\bvar(x \to \neg x), \bvar(\neg f(x \to \neg x))) \\
                               & = \xnor(\bvar(x \to \neg x), \neg \bvar f(x \to \neg x))  \\
                               & = \neg \xnor(\bvar(x \to \neg x), \bvar f(x \to \neg x))  \\
                               & = \neg f'(x).
              \end{align*}
        \item Using definition, property (i), and associativity of $\xnor$, $\forall x \in \bbB$ we have:
              \begin{align*}
                  (g\circ f)'(x) & = \xnor(\bvar(x \to \neg x), \bvar g(f(x) \to f(\neg x)))                            \\
                                 & = \xnor\pren{\bvar(x \to \neg x), \xnor\pren{\bvar f(x \to \neg x), g'\pren{f(x)} }} \\
                                 & = \xnor\pren{g'(f(x)), \xnor\pren{\bvar(x \to \neg x), \bvar f(x \to \neg x) } }     \\
                                 & = \xnor(g'(f(x)), f'(x) ).
              \end{align*} \qedhere
    \end{enumerate}
\end{proof}

\begin{mybox}
    \begin{proposition}  \citep{nguyen2023boolean,nguyen2024bold} \label{prop:B2NVariation}
        For $f \in \cF(\bbB, \bbN)$, the following properties hold:
        \begin{enumerate}
            \item $x, y \in \bbB$: $\bvar f(x \to y) = \xnor(\bvar(x \to y), f'(x))$.
            \item $\alpha \in \bbN$: $(\alpha f)'(x) = \alpha f'(x)$.
            \item $g \in \cF(\bbB, \bbN)$: $(f + g)'(x) = f'(x) + g'(x)$.
        \end{enumerate}
    \end{proposition}
\end{mybox}

\begin{proof}
    The proof is as follows:
    \begin{enumerate}
        \item For $x, y \in \bbB$. Firstly, the result is trivial if $y = x$. For $y \neq x$, i.e., $y = \neg x$, by definition:
              \begin{equation*}
                  f'(x) = \xnor(\bvar(x \to \neg x), \bvar f(x \to \neg x)).
              \end{equation*}
              Hence, $|\bvar f(x \to \neg x)| = |f'(x)|$ since $|\bvar(x \to \neg x)| = 1$, and
              \begin{align*}
                  \proj(f'(x))                                       & = \xnor(\bvar(x \to \neg x), \proj(\bvar f(x \to \neg x))) \\
                  \Leftrightarrow \quad \proj(\bvar f(x \to \neg x)) & = \xnor(\bvar(x \to \neg x), \proj(f'(x))),
              \end{align*}
              where $\proj(\cdot)$ is the logic projector \cref{eq:Projector}. Thus, $\bvar f(x \to \neg x) = \xnor(\bvar(x \to \neg x), f'(x))$. Hence the result.
        \item Firstly $\forall x, y \in \bbB$, we have
              \begin{equation*}
                  \bvar(\alpha f(x \to y)) = \alpha f(y) - \alpha f(x) = \alpha \bvar{f(x \to y)}.
              \end{equation*}
              Hence, by definition,
              \begin{align*}
                  (\alpha f)'(x) & = \xnor(\bvar(x \to \neg x), \bvar(\alpha f(x \to \neg x)))                                                 \\
                                 & = \xnor(\bvar(x \to \neg x), \alpha\bvar{f(x \to \neg x)})                                                  \\
                                 & = \alpha \, \xnor(\bvar(x \to \neg x), \bvar{f(x \to \neg x)}), \textrm{ due to \cref{prop:XNORAlgebra}(4)} \\
                                 & = \alpha f'(x).
              \end{align*}
        \item For $f, g \in \cF(\bbB, \bbN)$,
              \begin{align*}
                  (f+g)'(x) & = \xnor(\bvar(x \to \neg x), \bvar(f+g)(x \to \neg x))                                                                  \\
                            & = \xnor(\bvar(x \to \neg x), \bvar f(x \to \neg x) + \bvar g(x \to \neg x))                                             \\
                            & \overset{(*)}{=} \xnor(\bvar(x \to \neg x), \bvar f(x \to \neg x)) + \xnor(\bvar(x \to \neg x), \bvar g(x \to \neg x)), \\
                            & = f'(x) + g'(x),
              \end{align*}
              where $(*)$ is due to \cref{prop:XNORAlgebra}(3). \qedhere
    \end{enumerate}
\end{proof}

For $f \in \cF(\bbZ, \bbN)$, its derivative, also known in terms of \emph{finite differences}, has been defined in the literature as $f'(x) = f(x+1) - f(x)$, see e.g. \cite{Jordan1950}.
With the logic variation as introduced above, we can make this definition more generic as follows.

\begin{definition}
    For $f \in \cF(\bbZ, \bbD)$, the variation of $f$ w.r.t $x \in \bbZ$ is defined as $f'(x) \triangleq \bvar f(x \to x+1)$,
    where $\bvar f$ is in the sense of the variation defined in $\bbD$.
\end{definition}

\begin{mybox}
    \begin{proposition}  \citep{nguyen2023boolean,nguyen2024bold} \label{prop:Composition}
        The following composition rules (chain rules) hold:
        \begin{enumerate}
            \item For $\bbB \overset{f}{\to} \bbB \overset{g}{\to} \bbD$: $(g \circ f)'(x) = \xnor(g'(f(x)), f'(x))$, $\forall x \in \bbB$.
            \item For $\bbB \overset{f}{\to} \bbZ \overset{g}{\to} \bbD$, $x \in \bbB$, if $|f'(x)| \leq 1$ and $g'(f(x)) =g'(f(x)-1)$, then:
                  \begin{equation*}
                      (g \circ f)'(x) = \xnor(g'(f(x)), f'(x)).
                  \end{equation*}
        \end{enumerate}
    \end{proposition}
\end{mybox}

\begin{proof}
    The proof is as follows.
    \begin{enumerate}
        \item The case of $\bbB \overset{f}{\to} \bbB \overset{g}{\to} \bbB$ is obtained from \cref{prop:B2BVariation}(3). For $\bbB \overset{f}{\to} \bbB \overset{g}{\to} \bbN$, by using \cref{prop:B2NVariation}(1), the proof is similar to that of \cref{prop:B2BVariation}(3).
        \item By definition, we have
              \begin{equation}\label{eq:proofComposition2_1}
                  (g \circ f)'(x) = \xnor(\bvar(x \to \neg x), \bvar g(f(x) \to f(\neg x))).
              \end{equation}
              Using property (1) of \cref{prop:B2NVariation}, we have:
              \begin{align}
                  f(\neg x) & = f(x) + \bvar f(x \to \neg x) \nonumber                                   \\
                            & = f(x) + \xnor(\bvar(x \to \neg x), f'(x)). \label{eq:proofComposition2_2}
              \end{align}
              Applying \cref{eq:proofComposition2_2} back to \cref{eq:proofComposition2_1}, the result is trivial if $f'(x) = 0$. The remaining case is $|f'(x)| = 1$ for which we have $\xnor(\bvar(x \to \neg x), f'(x)) = \pm 1$. First, for $\xnor(\bvar(x \to \neg x), f'(x)) = 1$, we have:
              \begin{align}
                  \bvar g(f(x) \to f(\neg x)) & = \bvar g(f(x) \to f(x) + 1) \nonumber                                                \\
                                              & = g'(f(x)) \nonumber                                                                  \\
                                              & = \xnor(g'(f(x)), 1) \nonumber                                                        \\
                                              & = \xnor(g'(f(x)), \xnor(\bvar(x \to \neg x), f'(x)) ) \label{eq:proofComposition2_3}.
              \end{align}
              Substitute \cref{eq:proofComposition2_3} back to \cref{eq:proofComposition2_1}, we obtain:
              \begin{align*}
                  (g \circ f)'(x) & = \xnor(\bvar(x \to \neg x), \bvar g(f(x) \to f(\neg x)))                          \\
                                  & = \xnor(\bvar(x \to \neg x), \xnor(g'(f(x)), \xnor(\bvar(x \to \neg x), f'(x)) ) ) \\
                                  & = \xnor(g'(f(x)), f'(x)),
              \end{align*}
              where that last equality is by the associativity of $\xnor$ and that $\xnor(x, x) = \True$ for $x \in \bbB$.
              Similarly, for $\xnor(\bvar(x \to \neg x), f'(x)) = -1$, we have:
              \begin{align}
                  \bvar g(f(x) \to f(\neg x)) & = \bvar g(f(x) \to f(x) - 1) \nonumber                                                  \\
                                              & = - g'(f(x)-1) \nonumber                                                                \\
                                              & = \xnor(g'(f(x)-1), -1) \nonumber                                                       \\
                                              & = \xnor(g'(f(x)-1), \xnor(\bvar(x \to \neg x), f'(x)) ) \label{eq:proofComposition2_4}.
              \end{align}
              Substitute \cref{eq:proofComposition2_4} back to \cref{eq:proofComposition2_1} and use the assumption that $g'(f(x)) = g'(f(x)-1)$, we have:
              \begin{align*}
                  (g \circ f)'(x) & = \xnor(\bvar(x \to \neg x), \bvar g(f(x) \to f(\neg x)))                            \\
                                  & = \xnor(\bvar(x \to \neg x), \xnor(g'(f(x)-1), \xnor(\bvar(x \to \neg x), f'(x)) ) ) \\
                                  & = \xnor(g'(f(x)), f'(x)).
              \end{align*}
              Hence the preposition is proved. \qedhere
    \end{enumerate}
\end{proof}

\begin{example}\label{ex:XNORVariation}
    From \cref{ex:XORVariation}, we have $\bvar{\xor(x,a)}/\bvar{x} = \neg a$ for $a, x \in \bbB$.
    Using \cref{prop:B2BVariation}-(2) we have: $\bvar{\xnor(x,a)}/\bvar{x} = a$ since $\xnor(x,a) = \neg \xor(x,a)$.
\end{example}

\subsubsection{Multivariate Case}
The properties of Boolean variation described above can be extended to the multivariate case in a straightforward manner.
For example, in the case of multivariate Boolean functions, the extension is as follows.

\begin{mybox}
    \begin{definition}\label{def:BoolFuncVar0}
        For $\mbx = (x_1, \ldots, x_n) \in \bbB^n$, denote $\mbx_{\neg i} \triangleq (x_1, \ldots, x_{i-1}, \neg x_i, x_{i+1}, \ldots, x_n)$ for $n \ge 1$ and $1 \leq i \leq n$.
        For $f \in \cF(\bbB^n, \bbB)$, the (partial) variation of $f$ w.r.t. $x_i$, denoted $f'_{i}(\mbx)$ or $\bvar f(\mbx)/\bvar x_i$, is defined as: $f'_{i}(\mbx) \equiv \bvar f(\mbx)/\bvar x_i \triangleq \xnor(\bvar(x_i \to \neg x_i), \bvar f(\mbx \to \mbx{\neg i}))$.
    \end{definition}
\end{mybox}

The composition rule then becomes:

\begin{mybox}
    \begin{proposition}  \citep{nguyen2024bold} \label{prop:MultiVariate}
        Let $f \in \cF(\bbB^n, \bbB)$, $n \geq 1$, and $g \in \cF(\bbB, \bbB)$. For $1 \le i \le n$:
        \begin{equation}
            (g \circ f)'_i(\mbx) = \xnor(g'(f(\mbx)), f'_i(\mbx)), \quad \forall \mbx \in \bbB^n.
        \end{equation}
    \end{proposition}
\end{mybox}

\begin{example}\label{ex:PreActVariation}
    Apply \cref{prop:B2NVariation}-(3) to $\mbY_{[k,j]}^{(l)}$ from \cref{eq:Preactivation}: $\bvar{\mbY_{[k,j]}^{(l)}}/\bvar{\mbW^{(l)}_{[i,j]}} = \bvar{\Lm(\mbX^{(l)}_{[k,i]}, \mbW^{(l)}_{[i,j]})}/\bvar{\mbW^{(l)}_{[i,j]}}$ and $\bvar{\mbY_{[k,j]}^{(l)}}/\bvar{\mbX^{(l)}_{[k,i]}} = \bvar{\Lm(\mbX^{(l)}_{[k,i]}, \mbW^{(l)}_{[i,j]})}/\bvar{\mbX^{(l)}_{[k,i]}}$.
    Then, for $\Lm = \xnor$ as an example, we have: $\bvar{\mbY_{[k,j]}^{(l)}}/\bvar{\mbW^{(l)}_{[i,j]}} = \mbX^{(l)}_{[k,i]}$ and $\bvar{\mbY_{[k,j]}^{(l)}}/\bvar{\mbX^{(l)}_{[k,i]}} = \mbW^{(l)}_{[i,j]}$.
\end{example}

\subsection{Boolean Backpropagation}

This section presents how to apply the above principles of Boolean variation to define backpropagation for Boolean neural networks.
The $l$-th layer (\cref{eq:Preactivation}), receives the backpropagation signal from the downstream layer $l+1$.
Specifically, $\mbZ^{(l)}_{[k,j]} \triangleq \frac{\delta \cL}{\delta \mbY_{[k,j]}^{(l)}}$ denotes the variation of the loss function $\cL$ w.r.t. the output at layer $l$.
To optimize the Boolean weights, we need to compute the corresponding loss signal, denoted as $\mbQ^{(l)}_{[i,j]} \triangleq \frac{\delta \cL}{\delta \mbW_{[i,j]}^{(l)}}$.
In addition, we also have to compute the loss signal for the upstream layer, defined as $\mbP^{(l)}_{[k,i]} \triangleq \frac{\delta \cL}{\delta \mbX_{[k,i]}^{(l)}}$.
Hereafter, we consider the logic gate $\text{L} = \xnor$ as a concrete example.

First, using \cref{prop:B2BVariation}, \cref{prop:B2NVariation}, \cref{prop:Composition} and its extension to the multivariate case by \cref{prop:MultiVariate} in the same manner as shown in \cref{ex:PreActVariation}, we have:

\begin{align}
    \frac{\delta \mbY_{[k,j]}^{(l)}}{\delta \mbW_{[i,j]}^{(l)}} & = \frac{\delta \xnor(\mbX_{[k,i]}^{(l)}, \mbW_{[i,j]}^{(l)}) }{\delta \mbW_{[i,j]}^{(l)}} = \mbX_{[k,i]}^{(l)} \\
    \frac{\delta \mbY_{[k,j]}^{(l)}}{\delta \mbX_{[k,i]}^{(l)}} & = \frac{\delta \xnor(\mbX_{[k,i]}^{(l)}, \mbW_{[i,j]}^{(l)}) }{\delta \mbX_{[k,i]}^{(l)}} = \mbW_{[i,j]}^{(l)}
\end{align}

Using the chain rules given by \cref{prop:Composition}, we have the following atomic variations:
\begin{align}
    \mbQ^{(l)}_{[k,i,j]} \triangleq \frac{\delta \cL}{\delta \mbW_{[i,j]}^{(l)}} |_{k} & = \xnor \left( \frac{\delta \cL}{\delta \mbY_{[k,j]}^{(l)}}, \frac{\delta \mbY_{[k,j]}^{(l)}}{\delta \mbW_{[i,j]}^{(l)}} \right) = \xnor \left( \mbZ_{[k,j]}^{(l)}, \mbX_{[k,i]}^{(l)} \right), \\
    \mbP^{(l)}_{[k,i,j]} \triangleq \frac{\delta \cL}{\delta \mbX_{[k,i]}^{(l)}} |_{j} & = \xnor \left( \frac{\delta \cL}{\delta \mbY_{[k,j]}^{(l)}}, \frac{\delta \mbY_{[k,j]}^{(l)}}{\delta \mbX_{[k,i]}^{(l)}} \right) = \xnor \left( \mbZ_{[k,j]}^{(l)}, \mbW_{[i,j]}^{(l)} \right).
\end{align}

The variations $\mbQ^{(l)}_{[i,j]}$ and $\mbG^{(l)}_{[k,i]}$ can be then  obtained by aggregating the above atomic variations over the batch dimension $k$ and output dimension $j$, respectively.
More specifically, denote $\mathbf{1}(\cdot)$ the indicator function.
Additionally, for $b \in \bbB$ and a variable $x$, we define $\mathbf{1}(x = b) = 1$ if $x_{\logic} = b$ and $\mathbf{1}(x = b) = 0$ otherwise.
Then, we have:
\begin{align}
    \mbQ^{(l)}_{[i,j]} \triangleq \frac{\delta \cL}{\delta \mbW_{[i,j]}^{(l)}} & = \sum_{k} \mathbf{1} ( \mbQ_{[k,i,j]}^{(l)} = \textsc{true} ) |\mbQ_{[k,i,j]}^{(l)}| - \sum_{k} \mathbf{1} ( \mbQ_{[k,i,j]}^{(l)} = \textsc{false} ) |\mbQ_{[k,i,j]}^{(l)}|, \label{eq:AggrWeightVariation} \\
    \mbP^{(l)}_{[i,j]} \triangleq \frac{\delta \cL}{\delta \mbX_{[k,i]}^{(l)}} & = \sum_{j} \mathbf{1} ( \mbP_{[k,i,j]}^{(l)} = \textsc{true} ) |\mbP_{[k,i,j]}^{(l)}| - \sum_{j} \mathbf{1} ( \mbP_{[k,i,j]}^{(l)} = \textsc{false} ) |\mbP_{[k,i,j]}^{(l)}|. \label{eq:AggrBprop}
\end{align}

\subsection{Boolean Optimizer}

\begin{algorithm}[H]
	\caption{Boolean learning process for a linear layer.}
	\label{algo:BooleanTraining}
	\SetKwInOut{Input}{Input}
	\SetKwInOut{Output}{Output}
	\SetKwBlock{Loop}{Loop}{end}
	\SetKwInOut{Initialize}{Initialize}
	\SetKwFor{When}{When}{do}{end}
	\SetKwFunction{Wait}{Wait}
	\SetAlgoLined
	\SetNoFillComment
	\Input{
		Learning rate $\eta$, number of iterations $T$\;
	}
	\Initialize{	
		$\mbM_{[i,j]}^{(l),0} = 0$; $\beta^0 = 1$\;
	}
	\For{$t = 0,\dots, T-1$}{
		\tcc{\textbf{1. Forward}}
		Compute $\mbY^{(l),t}$ following \cref{eq:Preactivation}\; %
		\tcc{\textbf{2. Backward}}
		Receive $\frac{\bvar{\cL}}{\bvar{\mbY_{[k,j]}^{(l),t}}}$ from downstream layer\;
		\tcc{\textbf{2.1  Backpropagation}}
		Compute and backpropagate $\mbP^{(l),t}$ to the upstream following \cref{eq:AggrBprop}\; %
		\tcc{\textbf{2.2  Weight update process}}
		$N_{\textrm{total}} := 0$, $N_{\textrm{unchanged}} := 0$\;
		\ForEach{$\mbW_{i,j}^{l}$}{
			Compute $\mbQ_{[i,j]}^{(l),t+1}$ following \cref{eq:AggrWeightVariation}\;  %
			Update $\mbM_{[i,j]}^{(l),t+1} = \beta^t \mbM_{[i,j]}^{(l),t} + \eta^t \mbQ_{[i,j]}^{(l),t+1}$\;
			$N_{\textrm{total}} \gets N_{\textrm{total}} + 1$\;
			\eIf{$\xnor(\mbM_{[i,j]}^{(l),t+1}, \mbW_{[i,j]}^{(l),t}) = \true$}{ %
                \tcc{\textbf{Flip weight}}
				$\mbW_{[i,j]}^{(l),t+1} = \neg \mbW_{[i,j]}^{(l),t}$\;
                \tcc{\textbf{Reset corresponding accumulator}}
				$\mbM_{[i,j]}^{(l),t+1} = 0$\;
			}{
                \tcc{\textbf{Weight is unchanged}}
				$\mbW_{[i,j]}^{(l),t+1} = \mbW_{[i,j]}^{(l),t}$\;
                \tcc{\textbf{Update statistics to update $\beta$}}
				$N_{\textrm{unchanged}} \gets N_{\textrm{unchanged}} + 1$\;
			}%
		}
		Update $\eta^{t+1}$, $\beta^{t+1} = N_{\textrm{unchanged}}/N_{\textrm{total}}$ \;
	}
\end{algorithm}

Given the above variations, the rule for updating the Boolean weight $\mbW_{[i,j]}^{(l)}$ to minimize the loss function $\cL$ is as follows:
\begin{align}
    \mbW_{[i,j]}^{(l)} = \lnot \mbW_{[i,j]}^{(l)} \quad \text{if } \mathbf{xnor} \left(\mbQ^{(l)}_{[i,j]}, \mbW^{(l)}_{[i,j]} \right) = \textsc{true}.
\end{align}

Based on this update rule, we can develop an optimizer that accumulates the signal $\mbQ_{[i,j]}^{(l)}$ over training iterations.
Specifically, let $\mbW_{[i,j]}^{(l),t}$ denotes the weight at iteration $t$, and $\mbM_{[i,j]}^{(l),t}$ represents its accumulator, initialized as $\mbM_{[i,j]}^{(l),0} = 0$.
The update rule for the accumulator is then defined as:
The update rule for the accumulator is then defined as:
\begin{align}
    \mbM_{[i,j]}^{(l),t+1} \leftarrow \beta^{t} \mbM_{[i,j]}^{(l),t} + \eta \mbQ_{[i,j]}^{(l),t}, \label{eq:bool_optimizer}
\end{align}
where $\eta$ is the accumulation factor acting as a learning rate, and $\beta^{t}$ is an auto-regularizing factor that reflects the system's state at time $t$.
In our work, we use brain plasticity \citep{Fuchs2014} and Hebbian theory \citep{Hebb2005} to adaptively set $\beta^{t}$, that force the weights to adapt to their neighborhood during.
For the chose weight's neighborhood, for instance, neuron, layer, or network level, $\beta^t$ is set as:
\begin{equation}
    \beta^t = \frac{\textrm{Number of unchanged weights at } t}{\textrm{Total number of weights}}.
\end{equation}
It to temper the importance of weight variational according to how much neurons have changed.
In our experiments, $\beta^t$ is set to per-layer basis and initialized as $\beta^0 = 1$
The learning process for a linear layer is described in \cref{algo:BooleanTraining}.

\section{Discussion on Hardware Considerations} \label{sec:hardware_discussion}

\subsection{Computation Proposed in \cref{sec:bool_linear}}

The Boolean framework supports both full and partial binary settings. The afforementioned Boolean variation calculus shows that:
\begin{align}
    \xnor (x_{\textrm{real}}, w_{\textrm{logic}}) = x_{\textrm{real}} \times w_{\textrm{binary}},
\end{align}
under the mapping $\textsc{true} \rightarrow +1$ and $\textsc{false} \rightarrow -1$. 
Consequently, matrix multiplication ($\mathbf{matmul}$) between a real tensor $\mbX$ and a logic tensor $\mbW$ can be implemented as follows:
\begin{itemize}
    \item \textbf{Using binary weights} $\{-1, +1\}$: Simply represent the logic weights in binary format.
    Then, $\mathbf{matmul} (x_{\textrm{real}}, w_{\textrm{logic}})$ is directly computed as $\mathbf{matmul} (x_{\textrm{real}}, w_{\textrm{binary}})$.
    \item \textbf{Using native logic} $\{\textsc{true}, \textsc{false}\}$: The multiplication reduces to:
    \begin{equation}
        \mathbf{matmul} (x_{\textrm{real}}, w_{\textrm{logic}}) =
            \begin{cases}
            x_{\textrm{real}}, & \text{if } w_{\textrm{logic}} = \textsc{true}  \\
            -x_{\textrm{real}}, & \text{if } w_{\textrm{logic}} = \textsc{false}
            \end{cases} \label{eq:matmul_real_bool}
    \end{equation}
    Thus, a sign flip of $x_{\textrm{real}}$ conditioned on $ w_{\textrm{logic}}$, followed by accumulation, suffices to perform $\mathbf{matmul}(\mbX_{\textrm{real}}, \mbW_{\textrm{logic}})$.
\end{itemize}

The first approach is well-supported by modern hardware such as CPUs, GPUs, etc, where different bit-widths can be used to represent and simulate weight values in $\{-1, +1\}$.
Additionally, this approach can be implemented directly in PyTorch \citep{Paszke2019}.
The second approach, in contrast, requires a specialized Boolean accelerator. Such hardware can massively accelerate the computation by directly leveraging logic operations instead of real-arithmetic.

\subsection{Multi-core Computation Strategy in \cref{sec:multi_kernel_boolean}}

Boolean design, as used in the paper, employs Boolean weights and operates using logic operations. It is distinct from bit-level operations.

\paragraph{Boolean design:} Weights are Boolean logic variables, taking values $\textsc{true}/\textsc{false}$ or $-1/+1$.
Operations are logic-based, such as $\mathbf{xnor}$, and $\mathbf{or}$, etc. See \cref{eq:matmul_real_bool} for an example.

\paragraph{Bit-level operations:} 
These, such as bit-serial implementations in C/C++, operate bit-by-bit on multi-bit variables. For instance, a bit-level AND between two $n$-bit variables produces an $n$-bit result, where each bit is the ADN of corresponding pair of bits from the inputs.
Bit-level operations like bit-serial are inefficient in terms of latency, whereas Boolean logic operations are significantly faster compared to real-arithmetic operations such as multiplication.

\newpage
\section{Code Samples of Core Implementation}

\subsection{Boolean Linear Layer and Optimizer}

In this section, we provide example Python code for implementing a Boolean linear layer based on the $\xor$ logic gate.
This implementation is based on the PyTorch framework \citep{Paszke2019}.
As done in \cite{nguyen2024bold}, the class definition for the Boolean linear layer is presented in \cref{alg:code_xorlinear}, and its backpropagation mechanism—customized via PyTorch's \texttt{autograd} system—is detailed in \cref{alg:code_xorfunction}.
Each Boolean kernel is primarily implemented using this Boolean linear layer.

We consider both cases of the incoming backpropagation signal: Boolean-valued (see \cref{alg:code_bpropbool}), and real-valued (see \cref{alg:code_bpropreal}).
The latter is the main use case in this paper.
An example implementation of the Boolean optimizer used to update the layer's parameters is provided in \cref{alg:code_optim}.

\begin{algorithm}[H]
    \centering
    \caption{Python code of \textsc{xor} linear layer}\label{alg:code_xorlinear}
\begin{lstlisting}[language=Python]
import torch

from torch import Tensor, nn, autograd
from typing import Any, List, Optional, Callable


class XORLinear(nn.Linear):

    def __init__(self, in_features: int, out_features: int, bool_bprop: bool, **kwargs):
        super(XORLinear, self).__init__(in_features, out_features, **kwargs)        
        self.bool_bprop = bool_bprop
    
    def reset_parameters(self):
        self.weight = nn.Parameter(torch.randint(0, 2, self.weight.shape))
        
        if self.bias is not None:
            self.bias = nn.Parameter(torch.randint(0, 2, (self.out_features,)))
    
    def forward(self, X):
        return XORFunction.apply(X, self.weight, self.bias, self.bool_bprop)
\end{lstlisting}
\end{algorithm}

\begin{algorithm}[H]
    \centering
    \caption{Python code of the backpropagation logic of \textsc{xor} linear layer}\label{alg:code_xorfunction}
\begin{lstlisting}[language=Python]
class XORFunction(autograd.Function):

    @staticmethod
    def forward(ctx, X, W, B, bool_bprop: bool):
        ctx.save_for_backward(X,W,B)
        ctx.bool_bprop = bool_bprop      
          
        # Elementwise XOR logic
        S = torch.logical_xor(X[:,None,:], W[None,:,:])   
             
        # Sum over the input dimension
        S = S.sum(dim=2) + B
        
        # 0-centered for use with BatchNorm when preferred
        S = S - W.shape[1]/2
        
        return S
    
    @staticmethod
    def backward(ctx, Z):
        if ctx.bool_bprop:
            G_X, G_W, G_B = backward_bool(ctx, Z)
        else:
            G_X, G_W, G_B = backward_real(ctx, Z)

        return G_X, G_W, G_B, None
\end{lstlisting}
\end{algorithm}

\newpage

\begin{algorithm}[H]
    \centering
    \caption{Backpropagation logic with Boolean received backpropagation}\label{alg:code_bpropbool}
\begin{lstlisting}[language=Python]
def backward_bool(ctx, Z):
    """
    Variation of input:
        - delta(xor(x,w))/delta(x) = neg w
        - delta(Loss)/delta(x) = xnor(z,neg w) = xor(z,w)
    Variation of weights:
        - delta(xor(x,w))/delta(w) = neg x
        - delta(Loss)/delta(x) = xnor(z,neg x) = xor(z,x)
    Variation of bias:
        - bias = xnor(bias,True) ==> Variation of bias is driven in 
          the same basis as that of weight with xnor logic and input True.
    Aggregation:
        - Count the number of TRUEs = sum over the Boolean data
        - Aggr = TRUEs - FALSEs = TRUEs - (TOT - TRUEs) = 2TRUES - TOT
          where TOT is the size of the aggregated dimension
    """
    X, W, B = ctx.saved_tensors
    
    # Boolean variation of input
    G_X = torch.logical_xor(Z[:,:,None], W[None,:,:])
    
    # Aggregate over the out_features dimension
    G_X = 2 * G_X.sum(dim=1) - W.shape[0]   
    
    # Boolean variation of weights
    G_W = torch.logical_xor(Z[:,:,None], X[:,None,:])
    
    # Aggregate over the batch dimension
    G_W = 2 * G_W.sum(dim=0) - X.shape[0]
    
    # Boolean variation of bias
    if B is not None:
        # Aggregate over the batch dimension
        G_B = 2 * Z.sum(dim=0) - Z.shape[0]

    # Return
    return G_X, G_W, G_B
\end{lstlisting}
\end{algorithm}

\begin{algorithm}[H]
    \centering
    \caption{Backpropagation logic with real received backpropagation}\label{alg:code_bpropreal}
\begin{lstlisting}[language=Python]
def backward_real(ctx, Z):
    X, W, B = ctx.saved_tensors

    """
    Boolean variation of input processed using torch avoiding loop:
        -> xor(Z: Real, W: Boolean) = -Z * emb(W)
        -> emb(W): T->1, F->-1 => emb(W) = 2W-1
        => delta(Loss)/delta(X) = Z*(1-2W) """
    G_X = Z.mm(1-2*W)
    
    """ 
    Boolean variation of weights processed using torch avoiding loop:
        -> xor(Z: Real, X: Boolean) = -Z * emb(X)
        -> emb(X): T->1, F->-1 => emb(X) = 2X-1
        => delta(Loss)/delta(W) = Z^T * (1-2X) """
    G_W = Z.t().mm(1-2*X)
    
    """ Boolean variation of bias """
    if B is not None:
        G_B = Z.sum(dim=0)
    
    # Return
    return G_X, G_W, G_B
\end{lstlisting}
\end{algorithm}

\newpage

\begin{algorithm}[H]
    \centering
    \caption{Python code of Boolean optimizer}\label{alg:code_optim}
\begin{lstlisting}[language=Python]
class BooleanOptimizer(torch.optim.Optimizer):
    
    def __init__(self, params, lr: float):
        super(BooleanOptimizer, self).__init__(params, dict(lr=lr))
        for param_group in self.param_groups:
            param_group['accums'] = [torch.zeros_like(p.data) for p in param_group['params']]
            param_group['ratios'] = [0 for p in param_group['params']]
        self._nb_flips = 0

    @property
    def nb_flips(self):
        n = self._nb_flips
        self._nb_flips = 0
        return n

    def step(self):
        for param_group in self.param_groups:
            for idx, p in enumerate(param_group['params']):
                self.update(p, param_group, idx)

    def update(self, param: Tensor, param_group: dict, idx: int):
        accum = param_group['ratios'][idx] * param_group['accums'][idx] + param_group['lr'] * param.grad.data
        param_group['accums'][idx] = accum
        param_to_flip = accum * (2*param.data-1) >= 1
        param.data[param_to_flip] = torch.logical_not(param.data[param_to_flip])
        param_group['accums'][idx][param_to_flip] = 0.
        param_group['ratios'][idx] = 1 - param_to_flip.float().mean()
        self._nb_flips += float(param_to_flip.float().sum())
\end{lstlisting}
\end{algorithm}

\subsection{Successive SVID for Kernel Extraction}

\cref{alg:code_svid} illustrate the Python code of the \gls{SVID} algortithm to extract the optimal Boolean weights and scaling factors for one kernel.
Based on this, \cref{alg:code_succesive_svid} illustrates the succesive \gls{SVID} algorithm to extract all kernels.

\begin{algorithm}[H]
    \centering
    \caption{Python code of \gls{SVID} approximation of a \gls{FP} matrix. \label{alg:code_svid}}
    \UseRawInputEncoding
\begin{lstlisting}[language=Python]
def svid_approximation(w):
    """
    Approximate the input matrix `w` by a boolean matrix and a rank-1 matrix:
        w = w_bool * (s_out * s_in.T)
    
    Args:
        w (torch.Tensor): Input tensor of shape (*, m, n).

    Returns:
        tuple: 
            - w_bool (torch.Tensor): Boolean matrix of the same shape as `w`.
            - w_res (torch.Tensor): Residual matrix, w - w_bool * (s_out * s_in.T).
            - s_in (torch.Tensor): Scaled first left singular vector of `w`.
            - s_out (torch.Tensor): Scaled first right singular vector of `w`.
    """
    U, S, Vh = torch.linalg.svd(abs(w.data.clone().float()), full_matrices=False)

    w_bool = torch.sign(w)
    s_in = torch.sqrt(S[0]) * Vh[0,:].reshape(1,-1)
    s_out = torch.sqrt(S[0]) * U[:,0].reshape(-1,1)

    w_res = w - w_bool * torch.matmul(s_out, s_in)

    return w_bool, w_res, s_in, s_out
\end{lstlisting}
\end{algorithm}

\begin{algorithm}[H]
    \centering
    \caption{Python code of successively extracts kernels from \gls{FP} matrix using \gls{SVID}.  \label{alg:code_succesive_svid}}
    \UseRawInputEncoding
\begin{lstlisting}[language=Python]
def successive_svid(w_fp, n_kernels):
    """
    Perform successive SVID on the input matrix to extract Boolean kernels.

    Args:
        w_fp (torch.Tensor): Input weight matrix.
        n_kernels (int): Number of iterations to extract kernels.

    Returns:
        list: List of dictionaries containing `n_kernels` kernels, each has:
            - w_bool (torch.Tensor): Boolean matrix.
            - s_in (torch.Tensor): Input scaling vector.
            - s_out (torch.Tensor): Output scaling vector.
    """
    boolean_kernels = []
    
    w = w_fp # The input to SVID at first iteration is the original weight
    
    for k in range(n_kernels):
        # Extract the Boolean weights, residual, and scaling vectors
        w_bool, w_res, s_in, s_out = svid_approximation(w)
        
        # Save the extracted kernel
        boolean_kernels.append({'w_bool': w_bool, 's_in': s_in, 's_out': s_out})
        
        # The input to SVID for the next iteration is the current residual matrix
        w = w_res
        
    return boolean_kernels
\end{lstlisting}
\end{algorithm}

\section{Proof of Propositions}

For completeness, we include the proofs of Propositions related to \gls{SVID} approximation used in the main paper.

\subsection{Proof of Boolean Linear Reformulation using SVID}

\begin{mybox}
    \begin{proposition}\citep{xu2024onebit}
        Given the weight matrix $\mbW_{\mathrm{FP}}$ and input $\mbX$, the linear layer can be reformulated as the following using \gls{SVID} approximation, $\mbW_{\mathrm{FP}} \approx \mbW_{\mathrm{bool}} \odot \left( \mbs_{\mathrm{out}} \mbs_{\mathrm{in}}^{\top} \right)$, as follows:
        \begin{align}
            \mbX \mbW_{\mathrm{FP}}^{\top} \approx \left[ \left( \mbX \odot \mbs_{\mathrm{in}}^{\top} \right) \wbool^{\top} \right] \odot \mbs_{\mathrm{out}}^{\top}.
        \end{align}
    \end{proposition}
\end{mybox}

\begin{proof}
    Due to the \gls{SVID} approximation, we have $\mbW_{\mathrm{FP}[i,j]} \approx \mbW_{\mathrm{bool}[i,j]} \mbs_{\uout[i]}\mbs_{\uin[j]}$.
    Then, we have:
    \begin{align}
        \left( \mbX \wfp^{\top} \right)_{[i,j]} & \approx \sum_{k} \mbX_{[i,k]} \mbW_{\ufp[k,j]}^{\top}                                                                \\
                                                & = \sum_{k} \mbX_{[i,k]} \mbW_{\ufp[j,k]}                                                                             \\
                                                & = \sum_{k} \mbX_{[i,k]} \mbW_{\mathrm{bool}[j,k]} \mbs_{\uout[j]}\mbs_{\uin[k]}                                      \\
                                                & = \sum_{k} \mbX_{[i,k]} \mbs_{\uin[k]} \mbW_{\mathrm{bool}[j,k]} \mbs_{\uout[j]}                                     \\
                                                & = \sum_{k} \left( \mbX \odot \ssin^{\top} \right)_{[i,k]} \mbW_{\mathrm{bool}[k,j]}^{\top} \mbs_{\uout[j]}           \\
                                                & = \left[ \left( \mbX \odot \ssin^{\top} \right) \wbool^{\top} \right]_{[i,j]} \mbs_{\uout[j]}                        \\
                                                & = \left\{  \left[ \left( \mbX \odot \ssin^{\top} \right) \wbool^{\top} \right] \odot \ssout^{\top} \right\}_{[i,j]}.
    \end{align}

    Thus, the proposition is proved. \qedhere

\end{proof}

\subsection{Proof of \cref{pro:svid}}

\begin{mybox}
    \begin{lemma}\citep{xu2024onebit} \label{lemma:biggest_singular}
        Denote $\sigma_i(\mbW)$ the $i$-th biggest singular value of matrix $\mbW$. The following inequality holds:
        \begin{align}
            \sigma_1(|\mbW|) \geq \sigma_1(\mbW).
        \end{align}
    \end{lemma}
\end{mybox}

\begin{proof}
    By the definition of induced norm, we have:
    \begin{align}
        \sigma_1(\mbW)   & = \|\mbW\|_2 = \max_{\mbx, \|\mbx\|_2 = 1} \|\mbW \mbx\|_2,     \\
        \sigma_1(|\mbW|) & = \||\mbW\||_2 = \max_{\mby, \|\mby\|_2 = 1} \||\mbW| \mby\|_2.
    \end{align}
    In addition, because $\forall \mbx, \|\mbx\|_2 = 1$, we have:
    \begin{align}
        \||\mbW| |\mbx|\|_2^2 & = \sum_i \left( \sum_j |\mbW_{[i,j]}| |\mbx_{[j]}| \right)^2   \\
                              & \geq \sum_i \left( |\sum_j \mbW_{[i,j]} \mbx_{[j]} | \right)^2 \\
                              & = \sum_i \left( \sum_j  \mbW_{[i,j]} \mbx_{[j]} \right)^2      \\
                              & = \|\mbW \mbx\|_2^2.
    \end{align}

    Therefore
    \begin{align}
        \max_{\mby, \|\mby\|_2 = 1} \||\mbW| \mby\|_2 & \geq \max_{\mbx, \|\mbx\|_2 = 1} \|\mbW \mbx\|_2 \\
        \Leftrightarrow \sigma_1(|\mbW|)              & \geq \sigma_1(\mbW).
    \end{align}

    Thus, the lemma is proved. \qedhere
\end{proof}

\begin{mybox}
    \begin{proposition}[Restated from \cite{xu2024onebit}]
        For $\mbW \in \mathbb{R}^{m\times n}$, write $\mbW = \widetilde{\mbU} \widetilde{\mbSigma} \widetilde{\mbV}^{\top}$ its \gls{SVD}.
        Let $\mba = \sqrt{\tilde{\sigma}_1} \widetilde{\mbU}_{[:,1]}$, and $\mbb = \sqrt{\tilde{\sigma}_1} \widetilde{\mbV}_{[:,1]}$.
        Similarly, denote $|\mbW| = \mbU \mbSigma \mbV^{\top}$ its \gls{SVD}; $\mbs_{\mathrm{in}}$ and $\mbs_{\mathrm{out}}$ are given as: $\mbs_{\mathrm{in}} = \sqrt{\sigma_1} \mbV_{[:,1]}$, and $\mbs_{\mathrm{out}} = \sqrt{\sigma_1} \mbU_{[:,1]}$.
        We decompose the matrix as $\mbW = \mbW_{\mathrm{bool}} \odot |\mbW| \approx \mbW_{\mathrm{bool}} \odot \left( \mbs_{\mathrm{out}} \mbs_{\mathrm{in}}^{\top} \right)$.
        We then have:

        \begin{align}
            \left\| \mbW - \mbW_{\mathrm{bool}} \odot \mbs_{\mathrm{out}} \mbs_{\mathrm{in}}^{\top}   \right\|_{F}^{2} \leq \left\| \mbW - \mba \mbb^{\top} \right\|_{F}^{2}.
        \end{align}
    \end{proposition}
\end{mybox}

\begin{proof}
    We denote the following error matrices:
    \begin{align}
        \mbE_1 & = \mbW - \mba \mbb^{\top},                         \\
        \mbE_2 & = |\mbW| - \ssout \ssin^{\top}. \label{eq:error_2}
    \end{align}

    Multiplying $\wbool$ with both sides of \cref{eq:error_2}, we have:
    \begin{align}
        \wbool \odot |\mbW| - \wbool \odot \ssout \ssin^{\top}  & = \wbool \odot \mbE_2  \\
        \Leftrightarrow \mbW - \wbool \odot \ssout \ssin^{\top} & = \wbool \odot \mbE_2.
    \end{align}

    Thus, we have:
    \begin{align}
        \| \mbW - \wbool \odot \ssout \ssin^{\top} \|_F^{2} & = \| \wbool \odot \mbE_2 \|_F^2                            \\
                                                            & = \sum_{i,j} \mbW_{\mathrm{bool}[i,j]}^2 + \mbE_{2[i,j]}^2 \\
                                                            & = \sum_{i,j} \mbE_{2[i,j]}^2                               \\
                                                            & = \| \mbE_2 \|_F^2 \label{eq:approx_e2}
    \end{align}

    For \gls{SVD} decomposition, the norm of the above error matrices in the rank-1 approximation is the um of squares of all singular values except the largest one.
    In particular, we have:
    \begin{align}
        \|\mbE_1\|_F^2 & = \sum_{i=2}^{n} \sigma_i^2(\mbW),   \\
        \|\mbE_2\|_F^2 & = \sum_{i=2}^{n} \sigma_i^2(|\mbW|).
    \end{align}

    Since $\|\mbW\|_F^2 = \||\mbW|\|_F^2$, we have:
    \begin{align}
        \sum_{i=1}^{n} \sigma_i^2(\mbW)                   & = \sum_{i=1}^{n} \sigma_i^2(|\mbW|)    \\
        \Leftrightarrow \|\mbE_1\|_F^2 + \sigma_1^2(\mbW) & =  \|\mbE_2\|_F^2  \sigma_1^2(|\mbW|).
    \end{align}
    Thus, according to \cref{lemma:biggest_singular} and \cref{eq:approx_e2}, we have:
    \begin{align}
        \|\mbE_2\|_F^2                                                 & \leq \|\mbE_1\|_F^2                                    \\
        \left\| \mbW - \wbool \odot \ssout \ssin^{\top} \right\|_F^{2} & \leq \left\| \mbW - \mba \mbb^{\top} \right\|_{F}^{2}.
    \end{align}

    Thus, the proposition is proved. \qedhere

\end{proof}

\subsection{Proof of \cref{pro:svid_rank_1}}
\label{proof:svid_rank_1}

\begin{mybox}
    \begin{proposition}
        For $\mbW \in \mathbb{R}^{m\times n}$, we denote $|\mbW| = \mbU \mbSigma \mbV^{\top}$ its \gls{SVD}.
        $\mbs_{\mathrm{in}}$ and $\mbs_{\mathrm{out}}$ are given as: $\mbs_{\mathrm{in}} = \sqrt{\sigma_1} \mbV_{[:,1]}$, and $\mbs_{\mathrm{out}} = \sqrt{\sigma_1} \mbU_{[:,1]}$.
        We decompose the matrix as $\mbW = \mbW_{\mathrm{bool}} \odot |\mbW| \approx \mbW_{\mathrm{bool}} \odot \left( \mbs_{\mathrm{out}} \mbs_{\mathrm{in}}^{\top} \right)$.
        We then have:
        \begin{align}
            \left\| \mbW - \mbW_{\mathrm{bool}} \odot \mbs_{\mathrm{out}} \mbs_{\mathrm{in}}^{\top}   \right\|_{F}^{2} \leq \left\| \mbW - \mbW_{\mathrm{bool}} \odot \mbc \mbd^{\top}   \right\|_{F}^{2}, \quad \forall \mbc \in \mathbb{R}^{m \times 1}, \forall \mbd \in \mathbb{R}^{n \times 1}.
        \end{align}
    \end{proposition}
\end{mybox}

\begin{proof}
    Similar to the proof of \cref{pro:svid_rank_1}, we denote the following error matrices $\mbE_1 = |\mbW| - \mbs_{\mathrm{out}} \mbs_{\mathrm{in}}^{\top}$ and $\mbE_2 = |\mbW| - \mbc \mbd^{\top}$.
    We have that
    \begin{align}
        \mbW_{\mathrm{bool}} \odot |\mbW| - \mbW_{\mathrm{bool}} \odot \mbs_{\mathrm{out}} \mbs_{\mathrm{in}}^{\top} & =  \mbW_{\mathrm{bool}} \odot \mbE_1  \\
        \Leftrightarrow \mbW - \mbW_{\mathrm{bool}} \odot \mbs_{\mathrm{out}} \mbs_{\mathrm{in}}^{\top}              & =  \mbW_{\mathrm{bool}} \odot \mbE_1.
    \end{align}
    Therefore,
    \begin{align}
        \left\| \mbW - \mbW_{\mathrm{bool}} \odot \mbs_{\mathrm{out}} \mbs_{\mathrm{in}}^{\top}   \right\|_{F}^{2} = \left\| \mbW_{\mathrm{bool}} \odot \mbE_1  \right\|_{F}^{2} = \sum_{i,j} \mbW_{\mathrm{bool}[i,j]}^2 \mbE_{1[i,j]}^2 = \sum_{i,j} \mbE_{1[i,j]}^2 = \|\mbE_1\|_{F}^{2}.
    \end{align}

    Similarly, we have that
    \begin{align}
        \left\| \mbW - \mbW_{\mathrm{bool}} \odot \mba \mbb^{\top}   \right\|_{F}^{2} = \|\mbE_2\|_{F}^{2}.
    \end{align}
    Thus, we need to show that
    \begin{align}
        \|\mbE_1\|_{F}^{2} \leq \|\mbE_2\|_{F}^{2} \label{eq:svd_rank_1_1}
    \end{align}

    Additionally, we denote the rank-$k$ approximation to $|\mbW|$ by \gls{SVD} as $\mbS_k$:
    \begin{align}
        \mbS_{k} = \sum_{i=1}^{k} \sigma_{i} \mbU_{[:,i]} \mbV_{[:,i]}^{\top}.
    \end{align}
    With this notation, we have that $\mbS_1 = \mbs_{\mathrm{out}} \mbs_{\mathrm{in}}^{\top}$ is the rank-1 approximation of $|\mbW|$ by \gls{SVD}.

    From \cref{eq:svd_rank_1_1}, we need to show that if there is an arbitrary rank-1 approximation  to $| \mbW |$, $\mbP_1 = \mbc \mbd^\top$, we then have
    \begin{align}
        \left\| |\mbW| - \mbs_{\mathrm{out}} \mbs_{\mathrm{in}}^{\top} \right\|_{F}^{2} \leq \left\| |\mbW| - \mbc \mbd^{\top} \right\|_{F}^{2}.
    \end{align}

    This can be done by using the Eckart-Young-Mirsky theorem \citep{eckart1936approx}.
    First, we have that
    \begin{align}
        \left\| |\mbW| - \mbS_1 \right\|^2_{F} = \left\| |\mbW| - \mbs_{\mathrm{out}} \mbs_{\mathrm{in}}^{\top} \right\|^2_{F} = \left\| \sum_{i=2}^{n} \sigma_i \mbU_{[:,i]} \mbV_{[:,i]}^{\top} \right\|^2_{F} = \sum_{i=2}^{n} \sigma^2_i.
    \end{align}

    By the triangle inequality with the spectral norm, if $|\mbW| = \mbC + \mbD$ then $\sigma_1(|\mbW|) \leq \sigma_1(\mbC) + \sigma_1(\mbD)$.
    Suppose the $\mbC_k$ and $\mbD_k$ denote the rank-$k$ approximation to $\mbC$ and $\mbD$ by \gls{SVD} method, respectively.
    Then, for any $i,j \geq 1$ we have
    \begin{align}
        \sigma_i(\mbC) + \sigma_j(\mbD) & = \sigma_1(\mbC - \mbC_{i-1}) + \sigma_1(\mbD - \mbD_{j-1})                                             \\
                                        & \geq \sigma_1(|\mbW| - \mbC_{i-1} - \mbD_{j-1})                                                         \\
                                        & \geq \sigma_1(|\mbW| - \mbS_{i+j-2}) \qquad (\text{since rank}(\mbC_{i-1} + \mbD_{j-1}) \leq i + j -2 ) \\
                                        & =\sigma_{i+j-1}(|\mbW|).
    \end{align}
    Because $\sigma_{2}(\mbP_1) = 0$, when $\mbC = |\mbW| - \mbP_1$ and $\mbD = \mbP_1$ we have that for $i \geq 1, j=2$, $\sigma_i(|\mbW| - \mbP_1) \geq \sigma_{i+1}(|\mbW|)$.
    As a result,
    \begin{align}
        \left\| |\mbW| - \mbP_1 \right\|^2_{F} = \sum{i=1}^{n} \sigma_i (|\mbW| - \mbP_1)^2            & \geq \sum{i=2}^{n} \sigma_i (|\mbW|)^2 = \left\| |\mbW| - \mbS_1 \right\|_F^2                                    \\
        \Leftrightarrow \|\mbE_2\|_{F}^{2}                                                             & \geq \|\mbE_1\|_{F}^{2}                                                                                          \\
        \Leftrightarrow  \left\| \mbW - \mbW_{\mathrm{bool}} \odot \mbc \mbd^{\top}   \right\|_{F}^{2} & \geq \left\| \mbW - \mbW_{\mathrm{bool}} \odot \mbs_{\mathrm{out}} \mbs_{\mathrm{in}}^{\top}   \right\|_{F}^{2}.
    \end{align}
    Hence the proposition is proved. \qedhere

\end{proof}

\newpage

\section{Details on Kernel Allocation}

\subsection{Weight Importance Estimation} \label{sub:weight_importance}

We assess the importance of a linear weight in the original \gls{FP} model by comparing the representations at its input and output.
Let $\mbX \in \bbR^{d \times n}$ and $\mbY \in \bbR^{d \times m}$ denote the input and output matrices of a linear layer, respectively, where $d$ is the number of samples, and $n$ and $m$ are the input and output feature dimensions.
We hypothesize that a weight is important if it significantly transforms the input representations.
For example, a weight matrix equivalent to the identity does not alter the representations and thus would be considered unimportant.
To quantify this transformation, we use a robust metric for comparing neural representations.

Various similarity measures can be used for this purpose, such as cosine similarity, as done in \citep{gromov2025the}. 
In this work, we adopt \gls{PWCCA} \cite{Morcos2018}, which is particularly well-suited for our setting: it is invariant to linear transformations—an essential property given that large language models (\glspl{LLM}) are primarily composed of linear layers—and effectively captures shared structure while filtering out noise \cite{Morcos2018}.

Specifically, we define the importance score as:
\begin{align}
    h = 1 - \frac{1}{c} \sum_{i=1}^{c} \rho_{\mathrm{PWCCA},i}(\mbX, \mbY),
\end{align}
where $c$ denotes the number of canonical vectors used in the comparison (typically, $c = \min(n, m)$). 
The matrices $\mbX$ and $\mbY$ are obtained by simply forwarding a set of data samples through the network.
In our experiments, we use 128 random samples from the WikiText2 training set to estimate the importance score.
Here, $\rho_{\mathrm{PWCCA},i}$ represents the projection-weighted correlation along the $i$-th canonical direction.
The following section describes in detail how this correlation is computed.

\begin{algorithm}[H]\small
	\caption{Kernel allocation.}
	\label{algo:KernelAllocation}
	\SetKwInOut{Output}{Output}
	\SetKwBlock{Loop}{Loop}{end}
	\SetKwBlock{Input}{Input}{end}
	\SetKwBlock{Initialize}{Initialize}{end}
	\SetKwFor{When}{When}{do}{end}
	\SetKwFor{While}{While}{do}{end}
	\SetKwFunction{Wait}{Wait}
	\SetAlgoLined
	\SetNoFillComment
	\Input{
		$T \ge 1$ \tcc*{model expansion limit}
		$\mbE = [e_l^{[k]}] \in \bbR^{N_{\mbW} \times K_{\max}}$ for $k \in [1, K_{\max}]$, $l \in [1, N_{\mbW}]$ \tcc*[r]{residual approx error}
		$\mbh = [h_l] \in \bbR^{N_{\mbW} \times 1}$ \tcc*[r]{weight importance scores}
		$\mbp = [p_l] \in \bbR^{N_{\mbW} \times 1}$ \tcc*[r]{weight size ratios}
	}
	\Initialize{
		$\mbk = \trans{[1, \ldots, 1]}$ of length $N_{\mbW}$ \tcc*[r]{starting choice}
		$\mbf = \mbk < K_{\max}$ \tcc*[r]{feasible indicator}
		$\mbC = \left(\frac{1}{\mbp}\log\frac{1}{\mbp}\right) \odot \mbh \odot \mbE$ \tcc*[r]{where $\odot$ is broadcasted over $\mbE$ columns}
	}
	\While{not all $\mbf$ is $\False$}{
		$\mbg := \emptyset$, $\mbl := \emptyset$\; 
		\For{$l = 1 : N_{\mbW}$}{
			\If{$\mbf[l] = \True$}{
				$g := \mbC[l, \mbk[l]] - \mbC[l, \mbk[l]+1]$ \tcc*[r]{gain by increasing kernel size by 1}
				Append $l$ to $\mbl$, append $g$ to $\mbg$\;
			}
		}
		Sort $\mbg$ in decreasing order, and arrange $\mbl$ accordingly\;
		\For{$(g,l)$ in $(\mbg, \mbl)$}{
			$\mbk_l := \mbk$\;
			$\mbk_l[l] = \mbk_l[l] +  1$\;
			\eIf{$\trans{\mbk_l} \mbp \le T$}{
				$\mbk[l] = \mbk[l] + 1$\;
				\textbf{break} \tcc*{escape the for loop}
			}{
				$\mbf[l] := \False$\;
			}
		}
		$\mbf \leftarrow \andd(\mbf, \mbk < K_{\max})$ \tcc*{element-wise logical \textbf{and}}
	}
	\Return{$\mbk$}
\end{algorithm}

\paragraph{Projection-weighted Canonical Correlation Analysis.}

\gls{CCA} finds bases for two matrices such that, when the original matrices are projected onto these bases, the resulting projections are maximally correlated.
Without loss of generality, we assume that $n \leq m$.
For $1 \leq i \leq n$, the $i$-th canonical correlation coefficient $\rho_i$ is given by:
\begin{align}
    \rho_i = \max_{\mbw_{\mbX}^{i},\mbw_{\mbY}^{i}} & \mathrm{corr}(\mbX \mbw_{\mbX}^{i}, \mbY \mbY \mbw_{\mbY}^{i}) \label{eq:cca_objective}               \\
    \text{subject to } \text{ }                     & \mbX \mbw_{\mbX}^{i} \text{ } \bot \text{ } \mbX \mbw_{\mbX}^{j} \quad \forall  j < i \nonumber  \\
                                                    & \mbY \mbw_{\mbY}^{i} \text{ } \bot \text{ } \mbY \mbw_{\mbY}^{j} \quad \forall  j < i. \nonumber
\end{align}

The vectors $\mbw_{\mbX}^i \in \bbR^{n}$ and $\mbw_{\mbY}^i \in \bbR^{m}$ that maximize $\rho_i$ are called the canonical weights.
These weights transform the original data into the canonical variables $\mbX \mbw_{\mbX}^{i}$ and $\mbY \mbw_{\mbY}^{i}$.
The constraints in \cref{eq:cca_objective} enforce orthogonality among the canonical variables, ensuring that each successive pair captures a distinct mode of correlation.

The mean \gls{CCA} correlation is then computed as:
\begin{align}
    \bar{\rho}_{\mathrm{CCA}} = \frac{\sum_{i=1}^{n} \rho_i}{n},
\end{align}
where $n$ is the number of canonical correlation coefficients considered.

\gls{CCA} is sensitive to perturbation when the condition number of $\mbX$ and $\mbY$ is large.
To imporve robustness, \cite{Morcos2018} propose a strategy to reduce this sensitivity, which they term ``projection-weighted \textsc{cca}'' (\textsc{pwcca}).
\begin{align}
    \rho_{\mathrm{PWCCA},i} & = \frac{ \sum_{i=1}^c \alpha_i \rho_i }{\sum_{i=1}^c \alpha_i},
    \quad \alpha_i = \sum_j |\langle \mbh_i, \mbx_j \rangle |,
\end{align}
where $\mbx_j$ is the $j$-th column of $\mbX$, and $\mbh_i = \mbX \mbw_{\mbX}^i$ is the vector of canonical variables formed by projecting $\mbX$ to the $i$-th canonical cooridate frame.

\subsection{Kernel Allocation Algorithm}

\cref{algo:KernelAllocation} illustrates the details of our algorithm for kernel allocation.

\section{Theoretical Analysis of Training Complexity} \label{sec:training_complexity}

Consider a linear layer without bias, defined as $\mbY = \mbX \mbW$
where \(\mbX \in \mathbb{R}^{B \times L \times N}\) and \(\mbW \in \mathbb{R}^{N \times M}\).  
Here, \(B\) is the mini-batch size, \(L\) is the sequence length, \(N\) is the input dimension, and \(M\) is the output dimension.  
We analyze the number of multiplications (MULs) required.  

\textbf{Latent-weight approach (same cost as full-precision training):} 
{
    \setlist[itemize]{leftmargin=4mm} 
\begin{itemize}
    \item Forward: \(B \times L \times N \times M\) (\halffp--\halffp MULs)
    \item Backward w.r.t. weights: \(B \times L \times N \times M\) (\halffp--\halffp MULs)
    \item Backward w.r.t. inputs: \(B \times L \times N \times M\) (\halffp--\halffp MULs)
    \item \textbf{Total:} $3 \times B \times L \times N \times M $ \halffp--\halffp MULs
\end{itemize}
}

\textbf{Boolean approach with \(K\) kernels:}  
(assuming \halffp gradients for a fair comparison). As shown in the main text, only the final Boolean kernel needs to be fine-tuned. The number of multiplications becomes:  

{
    \setlist[itemize]{leftmargin=4mm}
\begin{itemize}
    \item Forward: \(K \times B \times L \times N \times M\) (\booll--\halffp MULs, using all \(K\) kernels)
    \item Backward w.r.t. weights: \(1 \times B \times L \times N \times M\) (\halffp--\halffp MULs, for last kernel only)
    \item Backward w.r.t. inputs: \(1 \times B \times L \times N \times M\) (\booll--\halffp MULs, for last kernel only)
    \item \textbf{Total:} $(K+1) \times B \times L \times N \times M$ \booll--\halffp MULs, and $B \times L \times N \times M$ \halffp--\halffp MULs
\end{itemize}
}

Since \(K\) is typically small (e.g., 2--4) while \(B\) and \(L\) are large (thousands), most computation shifts from \halffp--\halffp to the more efficient \booll--\halffp operations.
If we ignore the \booll--\halffp MULs, the \halffp--\halffp  operations are reduced by a factor of \(2/3\) (i.e., a 66.7\% reduction).
Remarkably, this reduction is achieved while using more kernels and attaining better performance, yet with significantly lower training complexity.  
According to \bitnet \citep{wang2023bitnet} (Table 1), for \(L = 512\) and a LLaMA-like 13B model on 7\,nm hardware, {\small{1}}\textsc{b}{\small{it}}--\halffp operations yield an energy saving of approximately \textbf{56$\times$} compared to \halffp--\halffp.
Hence, our method achieves substantial training efficiency.
Importantly, \bitnet is a latent-weight approach, with efficiency gains realized primarily during inference, whereas our method provides significant benefits already during training and fine-tuning.  

We note that the above analysis does not include optimizer cost. The latent-weight approach typically relies on Adam, which requires two full-precision momenta per parameter and a complex update rule involving multiple normalization statistics.
By contrast, our Boolean approach employs a Boolean optimizer requiring only one full-precision momentum per parameter, coupled with a much simpler update rule (see \cref{eq:bool_optimizer_main}).
This further underscores the reduction in overall training complexity offered by our method.

\section{Additional Experiemental Results}

\subsection{Additional Information of Experiemental Settings}

We use $12$ Nvidia GPUs of Tesla V100 for our experiments.
We follow exactly the experimental settings in \cite{jo2024mixture}.
The results of the baselines in \cref{tab:benchmark_main} are taken from \cite{xu2024onebit,jo2024mixture}.

\subsection{On the Choice of KD Loss} \label{sec:kd_loss_choice}

\begin{figure}[H]
    \centering
    \begin{subfigure}[c]{0.63\textwidth}

        \tikzexternaldisable
        \centering
        \scriptsize
        \setlength{\figurewidth}{5.1cm}
        \setlength{\figureheight}{4.7cm}
        \input{figures/loss_study.tikz}
        \tikzexternalenable

    \end{subfigure}
    \begin{subfigure}[c]{0.35\textwidth}
        \tikzexternaldisable

\scalebox{.72}{
    \setlength{\tabcolsep}{3.9pt}
    \renewcommand{\arraystretch}{1.2}
    \begin{tabular}{lcc}
        \toprule
        $\text{D}_{\textrm{logits}}$                                                                                                                                               & Wiki2          & C4             \\
        \midrule
        {\protect\tikz[baseline=-1ex]\protect\draw[color=color_blue, fill=color_blue, opacity=0.99, mark size=1.7pt, line width=1.7pt] plot[] (-0.0,0)--(-0.45,0);} Forward KL     & \textbf{31.39} & \textbf{28.50} \\
        {\protect\tikz[baseline=-1ex]\protect\draw[color=color_orange, fill=color_orange, opacity=0.99, mark size=1.7pt, line width=1.7pt] plot[] (-0.0,0)--(-0.45,0);} Reverse KL & 33.14          & 29.46          \\
        {\protect\tikz[baseline=-1ex]\protect\draw[color=color_green, fill=color_green, opacity=0.99, mark size=1.7pt, line width=1.7pt] plot[] (-0.0,0)--(-0.45,0);} Symmetric KL & 32.67          & 29.26          \\
        {\protect\tikz[baseline=-1ex]\protect\draw[color=color_red, fill=color_red, opacity=0.99, mark size=1.7pt, line width=1.7pt] plot[] (-0.0,0)--(-0.45,0);} JS Divergence    & 31.78          & 28.69          \\
        {\protect\tikz[baseline=-1ex]\protect\draw[color=color_purle, fill=color_purle, opacity=0.99, mark size=1.7pt, line width=1.7pt] plot[] (-0.0,0)--(-0.45,0);} TV Distance  & 33.02          & 29.56          \\
        \bottomrule
    \end{tabular}
}

\tikzexternalenable

    \end{subfigure}

    \vspace{-1ex}

    \caption{The training convergence of $\cL_{\mathrm{is}}$, and $\cL_{\mathrm{logits}}$, measured by Forward \gls{KL}, and the final results with respect to the choice of $\textrm{D}_{\textrm{logits}}$.}
    \label{fig:loss_study}
\end{figure}

\cref{fig:loss_study} illustrates the convergence and results of using different choices for $\mathrm{D}_{\textrm{logits}}$ in \cref{eq:loss_logits}.
Despite its simplicity, forward \gls{KL} achieves the best performance.
More complex measures, such as total variance (TV) distance \citep{wen2023fdivergence} and Jensen-Shannon (JS) divergence \citep{agarwal2024onpolicy}, offer no significant benefits in our case.
Furthermore, we observe that the final perplexity is strongly correlated with  $\cL_{\textrm{logits}}$ using forward \gls{KL}, but not with $\cL_{\textrm{is}}$, as shown in \cref{fig:loss_study} and \cref{fig:optim_strategy}.
As a result, we employ the forward \gls{KL} in all experiments.

\subsection{Results of Different Number of Kernels on LLMs} \label{sec:benchmark_more_kernels}

To complement the \cref{tab:benchmark_main}, \cref{tab:benchmark_main_more_kernels} shows the benchmarking results of \glspl{LLM} using our \ours{} method with varying numbers of kernels per weight. 
Consistent with the observations made on smaller models in \cref{sec:exp_num_kernels}, we observe that increasing the number of kernels generally improves performance. 
However, the performance gains begin to diminish noticeably beyond three kernels.

\begin{table}[H]
  \setlength{\tabcolsep}{3.8pt}
  \centering
  \caption{Perplexity and zero-shot accuracy results of our \ours method with different number of kernels.}
  \label{tab:benchmark_main_more_kernels}
  \renewcommand{\arraystretch}{1.15}
  \scalebox{.79}{
    \begin{tabular}{llcccccccccc}
      \toprule
      \multirow{2}{*}{\textbf{Model}}                                           & \multirow{2}{*}{\textbf{Method}} & \multirow{2}{*}{\textbf{Wbits}} & \multicolumn{2}{c}{\textbf{Perplexity ($\downarrow$})} & \multicolumn{7}{c}{\textbf{Zero-shot Accuracy ($\uparrow$})}                                                                                                                                                                                                                                                                                                  \\
                                                                                &                                  &                                 & \textbf{Wiki2}                                         & \textbf{C4}                                                  & BoolQ                                 & PIQA                                  & Hella.                                & WinoG.                                & ARC-e                                 & ARC-c                                 & \textbf{Average}                               \\
      \midrule
      \midrule

      \multicolumn{1}{l}{\multirow{3}{*}{\optbig}}         & \ours (2 kernels)                & \small{2$\times$1}              & \cellcolor[HTML]{dfebdd}\small{{16.13}}                & \cellcolor[HTML]{dfebdd}\small{{16.61}}                      & \cellcolor[HTML]{dfebdd}\small{58.53} & \cellcolor[HTML]{dfebdd}\small{70.67} & \cellcolor[HTML]{dfebdd}\small{48.11} & \cellcolor[HTML]{dfebdd}\small{56.75} & \cellcolor[HTML]{dfebdd}\small{48.19} & \cellcolor[HTML]{dfebdd}\small{27.90} & \cellcolor[HTML]{dfebdd}\small{{51.69}}        \\
      \multicolumn{1}{c}{}                                                      & \ours (3 kernels)                & \small{3$\times$1}              & \cellcolor[HTML]{d2ebce}\small{{15.30}}                & \cellcolor[HTML]{d2ebce}\small{{15.68}}                      & \cellcolor[HTML]{d2ebce}\small{60.64} & \cellcolor[HTML]{d2ebce}\small{70.78} & \cellcolor[HTML]{d2ebce}\small{50.71} & \cellcolor[HTML]{d2ebce}\small{56.83} & \cellcolor[HTML]{d2ebce}\small{48.82} & \cellcolor[HTML]{d2ebce}\small{28.49} & \cellcolor[HTML]{d2ebce}\small{{52.71}}        \\
      \multicolumn{1}{c}{}                                                      & \ours (4 kernels)                & \small{4$\times$1}              & \cellcolor[HTML]{c4edbe}\small{\textbf{14.83}}         & \cellcolor[HTML]{c4edbe}\small{\textbf{14.92}}               & \cellcolor[HTML]{c4edbe}\small{60.95} & \cellcolor[HTML]{c4edbe}\small{70.85} & \cellcolor[HTML]{c4edbe}\small{51.02} & \cellcolor[HTML]{c4edbe}\small{56.85} & \cellcolor[HTML]{c4edbe}\small{49.13} & \cellcolor[HTML]{c4edbe}\small{29.24} & \cellcolor[HTML]{c4edbe}\small{\textbf{53.01}} \\

      \midrule

      \multicolumn{1}{l}{\multirow{3}{*}{\llamasmall }} & \ours (2 kernels)                & \small{2$\times$1}              & \cellcolor[HTML]{dfebdd}\small{{6.83}}                 & \cellcolor[HTML]{dfebdd}\small{{8.53}}                       & \cellcolor[HTML]{dfebdd}\small{69.20} & \cellcolor[HTML]{dfebdd}\small{74.32} & \cellcolor[HTML]{dfebdd}\small{64.80} & \cellcolor[HTML]{dfebdd}\small{60.30} & \cellcolor[HTML]{dfebdd}\small{49.05} & \cellcolor[HTML]{dfebdd}\small{34.90} & \cellcolor[HTML]{dfebdd}\small{{58.76}}        \\
      \multicolumn{1}{c}{}                                                      & \ours (3 kernels)                & \small{3$\times$1}              & \cellcolor[HTML]{d2ebce}\small{{6.20}}                 & \cellcolor[HTML]{d2ebce}\small{{7.76}}                       & \cellcolor[HTML]{d2ebce}\small{67.89} & \cellcolor[HTML]{d2ebce}\small{76.15} & \cellcolor[HTML]{d2ebce}\small{68.91} & \cellcolor[HTML]{d2ebce}\small{63.30} & \cellcolor[HTML]{d2ebce}\small{48.94} & \cellcolor[HTML]{d2ebce}\small{37.62} & \cellcolor[HTML]{d2ebce}\small{{60.47}}        \\
      \multicolumn{1}{c}{}                                                      & \ours (4 kernels)                & \small{4$\times$1}              & \cellcolor[HTML]{c4edbe}\small{\textbf{6.01}}          & \cellcolor[HTML]{c4edbe}\small{\textbf{7.53}}                & \cellcolor[HTML]{c4edbe}\small{68.16} & \cellcolor[HTML]{c4edbe}\small{76.71} & \cellcolor[HTML]{c4edbe}\small{69.85} & \cellcolor[HTML]{c4edbe}\small{62.09} & \cellcolor[HTML]{c4edbe}\small{49.24} & \cellcolor[HTML]{c4edbe}\small{38.14} & \cellcolor[HTML]{c4edbe}\small{\textbf{60.70}} \\

      \midrule

      \multicolumn{1}{l}{\multirow{3}{*}{\llamabig }}   & \ours (2 kernels)                & \small{2$\times$1}              & \cellcolor[HTML]{dfebdd} {\small{6.17}}                & \cellcolor[HTML]{dfebdd} {\small{7.88}}                      & \cellcolor[HTML]{dfebdd}\small{68.10} & \cellcolor[HTML]{dfebdd}\small{76.33} & \cellcolor[HTML]{dfebdd}\small{69.88} & \cellcolor[HTML]{dfebdd}\small{64.17} & \cellcolor[HTML]{dfebdd}\small{52.34} & \cellcolor[HTML]{dfebdd}\small{37.88} & \cellcolor[HTML]{dfebdd}{\small{61.45}}        \\
      \multicolumn{1}{c}{}                                                      & \ours (3 kernels)                & \small{3$\times$1}              & \cellcolor[HTML]{d2ebce}\small{{5.58}}                 & \cellcolor[HTML]{d2ebce}\small{{7.15}}                       & \cellcolor[HTML]{d2ebce}\small{67.39} & \cellcolor[HTML]{d2ebce}\small{77.74} & \cellcolor[HTML]{d2ebce}\small{73.37} & \cellcolor[HTML]{d2ebce}\small{66.61} & \cellcolor[HTML]{d2ebce}\small{54.04} & \cellcolor[HTML]{d2ebce}\small{41.21} & \cellcolor[HTML]{d2ebce}\small{{63.39}}        \\
      \multicolumn{1}{c}{}                                                      & \ours (4 kernels)                & \small{4$\times$1}              & \cellcolor[HTML]{c4edbe}\small{\textbf{5.38}}          & \cellcolor[HTML]{c4edbe}\small{\textbf{6.91}}                & \cellcolor[HTML]{c4edbe}\small{68.69} & \cellcolor[HTML]{c4edbe}\small{77.63} & \cellcolor[HTML]{c4edbe}\small{74.23} & \cellcolor[HTML]{c4edbe}\small{66.53} & \cellcolor[HTML]{c4edbe}\small{56.14} & \cellcolor[HTML]{c4edbe}\small{41.38} & \cellcolor[HTML]{c4edbe}\small{\textbf{64.10}} \\
      \bottomrule
    \end{tabular}}
  \vspace{-3ex}
\end{table}

\subsection{Additional Results on LLaMA-2} \label{sec:bencmark_llama2}

\cref{tab:benchmark_main_llama_2} shows the results on \llamatwobig \citep{touvron2023llama2}.
Similar to the \cref{tab:benchmark_main}, the results of the baselines are taken from \cite{xu2024onebit} and \cite{jo2024mixture}.
It is clear that our method consistently outperforms the baselines across different metrics and model sizes.This further emphasizes the robustness of our approach across various types of models.

\begin{table}[H]
  \setlength{\tabcolsep}{3.8pt}
  \centering
  \caption{Perplexity and zero-shot accuracy results of Float16, quantized and binarized \llamatwo models.}
  \label{tab:benchmark_main_llama_2}
  \renewcommand{\arraystretch}{1.02}
  \scalebox{.83}{
    \begin{tabular}{llcccccccccc}
      \toprule
      \multirow{2}{*}{\textbf{Model}}                     & \multirow{2}{*}{\textbf{Method}} & \multirow{2}{*}{\textbf{Wbits}} & \multicolumn{2}{c}{\textbf{Perplexity ($\downarrow$})} & \multicolumn{7}{c}{\textbf{Zero-shot Accuracy ($\uparrow$})}                                                                                                                                                                                                                                                                                                  \\
                                                          &                                  &                                 & \textbf{Wiki2}                                         & \textbf{C4}                                                  & BoolQ                                 & PIQA                                  & Hella.                                & WinoG.                                & ARC-e                                 & ARC-c                                 & \textbf{Average}                               \\
      \midrule
      \midrule

      \multicolumn{1}{l}{\multirow{9}{*}{\llamatwosmall}} & \halffp                          & \small{16}                      & \small{5.47}                                           & \small{6.97}                                                 & \small{71.10}                         & \small{76.88}                         & \small{72.94}                         & \small{67.09}                         & \small{53.58}                         & \small{40.61}                         & \small{63.70}                                  \\
      \cmidrule(l){2-12}
      \multicolumn{1}{c}{}                                & \pbllm                           & \small{1.7}                     & \small{76.75}                                          & \small{85.92}                                                & \small{62.17}                         & \small{52.82}                         & \small{26.87}                         & \small{50.11}                         & \small{26.89}                         & \small{24.31}                         & \small{40.53}                                  \\
      \multicolumn{1}{c}{}                                & \billm                           & \small{1.11}                    & \small{27.72}                                          & \small{36.34}                                                & \small{62.14}                         & \small{59.19}                         & \small{35.18}                         & \small{53.11}                         & \small{34.22}                         & \small{26.54}                         & \small{45.06}                                  \\
      \multicolumn{1}{c}{}                                & \onebit                          & \small{1}                       & \small{8.60}                                           & \small{10.74}                                                & \small{63.06}                         & \small{70.40}                         & \small{54.24}                         & \small{56.67}                         & \small{40.82}                         & \small{29.35}                         & \small{52.42}                                  \\
      \multicolumn{1}{c}{}                                & \mos                             & \small{1}                       & \small{7.88}                                           & \small{9.75}                                                 & \small{65.02}                         & \small{71.55}                         & \small{59.41}                         & \small{56.18}                         & \small{41.84}                         & \small{30.03}                         & \small{54.01}                                  \\
      \cmidrule(l){2-12}
      \multicolumn{1}{c}{}                                & \qptq                            & \small{2}                       & \small{7.7e3}                                          & \small{NaN}                                                  & \small{42.97}                         & \small{49.46}                         & \small{26.19}                         & \small{50.28}                         & \small{26.77}                         & \small{28.58}                         & \small{37.38}                                  \\
      \multicolumn{1}{c}{}                                & \llmqat                          & \small{2}                       & \small{1.1e3}                                          & \small{6.6e2}                                                & \small{59.14}                         & \small{50.12}                         & \small{25.10}                         & \small{49.08}                         & \small{26.26}                         & \small{26.96}                         & \small{35.89}                                  \\
      \multicolumn{1}{c}{}                                & \omniquant                       & \small{2}                       & \small{31.21}                                          & \small{64.34}                                                & \small{58.69}                         & \small{56.53}                         & \small{33.87}                         & \small{51.22}                         & \small{33.63}                         & \small{24.32}                         & \small{43.12}                                  \\
      \cmidrule(l){2-12}
      \multicolumn{1}{c}{}                                & \ours [Ours]                     & \small{2$\times$1}              & \cellcolor[HTML]{d6e9c9}\small{\textbf{6.87}}          & \cellcolor[HTML]{d6e9c9}\small{\textbf{8.74}}                & \cellcolor[HTML]{d6e9c9}\small{66.94} & \cellcolor[HTML]{d6e9c9}\small{74.97} & \cellcolor[HTML]{d6e9c9}\small{65.59} & \cellcolor[HTML]{d6e9c9}\small{61.72} & \cellcolor[HTML]{d6e9c9}\small{44.82} & \cellcolor[HTML]{d6e9c9}\small{34.21} & \cellcolor[HTML]{d6e9c9}\small{\textbf{58.04}} \\
      \multicolumn{1}{c}{}                                & \ours [Ours]                     & \small{3$\times$1}              & \cellcolor[HTML]{d6e9c9}\small{\textbf{6.12}}          & \cellcolor[HTML]{d6e9c9}\small{\textbf{7.81}}                & \cellcolor[HTML]{d6e9c9}\small{65.46} & \cellcolor[HTML]{d6e9c9}\small{75.79} & \cellcolor[HTML]{d6e9c9}\small{69.59} & \cellcolor[HTML]{d6e9c9}\small{62.04} & \cellcolor[HTML]{d6e9c9}\small{49.11} & \cellcolor[HTML]{d6e9c9}\small{37.80} & \cellcolor[HTML]{d6e9c9}\small{\textbf{59.97}} \\

      \midrule
      \midrule

      \multicolumn{1}{l}{\multirow{9}{*}{\llamatwobig }}  & \halffp                          & \small{16}                      & \small{4.88}                                           & \small{6.47}                                                 & \small{68.99}                         & \small{79.05}                         & \small{76.62}                         & \small{69.77}                         & \small{57.95}                         & \small{44.20}                         & \small{66.10}                                  \\
      \cmidrule(l){2-12}
      \multicolumn{1}{c}{}                                & \pbllm                           & \small{1.7}                     & \small{155.25}                                         & \small{151.15}                                               & \small{37.82}                         & \small{53.26}                         & \small{28.89}                         & \small{49.48}                         & \small{28.28}                         & \small{23.72}                         & \small{36.91}                                  \\
      \multicolumn{1}{c}{}                                & \billm                           & \small{1.11}                    & \small{20.71}                                          & \small{27.19}                                                & \small{62.20}                         & \small{62.51}                         & \small{38.05}                         & \small{56.35}                         & \small{40.69}                         & \small{27.73}                         & \small{47.92}                                  \\
      \multicolumn{1}{c}{}                                & \onebit                          & \small{1}                       & \small{7.56}                                           & \small{9.67}                                                 & \small{65.66}                         & \small{71.60}                         & \small{60.07}                         & \small{56.91}                         & \small{45.76}                         & \small{31.74}                         & \small{55.29}                                  \\
      \multicolumn{1}{c}{}                                & \mos                             & \small{1}                       & \small{7.08}                                           & \small{8.91}                                                 & \small{66.12}                         & \small{73.72}                         & \small{63.80}                         & \small{58.98}                         & \small{45.71}                         & \small{33.19}                         & \small{57.09}                                  \\
      \cmidrule(l){2-12}
      \multicolumn{1}{c}{}                                & \qptq                            & \small{2}                       & \small{2.1e3}                                          & \small{3.2e2}                                                & \small{40.61}                         & \small{51.74}                         & \small{25.67}                         & \small{51.85}                         & \small{25.46}                         & \small{27.30}                         & \small{37.11}                                  \\
      \multicolumn{1}{c}{}                                & \llmqat                          & \small{2}                       & \small{5.1e2}                                          & \small{1.1e3}                                                & \small{39.85}                         & \small{49.08}                         & \small{24.37}                         & \small{51.38}                         & \small{27.15}                         & \small{24.32}                         & \small{36.03}                                  \\
      \multicolumn{1}{c}{}                                & \omniquant                       & \small{2}                       & \small{16.88}                                          & \small{27.02}                                                & \small{62.05}                         & \small{62.24}                         & \small{50.34}                         & \small{53.20}                         & \small{40.66}                         & \small{29.61}                         & \small{49.68}                                  \\
      \cmidrule(l){2-12}
      \multicolumn{1}{c}{}                                & \ours [Ours]                     & \small{2$\times$1}              & \cellcolor[HTML]{d6e9c9} \textbf{\small{5.97}}         & \cellcolor[HTML]{d6e9c9} \textbf{\small{7.85}}               & \cellcolor[HTML]{d6e9c9}\small{66.32} & \cellcolor[HTML]{d6e9c9}\small{75.84} & \cellcolor[HTML]{d6e9c9}\small{70.24} & \cellcolor[HTML]{d6e9c9}\small{62.51} & \cellcolor[HTML]{d6e9c9}\small{50.00} & \cellcolor[HTML]{d6e9c9}\small{37.46} & \cellcolor[HTML]{d6e9c9}\textbf{\small{60.40}} \\
      \multicolumn{1}{c}{}                                & \ours [Ours]                     & \small{3$\times$1}              & \cellcolor[HTML]{d6e9c9} \textbf{\small{5.35}}         & \cellcolor[HTML]{d6e9c9} \textbf{\small{7.07}}               & \cellcolor[HTML]{d6e9c9}\small{66.80} & \cellcolor[HTML]{d6e9c9}\small{77.59} & \cellcolor[HTML]{d6e9c9}\small{73.79} & \cellcolor[HTML]{d6e9c9}\small{65.27} & \cellcolor[HTML]{d6e9c9}\small{54.47} & \cellcolor[HTML]{d6e9c9}\small{41.89} & \cellcolor[HTML]{d6e9c9}\textbf{\small{63.30}} \\

      \bottomrule
    \end{tabular}}
  \vspace{-3ex}
\end{table}

\subsection{Generation Quality}

\cref{tab:generation} illustrates the generated text given zero-shot arbitrary questions from \llamabig with \halffp weights and \ours using three kernels.
It is clearly that our model is able to produce meaningful and coherennt text compared to the original \halffp model.

\begin{table}[H]
    \setlength{\tabcolsep}{5.8pt}
      \centering
      \caption{Comparison of generation quality (\emph{non-curated}) on \llamabig models with \halffp and \ours using three kernels.} \label{tab:generation}
      \label{tab:prompts}
      \renewcommand{\arraystretch}{1.5}
      \scalebox{.93}{
        \begin{tabular}{rp{12cm}}
            \toprule
            \textbf{Prompt:} & \textcolor{Blue}{\texttt{What are the capitals of China, US and Russia?}}                                                                            \\
            \textbf{FP16:}  & \textcolor{PineGreen}{\texttt{The capital of China is Beijing. The capital of the United States is Washington D.C. and the capital of Russia is Moscow.}} \\
            \textbf{MBOK:}   & \textcolor{Mahogany}{\texttt{The capital of China is Beijing. The capital of the US is Washington D.C. and the capital of Russia is Moscow.}}         \\
            \midrule 
            \textbf{Prompt:} & \textcolor{Blue}{\texttt{I have a friend who lives in London and I would like to visit him. How would I go there from Paris?}}                                                                            \\
            \textbf{FP16:}  & \textcolor{PineGreen}{\texttt{You can take the Eurostar from Gare du Nord in Paris to St. Pancras Station in London. The trip takes about 2 hours and 15 minutes.}} \\
            \textbf{MBOK:}   & \textcolor{Mahogany}{\texttt{The fastest way to get from Paris to London is to take the Eurostar train. The Eurostar departs from the Gare du Nord station in Paris and arrives at the St. Pancras International station in London. The trip takes about 2 hours and 15 minutes.}}         \\ 
            \midrule 
            \textbf{Prompt:} & \textcolor{Blue}{\texttt{What is Fete du Travail?}   }                                                                         \\
            \textbf{FP16:}  & \textcolor{PineGreen}{\texttt{It is an annual holiday celebrated all over the world that resulted from the labour union movement, to celebrate the economic and social achievements of workers. The majority of countries celebrate Labour Day on May Day, or May 1, and it is popularly known as May Day and International Workers' Day}} \\
            \textbf{MBOK:}   & \textcolor{Mahogany}{\texttt{In France, May 1st is a public holiday. It is also known as Labour Day, International Workers\_ Day, or May Day. It is a celebration of the social and economic achievements of the labour movement. It originated from the eight-hour day movement, which advocated eight hours for work, eight for recreation, and eight for rest.}}         \\
            \midrule 
            \textbf{Prompt:} & \textcolor{Blue}{\texttt{What are the main ingredients in pho, Vietnam\_s iconic noodle soup?}  }                                                                          \\
            \textbf{FP16:}  & \textcolor{PineGreen}{\texttt{Rice noodles, beef or chicken, onions, ginger, cilantro, bean sprouts, basil, lime, hoisin sauce, Sriracha, and fish sauce.}} \\
            \textbf{MBOK:}   & \textcolor{Mahogany}{\texttt{Bone broth, rice noodles, fish sauce, herbs and spices, and beef, chicken, or seafood, depending on the type of pho you\_re eating. Pho is usually served with bean sprouts, Thai basil, lime wedges, and hoisin and sriracha sauces.}}         \\
            \bottomrule 
        \end{tabular}
      }
\end{table}

\subsection{Additional Baselines} \label{sec:addtional_baselines}

\subsubsection{Comparions with QuIP and ShiftAddLLM}

Both \quip \citep{chee2023quip} and \shiftaddllm \citep{haoran2024addshiftllm} are \gls{PTQ} method for \glspl{LLM}.
\quip is a two-step process that leverages the insight that quantization performs better when weight and Hessian matrices are incoherent.
It uses an adaptive rounding procedure to minimize a quadratic proxy objective, which measures the error between the original and quantized weights.
Additionally, it applies pre- and post-processing steps using random orthogonal matrices to ensure the weight and Hessian matrices are incoherent.
Conversely, our method does not employ either these complicated pre- and post-processing steps or costly Hessian matrices.
Meanwhile, \shiftaddllm is a post-training reparameterization process,  which quantizes each weight matrix in the LLM into a set of binary matrices and group-wise scaling factors.
he original multiplication between activations and weights is then reparameterized into: (1) bitwise shifts for the activations, using the power-of-two quantized scaling factors, and (2) additions of the results, guided by the binary weight matrices; this process can be implemented using look-up tables (LUTs) on GPUs.

\cref{tab:addshiftllm} presents results on \textsc{opt} models, with competitor results extracted from their respective original papers.
Notably, \shiftaddllm utilizes a more computationally expensive group quantization, whereas our method does not.
Our results clearly demonstrate that our approach consistently and significantly outperforms these baselines, particularly in the 2-bit scenario.

\begin{table}[H]
    \setlength{\tabcolsep}{3.8pt}
    \centering
    \caption{Comparisons with \quip, \shiftaddllm using \textsc{opt} models.}
    \label{tab:addshiftllm}

    \renewcommand{\arraystretch}{1.2}

    \scalebox{.93}{
        \begin{tabular}{clccc}
            \toprule
            \textsc{bit-width}  & \textsc{method}                            & \optsmall              & \optmedium             & \optbig                \\
            \midrule
            \midrule
                                & \quip \citep{chee2023quip}                 & \small{34.22}          & \small{25.19}          & \small{16.21}          \\
                                & \shiftaddllm \citep{haoran2024addshiftllm} & \small{31.29}          & \small{24.24}          & \small{21.53}          \\
            \multirow{-3}{*}{2} & \ours [Ours]                               & \textbf{\small{29.10}} & \textbf{\small{23.12}} & \textbf{\small{15.03}} \\
            \midrule
                                & \quip \citep{chee2023quip}                 & \small{347.40}         & \small{672.30}         & \small{41.64}          \\
                                & \shiftaddllm \citep{haoran2024addshiftllm} & \small{51.15}          & \small{40.24}          & \small{29.03}          \\
            \multirow{-3}{*}{3} & \ours [Our]                                & \textbf{\small{28.60}} & \textbf{\small{24.54}} & \textbf{\small{16.13}} \\
            \bottomrule
        \end{tabular}
    }

\end{table}

\subsubsection{Comparions with BitStack, DB-LLM and AWQ}

While \bitstack \citep{wang2025bitstack} also decompose weights using SVD, its core method and goal fundamentally differ from our method.
\bitstack is  a training-free method primarily aimed at saving storage for inference. 
In contrast, our method not only converts \gls{FP} models into Boolean models but also includes further fine-tuning, with the goal of achieving low complexity in both training and inference.
Furthermore, while \bitstack packs the extracted binary matrix into GPU-supported data types to reduce inference memory, and its approach to loading residual blocks relies on their influence on perplexity, our approach to residual block management is distinct.

\dbllm \citep{chen2024dbllm}  is limited to a fixed decomposition into two binary matrices, whereas our \ours method generalizes to an arbitrary number of Boolean kernels.
In \dbllm, the full-precision knowledge is preserved only through scaling factors and binary matrices derived implicitly via thresholding.
There is no formal analysis proving the optimality of this formulation.
In contrast, thanks to the \gls{SVID} in our approach, each extracted kernel is accompanied by an optimal scaling vector and Boolean matrix.
This allows us to only finetune the last kernel to calibrate the entire model.
Like most existing binary \glspl{LLM}, \dbllm relies on full-precision latent weights 
 during training and finetuning. Our method does not require this, as it directly operates in the Boolean domain.
 This distinction is particularly important in the \gls{LLM} context, where training and finetuning can be computationally expensive.

\cref{tab:awq_bitstack} compares our method, \ours (with 2 kernels), against \bitstack, \dbllm, and \awq \citep{Lin2024AWQ} on \llamatwosmall.  
It is evident that our method consistently outperforms all baselines.

\begin{table}[H]
    \setlength{\tabcolsep}{3.8pt}
    \centering
    \caption{Comparisons with \awq, \bitstack, \dbllm using \llamatwosmall with 2-bit setting.}
    \label{tab:awq_bitstack}

    \renewcommand{\arraystretch}{1.2}

    \scalebox{.93}{
        \begin{tabular}{lcccccc}
            \toprule
            \textsc{method}                    & \small{Wiki2} $(\downarrow)$ & \small{ARc-e} $(\uparrow)$ & \small{ARC-c} $(\uparrow)$ & \small{PIQA} $(\uparrow)$ & \small{Hella.} $(\uparrow)$ & \small{WinoG.} $(\uparrow)$ \\
            \midrule \midrule
            \awq \citep{Lin2024AWQ}            & \small{1.8e5}                & \small{26.3}               & \small{26.7}               & \small{50.9}              & \small{26.5}                & \small{49.3}                \\
            \bitstack \citep{wang2025bitstack} & \small{29.93}                & \small{32.3}               & \small{25.6}               & \small{62.4}              & \small{42.8}                & \small{53.6}                \\
            \dbllm \citep{chen2024dbllm}       & \small{7.23}                 & \textbf{\small{45.2}}      & \small{33.5}               & \small{73.1}              & \small{61.9}                & \small{61.7}                \\
            \ours [Ours]                       & \textbf{\small{6.87}}        & \small{44.8}               & \textbf{\small{34.2}}      & \textbf{\small{75.0}}     & \textbf{\small{65.6}}       & \textbf{\small{61.7}}       \\
            \bottomrule
        \end{tabular}
    }

\end{table}

\subsection{Effects of Knowledge Distillation} \label{sec:effect_kd}

\begin{figure}[H]
    \begin{subfigure}[c]{0.76\textwidth}
        \centering
        \tikzexternaldisable
        \centering
        \scriptsize
        \setlength{\figurewidth}{4.1cm}
        \setlength{\figureheight}{3.2cm}
        \input{figures/opt_kd_ablation.tikz}
        \tikzexternalenable
    \end{subfigure}
    \hfill
    \begin{subfigure}[c]{0.23\textwidth}
        \centering
        \tikzexternaldisable

\scalebox{.7}{
    \setlength{\tabcolsep}{3.9pt}
    \renewcommand{\arraystretch}{1.2}
    \begin{tabular}{lcc}
        \toprule
        & Wiki2 & C4 \\
        \midrule
        {\protect\tikz[baseline=-1ex]\protect\draw[color=color_red, fill=color_red, opacity=0.99, mark size=1.7pt, line width=1.7pt] plot[] (-0.0,0)--(-0.45,0);} KD & 
        \textbf{31.47} & \textbf{28.62}    \\
        {\protect\tikz[baseline=-1ex]\protect\draw[color=color_blue, fill=color_blue, opacity=0.99, mark size=1.7pt, line width=1.7pt] plot[] (-0.0,0)--(-0.45,0);} No KD & 32.21 & 29.84     \\
        \bottomrule
    \end{tabular}
}

\tikzexternalenable
    
    \end{subfigure}
    \caption{Study on the effect of using knowledge distillation on \optsmall with 2 Boolean kernels.  \label{fig:opt_kd_ablation}}
\end{figure}

\cref{fig:opt_kd_ablation} presents a comparison between training with and without Knowledge Distillation (KD).
It is evident that employing KD outperforms the baseline in terms of test perplexity on the Wiki2 and C4 datasets, as it provides more informative guidance during training.
To investigate this behavior further, we visualize the convergence of $\cL_{\textrm{logits}}$ and $\cL_{\textrm{is}}$.
Aided by the informative guidance from the teacher, convergence with KD is significantly faster. Furthermore, the model learns more effectively—leveraging the additional signal from the teacher—as evidenced by the higher flipping rates compared to training without KD.

\subsection{Analysis of Scaling Values}

\begin{figure}[H]
    \centering
    \includegraphics[scale=0.9]{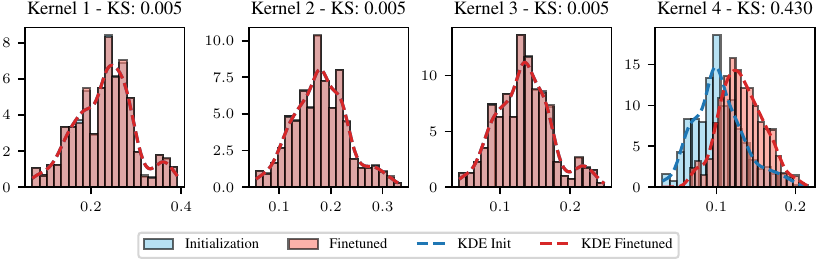}
    \caption{Histogram of output-scaling values for the first linear layer of \optsmall with four kernels, shown at initialization and after finetuning. The Kolmogorov–Smirnov (KS) statistic is also reported to quantify the difference between the scaling-value distributions before and after finetuning.
        \label{fig:out_scales_comparison}
    }   
\end{figure}

\cref{fig:out_scales_comparison} illustrates how the output-scaling values of the four kernels in the first linear layer change from initialization to after finetuning. All output-scaling values are learnable; however, only those associated with the last kernel exhibit a substantial shift during training. This is evident from both the histogram changes and the corresponding Kolmogorov–Smirnov distance.

After finetuning, the scaling values of the last kernel become significantly larger and more dominant, whereas the scaling values of the other kernels change only minimally. This observation supports our theoretical analysis: successive SVID extraction provides sufficiently strong initialization for the low-order kernels, and finetuning primarily the last kernel is already adequate.

\subsection{Convergences of OPT Models}

\cref{fig:opt_convergences} shows the training convergences of \ours using 3 kernels with OPT models.

\begin{figure}
    \begin{subfigure}[c]{\textwidth}
        \centering
        \tikzexternaldisable
        \centering
        \scriptsize
        \setlength{\figurewidth}{5.1cm}
        \setlength{\figureheight}{3.3cm}
        \input{figures/opt_convergence_125m.tikz}
        \tikzexternalenable
        \vspace{-2ex}
        \caption{OPT-125M}
    \end{subfigure}

    \vspace{3ex}

    \begin{subfigure}[c]{\textwidth}
        \centering
        \tikzexternaldisable
        \centering
        \scriptsize
        \setlength{\figurewidth}{5.1cm}
        \setlength{\figureheight}{3.3cm}
        \input{figures/opt_convergence_350m.tikz}
        \tikzexternalenable
        \vspace{-2ex}
        \caption{OPT-350M}
    \end{subfigure}

    \vspace{3ex}

    \begin{subfigure}[c]{\textwidth}
        \centering
        \tikzexternaldisable
        \centering
        \scriptsize
        \setlength{\figurewidth}{5.1cm}
        \setlength{\figureheight}{3.3cm}
        \input{figures/opt_convergence_1.3b.tikz}
        \tikzexternalenable
        \vspace{-2ex}
        \caption{OPT-1.3B}
    \end{subfigure}

    \vspace{3ex}

    \begin{subfigure}[c]{\textwidth}
        \centering
        \tikzexternaldisable
        \centering
        \scriptsize
        \setlength{\figurewidth}{5.1cm}
        \setlength{\figureheight}{3.3cm}
        \input{figures/opt_convergence_2.7b.tikz}
        \tikzexternalenable
        \vspace{-2ex}
        \caption{OPT-2.7B}
    \end{subfigure}

    \vspace{3ex}

    \begin{subfigure}[c]{\textwidth} 
        \centering
        \tikzexternaldisable
        \centering
        \scriptsize
        \setlength{\figurewidth}{5.1cm}
        \setlength{\figureheight}{3.3cm}
        \input{figures/opt_convergence_6.7b.tikz}
        \tikzexternalenable
        \vspace{-2ex}
        \caption{OPT-6.7B}
    \end{subfigure}

    \vspace{3ex}

    \begin{subfigure}[c]{\textwidth}
        \centering
        \tikzexternaldisable
        \centering
        \scriptsize
        \setlength{\figurewidth}{5.1cm}
        \setlength{\figureheight}{3.3cm}
        \input{figures/opt_convergence_30b.tikz}
        \tikzexternalenable
        \vspace{-2ex}
        \caption{OPT-30B}
    \end{subfigure}

    \caption{The training convergences of \ours using 3 kernels with \textsc{opt} models. \label{fig:opt_convergences}}
\end{figure}

\subsection{Effects of Successive SVID Initialization}

\begin{figure}[H]
    \begin{subfigure}[c]{0.75\textwidth}
        \centering
        \tikzexternaldisable
        \centering
        \scriptsize
        \setlength{\figurewidth}{4.1cm}
        \setlength{\figureheight}{3.2cm}
        \input{figures/opt_random_init.tikz}
        \tikzexternalenable
    \end{subfigure}
    \hfill
    \begin{subfigure}[c]{0.24\textwidth}
        \centering
        \tikzexternaldisable

\scalebox{.7}{
    \setlength{\tabcolsep}{3.9pt}
    \renewcommand{\arraystretch}{1.2}
    \begin{tabular}{lcc}
        \toprule
        & Wiki2 & C4 \\
        \midrule
        {\protect\tikz[baseline=-1ex]\protect\draw[color=color_red, fill=color_red, opacity=0.99, mark size=1.7pt, line width=1.7pt] plot[] (-0.0,0)--(-0.45,0);} Suc. SVID & 
        \textbf{31.47} & \textbf{28.62}    \\
        {\protect\tikz[baseline=-1ex]\protect\draw[color=color_blue, fill=color_blue, opacity=0.99, mark size=1.7pt, line width=1.7pt] plot[] (-0.0,0)--(-0.45,0);} Rand. Init. & 1.6E4 & 1.1E4     \\
        \bottomrule
    \end{tabular}
}

\tikzexternalenable
    
    \end{subfigure}
    \caption{Study on the effect of using our successive \gls{SVID} strategy and random initialization on \optsmall with 2 Boolean kernels.  \label{fig:opt_random_init}}
\end{figure}

\cref{fig:opt_random_init} compares our proposed successive \gls{SVID} initialization with a random initialization. It is clear that our method delivers far better results, while the random initialization often fails to converge. Moreover, our initialization enables the model to learn efficiently, whereas the random initialization causes the model to struggle, as reflected by the large number of Boolean flips.
\subsection{Discussion on Latency and Comparison with Vector Quantization } \label{sec:latency_vq}

\paragraph{Scalar and Vector Quantization.} In the context of \glspl{LLM}, scalar quantization and vector quantization are two different approaches for compressing weights.
Scalar quantization maps each weight or activation independently to a smaller set of discrete levels (e.g., 32-bit floating-point to 8- or 4-bit integers). It is simple, hardware-friendly, and widely used in practice, but it ignores correlations across dimensions, potentially discarding fine-grained structure.
Vector quantization (VQ) instead compresses entire vectors (e.g.,  weight groups) by replacing them with indices into a learned codebook of representative vectors. By capturing cross-dimensional correlations, VQ often achieves higher compression, particularly for large embedding tables. However, codebook training is more complex, and inference requires index lookups to reconstruct vectors. This adds significant overhead to both quantization and dequantization, leading to much higher latency compared to scalar methods.

Our method is native 1-bit weight design, its nearest baselines are scalar weight quantization.
As a result, for a fair comparison, in the main text we mainly consider state-of-the-art scalar quantization like \omniquant \citep{shao2024omniquant}, \qptq \citep{frantar2023optq}, \llmqat \citep{liu2024llmqat} as the main baselines.
Nevertheless, for completeness, we also compare our approach against state-of-the-art ultra low-bit vector quantization (VQ) methods for \glspl{LLM}, including \qtip \citep{tseng2024qtip} and \quipsharp \citep{tseng2024quipsharp} in a 2-bit setting, specifically on \llamasmall with a sequence length of 2048 (results taken from the \qtip paper).
The results are summarized in \cref{tab:qtip_quipsharp}.
Remarkably, our method's performance is comparable to these \gls{SOTA} VQ methods.
This is noteworthy given that our approach directly utilizes native Boolean weights, eliminating the need for the very costly quantization and dequantization of high-dimensional vectors inherent in VQ.

\begin{table}[H]
    \setlength{\tabcolsep}{3.8pt}
    \centering
    \caption{Perplexity comparison with SOTA vector quantization methods using \llamasmall. }
    \label{tab:qtip_quipsharp}

    \renewcommand{\arraystretch}{1.2}

    \scalebox{.97}{
        \begin{tabular}{lcc}
            \toprule
            \textsc{method}                       & \small{Wiki2} ($\downarrow$) & \small{C4} ($\downarrow$) \\
            \midrule
            \midrule
            \quipsharp \citep{tseng2024quipsharp} & \small{6.86}                 & \small{8.36}              \\
            \qtip \citep{tseng2024qtip}           & \small{6.52}                 & \small{7.99}              \\
            \ours [Ours]                          & \small{6.83}                 & \small{8.53}              \\
            \bottomrule
        \end{tabular}
    }
\end{table}

\paragraph{Empirical Evidence of Latency Gains.}

To demonstrate the practicality of our approach even on modern hardware such as GPUs, we leverage the recently introduced BitBLAS library \footnote{\url{https://github.com/microsoft/BitBLAS}} \citep{ladderosdi24} for 1-bit matrix multiplications.
Using FP16 activations with INT1 weights, we measure the latency of linear layers in \llamasmall (\cref{tab:latency_llama_7b}) and \llamabig (\cref{tab:latency_llama_13b}) under an inference batch size of 1, evaluating our method \ours with two kernels.
Our results show that \ours achieves up to an $8.7\times$ speedup over FP16 baselines, while substantially outperforming existing binarization and scalar quantization methods, as detailed in the main text.
We also benchmark against 2-bit \quipsharp and \qtip using the authors’ official implementations\footnote{\url{https://github.com/Cornell-RelaxML/quip-sharp}}\footnote{\url{https://github.com/Cornell-RelaxML/qtip}}.
All experiments are conducted on an A100 GPU.

Remarkably, our method is not only much faster than these VQ baselines but also delivers comparable performance. This is expected, as VQ-based methods incur significant overhead from the costly encoding and decoding steps required to realize their high compression ratios.
Taken together, the results highlight that our native Boolean approach offers a compelling and efficient alternative to state-of-the-art vector quantization methods. With dedicated Boolean hardware accelerators, the performance gains would be even more pronounced.

\begin{table}[H]
    \setlength{\tabcolsep}{3.8pt}
    \centering
    \caption{Measured latency (ms) of linear layers in  \llamasmall, with values in parentheses denoting speed-up relative to the \halffp baseline.}
    \label{tab:latency_llama_7b}

    \renewcommand{\arraystretch}{1.2}

    \scalebox{.95}{
        \begin{tabular}{cccccccc}
            \toprule
            \textsc{weight size}         & \halffp           & \quipsharp  \citep{tseng2024quipsharp}                   & \qtip \citep{tseng2024qtip}                           & \ours (Ours)                                                        \\
            \midrule
            \midrule
            \small{$4096 \times 4096$}   & \small{$0.10697$} & \small{$0.46196$} ($0.23\times$) & \small{$1.37137$} ($0.08\times$) & \cellcolor[HTML]{d6e9c9} \small{$\mathbf{0.04989}$ ($\mathbf{2.14}\times$) } \\
            \small{$4096 \times 11008$}  & \small{$0.27935$} & \small{$0.55526$} ($0.50\times$) & \small{$3.13633$} ($0.09\times$) & \cellcolor[HTML]{d6e9c9} \small{$\mathbf{0.05136}$ ($\mathbf{5.44}\times$) } \\
            \small{$11008 \times  4096$} & \small{$0.27664$} & \small{$0.55988$} ($0.49\times$) & \small{$3.16067$} ($0.09\times$) & \cellcolor[HTML]{d6e9c9} \small{$\mathbf{0.05117}$ ($\mathbf{5.41}\times$) } \\
            \bottomrule
        \end{tabular}
    }

\end{table}

\begin{table}[H]
    \setlength{\tabcolsep}{3.8pt}
    \centering
    \caption{Measured latency (ms) of linear layers in  \llamabig, with values in parentheses denoting speed-up relative to the \halffp baseline.}
    \label{tab:latency_llama_13b}

    \renewcommand{\arraystretch}{1.2}

    \scalebox{.95}{
        \begin{tabular}{cccccccc}
            \toprule
            \textsc{weight size}          & \halffp           & \quipsharp  \citep{tseng2024quipsharp} & \qtip \citep{tseng2024qtip}      & \ours (Ours)                                                                 \\
            \midrule
            \midrule
            \small{$5120  \times 5120 $}  & \small{$0.16540$} & \small{$0.62260$} ($0.27\times$)       & \small{$1.96368$} ($0.08\times$) & \cellcolor[HTML]{d6e9c9} \small{$\mathbf{0.05074}$ ($\mathbf{3.25}\times$) } \\
            \small{$5120  \times 13824$}  & \small{$0.42830$} & \small{$0.62836$} ($0.68\times$)       & \small{$5.23681$} ($0.09\times$) & \cellcolor[HTML]{d6e9c9} \small{$\mathbf{0.05098}$ ($\mathbf{8.40}\times$) } \\
            \small{$13824  \times  5120$} & \small{$0.43411$} & \small{$0.62840$} ($0.69\times$)       & \small{$5.21193$} ($0.08\times$) & \cellcolor[HTML]{d6e9c9} \small{$\mathbf{0.04987}$ ($\mathbf{8.70}\times$) } \\
            \bottomrule
        \end{tabular}
    }

\end{table}

\section{Ethics Statement}

This work makes a fundamental contribution to machine learning methodology. It does not involve human subjects, sensitive data, or applications with direct societal or ethical risks. We do not foresee any immediate ethical concerns arising from this research.

\section{Reproducibility Statement}
We provide detailed descriptions of all algorithms and illustrative code for the core components. Experiments are conducted on standard benchmarks using established testing procedures, and all experimental details and settings are fully declared to facilitate independent reproduction of our results.

\section{The Use of Large Language Models}
We used large language models (LLMs) solely for non-substantive assistance, including grammar refinement and summarizing relevant literature. All research ideas, analyses, and conclusions are the authors' own.

\end{document}